%% file: main.tex
\documentclass[runningheads]{llncs}

\usepackage{accv}

\usepackage{accvabbrv}

\usepackage{graphicx}
\usepackage{booktabs}
\usepackage{fontawesome}

\usepackage[accsupp]{axessibility}  %

\usepackage[pagebackref,breaklinks,colorlinks,citecolor=accvblue]{hyperref}

\usepackage{orcidlink}
\usepackage{epsfig}
\usepackage{amsmath}
\usepackage{amssymb}
\usepackage{booktabs}
\usepackage{colortbl}
\usepackage{textcomp}
\usepackage{subcaption}
\usepackage{standalone}
\usepackage{pgfplots} 
\usepackage{multirow}
\usepackage{xcolor}
\usepackage[inkscapelatex=false]{svg}
\usepackage{siunitx}
\usepackage{lipsum}
\usepackage{pifont}
\usepackage{yfonts}
\usepackage{listings}
\usepackage{algorithm}
\usepackage{algpseudocode}
\usepackage{makecell}
\usepackage[T1]{fontenc}
\usepackage[normalem]{ulem}
\usepackage{nicematrix}
\usepackage{placeins}
\usepackage{circledsteps}
\usepackage{scalerel}
\usepackage{todonotes}
\usepackage{ulem}
\usepackage{appendix}
\usepackage{comment}
\usepackage{wrapfig}
\usepackage{tikz}
\usepackage[skip=3pt]{caption}
\newcommand{\fakeparagraph}[1]{\noindent\textbf{#1}}

\input{mathematical_formulations}

\input{ter-colors}

\input{mathmacros}
\newcolumntype{g}{>{\columncolor{tu73}}c}
\newcolumntype{h}{>{\columncolor{tu74}}c}

\newcommand{\rot}[1]{\rotatebox{90}{#1}}
\definecolor{ForestGreen}{RGB}{34,170,34}
\definecolor{textcolor}{RGB}{70,170,34}

\begin{document}

\title{Strong but simple: A Baseline for Domain Generalized Dense Perception by CLIP-based Transfer Learning} 

\titlerunning{Strong but simple: A Baseline for Domain Generalized Dense Perception}

\author{Christoph Hümmer\inst{1,3}* \and
Manuel Schwonberg\inst{1,3}* \and
Liangwei Zhou\inst{2}*\and 
Hu Cao\inst{2} \and
Alois Knoll\inst{2}\and 
Hanno Gottschalk\inst{1}\\
\footnotesize * equal contribution, alphabetically ordered}

\authorrunning{Hümmer, Schwonberg, Zhou et al.}

\institute{Technical University Berlin\\ 
{\tt\small gottschalk@math.tu-berlin.de}
\and
Technical University Munich\\
{\tt\small \{hu.cao,liangwei.zhou,k\}@tum.de}\and
CARIAD SE\\
{\tt \small \{christoph.huemmer,manuel.schwonberg\}@cariad.technology}}

\maketitle

\begin{abstract}
  Domain generalization (DG) remains a significant challenge for perception based on deep neural networks (DNNs), where domain shifts occur due to synthetic data, lighting, weather, or location changes. Vision-language models (VLMs) marked a large step for the generalization capabilities and have been already applied to various tasks. Very recently, first approaches utilized VLMs for domain generalized segmentation and object detection and obtained strong generalization. However, all these approaches rely on complex modules, feature augmentation frameworks or additional models. Surprisingly and in contrast to that, we found that simple fine-tuning of vision-language pre-trained models yields competitive or even stronger generalization results while being extremely simple to apply. Moreover, we found that vision-language pre-training consistently provides better generalization than the previous standard of vision-only pre-training. This challenges the standard of using ImageNet-based transfer learning for domain generalization. Fully fine-tuning a vision-language pre-trained model is capable of reaching the domain generalization SOTA when training on the synthetic GTA5 dataset. Moreover, we confirm this observation for object detection on a novel synthetic-to-real benchmark. We further obtain superior generalization capabilities by reaching \textbf{77.9\%} mIoU on the popular Cityscapes$\rightarrow$ ACDC benchmark. We also found improved in-domain generalization, leading to an improved SOTA of \textbf{86.4\%} mIoU on the Cityscapes test set marking the first place on the leaderboard. 
  \keywords{Domain Generalization \and Semantic Segmentation \and Object Detection}
\end{abstract}

\section{Introduction}
\label{sec:intro}

Despite the recent advancements in computer vision due to deep learning \cite{rombach2021highresolution, kirillov2023segment}, domain shifts are still a major challenge for computer vision tasks. They can cause a significant performance decrease during inference time when the distribution differs from the training distribution \cite{tsai2018learning, Hoyer2022daformer}.
For this reason, two research fields have emerged to tackle domain shifts. Unsupervised domain adaptation (UDA) works with the assumption that unlabeled data from the target domain is available to adapt the network towards this particular domain. A large variety of methods have been developed in this field for both semantic segmentation \cite{Schwonberg2023Survey, hoffman2018cycada, tsai2018learning, tranheden2021dacs, zhang2021prototypical, zhang2021multiple, hoyer2022hrda, hoyer2023mic, Hoyer2022daformer} and objection detection \cite{mattolin2023confmix, chen2020harmonizing}. Nevertheless, unlabeled target data or information about the target domain(s) might not always be available during training time. In addition, data collection can be difficult due to  the sheer size of the operational domain.
\begin{figure}[t!]
  \centering
  \includegraphics[width=0.85\columnwidth]{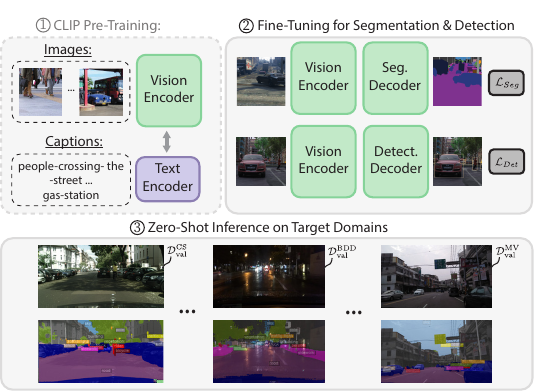}
  \caption{\textbf{CLIP-Based Fine-tuning for Domain Generalized Dense Perception}: Previous works in domain generalization research mostly focus on applying methods (e.g. augmentations, consistency losses) in downstream training. In contrast, we simply transfer CLIP-based pre-trained weights (1), conduct fine-tuning on two different tasks (2) and evaluate the performance across various unseen real-world domains (3). Hereby, we reach a SOTA performance in domain generalized semantic segmentation and object detection.}
  \label{fig:eyecatcher_fig1}
\end{figure}
For this reason, the field of domain generalization (DG) developed to improve the generalization from only a single (or multiple) source domain(s) to unseen target domains \cite{Lee2022wildnet, Pan2018, Choi2021}. A variety of methods exist for semantic segmentation and several of them utilize domain randomization techniques in the input space \cite{schwonberg2023augmentation, Lee2022wildnet, Huang2021, peng2021global} or constrained feature representation learning \cite{Peng2022semanticaware}. For object detection a few DG methods exist yet also utilizing consistency losses and contrastive learning \cite{bi2024good, vidit2023clip, wu2022single}. Recently, vision-language models like CLIP \cite{radford2021learning} with their remarkable zero-shot capabilities were harnessed to improve domain generalization for street scene segmentation \cite{benigmim2024collaborating, wei2024stronger, fahes2024simple, vidit2023clip}. However, all of these works require additional modules \cite{wei2024stronger}, additional models like SAM \cite{kirillov2023segment} and large language models \cite{benigmim2024collaborating} or feature augmentations \cite{fahes2024simple, vidit2023clip}. Fahes \etal \cite{fahes2024simple} show that fine-tuning with CLIP initialized weights does not work and gives a significantly worse performance compared to ImageNet. With our work, we challenge these results and present a simple yet effective solution solely based on fine-tuning to utilize the generalized vision-language representations. We achieve a competitive or higher performance for both segmentation and object detection compared to the mentioned more complex approaches. For this, we leverage CLIP \cite{radford2021learning} and EVA-CLIP \cite{EVA-CLIP} pre-trained ViT-encoders with a Mask2Former \cite{cheng2022masked} decoder along with a simple fine-tuning protocol. Without any additional modules or losses, our technique demonstrates strong generalization capabilities on synthetic-to-real and real-to-real benchmarks. We refer to these techniques as \textbf{VLTSeg} and \textbf{VLTDet}, which is an abbreviation for \textbf{V}ision-\textbf{L}anguage representation \textbf{T}ransfer for domain generalized \textbf{Seg}mentation or \textbf{Det}ection, respectively. Unlike previous DG approaches for dense perception, fine-tuning does not rely on or require input-level augmentations \cite{schwonberg2023augmentation, zhong2022adversarial, sun2023augment, zhao2022style, vidit2023clip}, consistency losses \cite{Lee2022wildnet, peng2021global, Yue2019, bi2024good} or complex whitening losses in the feature space \cite{pan2019switchable, Peng2022semanticaware, Choi2021} but offers a strong and easily applicable method. The contribution of our work is, therefore, multi-folded: 

\begin{itemize}
    \item In contrast to recent works, we show that our simple CLIP-based transfer-learning shows similar or stronger generalization capabilities without any additional modules or methods.
    \item We compare the generalization of state-of-the-art pre-trainings for dense perception and validate the effectiveness of vision-language pre-training.
    \item 
    In synthetic-to-real DG we reach a SOTA competitive performance in semantic segmentation of 63.2 \% in DG mean. We evaluate synthetic-to-real performance on a new multi-object detection benchmark based on UrbanSyn \cite{gomez2023all} with an absolute improvement ranging from 5.7 - 8.4\% in $\textrm{mAP}_{50}$. For real-to-real DG, we reach an improved SOTA on Cityscapes$\rightarrow$ACDC of 77.9\% and for the in-domain Cityscapes test dataset with 86.4\% mIoU.

\end{itemize}

\section{Related Work}
\textbf{Domain Generalization}      
Domain generalization (DG) approaches do not require any target data at all and only labeled source data. They can be divided into three major categories: distribution randomization, constrained feature learning and vision-language utilization which are related to this work. To constrain the learning to domain invariant features, instance normalization and whitening techniques are commonly used \cite{Pan2018, pan2019switchable, Choi2021, xu2022dirl, peng2021global} with different improvements like, e.g. semantic-aware whitening \cite{Peng2022semanticaware}.\\
In domain randomization, the visual appearance of the input is modified to learn domain-invariant features. Several of these approaches use additional external real-world data like ImageNet for style randomization, often in combination with consistency losses \cite{Yue2019, peng2021global, Lee2022wildnet, kim2023texture}. FSDR \cite{Huang2021} and PASTA \cite{2023iccv_PASTA} differ by applying style randomization and augmentation in the frequency space. WEDGE \cite{kim2021wedge} samples real images from the internet and utilizes them and self-training. \\
In \cite{schwonberg2023augmentation}, the authors show that augmentations can provide a state-of-the-art DG performance without external data. In contrast, both GBFA \cite{sun2023augment} and SHADE \cite{zhao2022style} employ feature-level augmentations. Recently, approaches tailored towards vision transformer architecture emerged in the field of DG, proposing enhanced attention mechanisms \cite{bi2023learning, ding2023hgformer, sun2023ibaformer}.
Parallel to our work, other works developed approaches based on vision-language models. PromptFormer, DGinStyle and DIDEX \cite{gong2023prompting, jia2023dginstyle, niemeijer2024generalization} all utilize generative latent diffusion models for a style transfer-based approach. Very recently, FAmix, CLOUDS and Rein \cite{fahes2024simple,benigmim2024collaborating, wei2024stronger} introduced methods with additional modules, feature-based augmentation and multiple different networks in addition to the vision-language trained models. In contrast to that, our simple but strong baseline does not need any additional modules, loss functions or networks but just relies on fine-tuning and achieves a competitive often even superior performance. Also, we demonstrate that our technique achieves SOTA performances for both segmentation and object detection which was only achieved by PASTA \cite{2023iccv_PASTA} before. Similar to our approach, Kerssies \etal \cite{kerssies2024benchmark} fine-tuned foundation models for semantic segmentation to investigate the influence of choices like the patch size on the model performance but didn't focus on domain generalization.\\
\textbf{Pre-Training and Generalization}
Several works investigated the impact of supervised ImageNet pre-training on the downstream task \cite{he2019rethinking, huh2016makes, zoph2020rethinking, feng2021rethinking, kornblith2019better, yamada2022does} or up-scaling it to the larger 21k-ImageNet pre-training \cite{ridnik2021imagenet}.\\  
In recent years, self-supervised image-based pre-training approaches have emerged. Earlier approaches \cite{he2020momentum, chen2020simple, grill2020bootstrap} already showed a strong transferability to downstream tasks. More self-supervised pre-training methods were proposed in the following namely MoCov3 \cite{chenempirical}, DINO \cite{caron2021emerging}, DINOv2 \cite{oquab2023dinov2}, iBOT \cite{zhou2021ibot}, masked image modeling (MIM) \cite{xie2022simmim} and BEiTv2 \cite{peng2022beit} demonstrating highly generalizing and robust representations which transfer well to downstream tasks. We exclude DinoV2 as it uses Cityscapes to sample its pre-training dataset and the knowledge of target datasets/domains
is a significant advantage compared to the largely uncurated datasets SO400M and Laion-2B employed in CLIP and EVA-02-CLIP.\\
With the development of CLIP \cite{radford2021learning}, supervised vision-language pre-training showed strong zero-shot robustness, making it a promising candidate for pre-training. Consequently, several works building upon CLIP were proposed, such as EVA-CLIP \cite{EVA-CLIP} further improving, upscaling, and robustifying the vision-language pre-training.\\
\textbf{Vision-Language Models}
Vision Language Models (VLMs) combine vision and text features by training on large image-text pair datasets. A significant advantage of this training is the availability of image-text caption data, which can be extracted from the web and does not require manual labeling. That leads to massive datasets like the publicly available LAION-5B \cite{schuhmann2022laionb} dataset. The training methods can be divided into three main branches: The first implements contrastive learning by enforcing the similarity of corresponding text caption and image embeddings \cite{radford2021learning, jia2021scaling, shen2022k, Zhou2022ConditionalPL, EVA-CLIP}. The second branch is about generative cost functions, for instance, masked image modeling and masked text modeling pre-training strategies \cite{singh2022, bao2022, Wang2023, geng2022multimodal, zeng2022x}, and the third branch introduces input modality adapters \cite{Alayrac2022, li2023otter, kosmos-2}. Several works adopted CLIP for tasks like object detection or semantic segmentation to profit from the classification capabilities and features. A straightforward extension is to apply CLIP to region-text pairs \cite{zhong2022regionclip, Wang_2023_CVPR, Shi_2023_ICCV} to enable classification in object detection. Similarly, semantic segmentation can be conducted based on CLIP by employing segment-text alignment \cite{He_2023_CVPR, Liang_2023_CVPR, yu2023fcclip, Zhou_2023_CVPR}.

\section{Method}
In this section, we will describe the general setting for our simple transfer learning setting and the investigation of pre-training for domain generalized dense perception.

\subsection{General Setting}
An input image is defined as $\mathbf{x}_n \in \mathbb{G}^{H \times W \times C}$ with $\mathbb{G}$ being the integer color values, $H$ and $W$ height and width respectively and $C$ denoting the number of channels. A deep segmentation network is defined as $\VEC{M}$ mapping the input images $\mathbf{x}_n$ to output probability maps denoted as $\mathbf{y}_n = (y_{n,i,s})\in \mathbb{I}^{H \times W \times S}$. They represent the class posterior probabilities $y_{n,i,s} = P(s|i,\mathbf{x}_n)$  for each class $s \in \mathcal{S}$ at pixel index $i \in \mathcal{I} = \{1,2,...,H\cdot W\}$ with $\mathbb{I}=[0,1]$. We furthermore define the encoder and decoder of a network $\VEC{M}$ as $\VEC{M}_E$ and $\VEC{M}_D$, respectively. For the vision and language encoders we introduce the superscripts $V$ and $L$ respectively. We describe the vision encoder as $\VEC{M}_E^V$ and the language encoder as $\VEC{M}_E^L$.  $\mathcal{T}$ denotes the space of text descriptions. Superscripts ``$\mathrm{S}$'' and ``$\mathrm{T}$'' on $\mathbf{x}_n$ and $\mathbf{y}_n$ denote the domain from which the variables stem, with e.g., ${\src}$ being the source domain and ${\tgt}$ being the target domain. Since there are multiple possible unseen target domains in domain generalization, the target domains $\mathcal{D}^{\mathrm{T}_k}$ are indexed with $k \in \mathcal{K} = \{1,2,...,K\}$. In domain generalization, we seek to train a model $\VEC{M}$ which generalizes well to these unseen target domains $\mathcal{D}^{\mathrm{T}_k}$. We provide a more extensive mathematical motivation of our approach in the supplement.\\
One of the intriguing properties of the multimodal vision-language models like CLIP \cite{radford2021learning} is the zero-shot robustness against domain shifts. That indicates that vision-language trained feature extractors exhibit a strong semantic generalization due to the domain-invariant language modality and the massive size of the datasets. This inherent generalization may be easy to transfer to downstream tasks without complex generalization methods as previous approaches. Therefore, we aim to utilize this robustness of VLMs against domain shifts with VLTSeg and VLTDet for domain generalized dense perception.   

\subsection{Vision-Language based Transfer learning}
\label{trwcbve}
In contrast to previous works, which employ complex auxiliary objectives, we investigate domain generalization by only fine-tuning strongly generalized encoders with a simple protocol.
The straightforward way to employ pre-trained encoder weights for semantic segmentation or other perception tasks such as object detection is the utilization of a task-specific head and then fine-tuning the entire network $\VEC{M}$ on the downstream task, which is often done in computer vision \cite{caron2021emerging,zhou2021ibot, oquab2023dinov2, xie2022simmim} and domain generalization \cite{Yue2019, peng2021global, Lee2022wildnet, kim2023texture} with additional DG methods. The encoder in our experiments $\VEC{M}_E^V$ is initialized with either CLIP \cite{radford2021learning} or EVA-CLIP \cite{fang2023eva} pre-trained weights, which are trained on image-text pairs.\\ We focus on transformer-based architectures ViT \cite{Dosovitskiy2021} and EVA \cite{EVA-CLIP} since they benefit more from large-scale pre-training and have a higher capacity than CNNs, which benefits generalization, as shown previously by \cite{goldblum2023battle}. Moreover, we select the recently introduced transformer-based Mask2Former \cite{cheng2022masked} as the decoder for semantic segmentation. This decoder based on mask classification was robust across multiple segmentation tasks and, therefore, fits well with our target objective of domain generalization. We chose the ViTDet \cite{LiMGH2022} as a simple decoder for our object detection experiments on top of the backbone. We are fully fine-tuning the entire network $\VEC{M}$ and will validate and discuss this strategy with an ablation study of (partial) layer freezing.

\subsection{Relative Performance under Domain Shift}
In domain adaptation and generalization the common evaluation metric is the mean intersection over union (mIoU) or mean average precision (mAP). However, for the real-to-real domain shift evaluation (see Table \ref{tab:r2r_ablation}) we analyze the performance drop from in- to out-of-domain performance. For this reason we adapt the common robustness metric relative performance under corruption (rPC) \cite{michaelis2019benchmarking} since corruptions are technically one form of domain shift. Based on their definition we define our relative performance under domain shift (rPD) metric as: 
\begin{equation}
    \textrm{rPD} = \frac{\frac{1}{K} \sum_{k=1}^{K} \textrm{mIoU}^{\mathrm{T}_k}}{\textrm{mIoU}^S}
\end{equation}
The intuition behind the rPD is similar to the rPC. It measures the, over all target domains $K$ averaged, relative performance drop caused by domain shift between the source domain with its oracle performance $\textrm{mIoU}^S$ and the reduced $\textrm{mIoU}^{\mathrm{T}_k}$ on the unseen target domains $\mathrm{T}_k$. We apply the same metric to object detection by exchanging $\textrm{mIoU}$ with $\textrm{mAP}_{50}$.

\section{Experimental Settings}
\label{sec:experiments}
In the following, we introduce firstly the employed datasets and metrics. Afterwards, we introduce the network architectures and implementation details.
\subsection{Datasets and Metrics}
\textbf{Datasets:} According to the common domain generalization standard setting \cite{gong2023prompting,Lee2022wildnet,Peng2022semanticaware,wang2020differential,sun2023augment} we employ the two synthetic datasets SYNTHIA (SYN) \cite{ros2016synthia} and GTA5 \cite{richter2016playing} as our source domains for our synthetic-to-real experiments. The datasets contain 9400 and 24966 images, respectively. UrbanSyn \cite{gomez2023all} with 7540 synthetic images is used for object detection.  We utilize Cityscapes \cite{cordts2016cityscapes}, BDD100k \cite{yu2020bdd100k}, Mapillary \cite{neuhold2017mapillary} and ACDC \cite{sakaridis2021acdc} as our real-world target domains with 500, 1000, 2000 and 406 validation images, respectively. The training sets of these datasets remain unused in the synthetic-to-real experiments. As a common practice in domain generalization \cite{kim2021wedge,yue2019domain,sun2023ibaformer}, we evaluate the experiments on the validation sets $\mathcal{D}_{\text{val}}$ of the real-world target domain datasets and compute the domain generalization (DG) mean over Cityscapes, Mapillary, BDD and ACDC. 
\textbf{Metrics:} Next to our novel rPD metric, we follow common DG research practice for the evaluation metrics \cite{klingner2022unsupervised, huang2021fsdr, Termoehlen2023arxiv}. The mean intersection over union (mIoU) of $S = 19$ (SYNTHIA trainings only $S = 16$) segmentation classes \cite{cordts2016cityscapes,richter2016playing,sakaridis2021acdc} and the $\textrm{mAP}_{50}$ of $S = 8$ classes is used for the majority of experiments. In addition, the mean of mIoUs and $\textrm{mAP}_{50}$ on the target datasets is computed and described as DG mean.

\subsection{Data Augmentations and Baselines}
We investigate a set of augmentations and domain generalization methods that were shown to be effective. The augmentations consist of PixMix \cite{Hendrycks2022pixmix}, RandAugment \cite{Cubuk2022}, and PASTA  \cite{2023iccv_PASTA} and are applied on every image except PixMix, which is applied randomly with a probability of 0.5. For PixMix, we employ the parameters $k=4$ for mixing rounds and $\beta=3$ for the mixing coefficients. For RandAugment, we only use color space augmentations which consist of ColorTransform, AutoContrast, Equalize, Sharpness, Posterize, Solarize, Color, Contrast, Brightness. Moreover, we incorporate the recent method TLDR \cite{kim2023texture}, which relies on texture regularization losses, and ReVT \cite{Termoehlen2023}, which aims to improve generalization by model averaging.

\subsection{Network Architectures} 

\indent\textbf{Encoder} 
We utilize vision transformer (ViT)-based backbones \cite{Dosovitskiy2021} for all our experiments but with different initialization. Our default versions are the ViT-Large encoder with patch-size 14, and the EVA-02 \cite{EVA-CLIP} encoder, denoted as \network{ViT-L-14} and \network{EVA-02-L-14} respectively. For our ablation studies also ViT-Base, denoted as \network{ViT-B-16}, is used. Note that we do not apply any own vision-language pre-training. For the synthetic-to-real experiments with a ResNet-backbone (see supplementary material), we also employed a \network{ResNet-101} \cite{he2016deep} encoder.  \\
\textbf{Decoder Architecture} Our standard decoder for semantic segmentation architecture is the Mask2Former \cite{cheng2022masked} due to its generality. For ablation purposes, we also used the \cite{rao2022denseclip} FPN decoder \cite{kirillov2019panoptic}, the ASPP-based DAFormer decoder \cite{Hoyer2022daformer} and the SegFormer decoder \cite{Xie2022segformer}. For object detection we used the ViTDet \cite{LiMGH2022} decoder without the mask head. All decoders were randomly initialized in our experiments. 

\subsection{Implementation Details}
Our experiments are implemented based on the open-source toolbox MMSegmentation \cite{contributors2020mmsegmentation} and MMDetection \cite{chen2019mmdetection}. All experiments are conducted on 2 $\times$ A100 GPUs with 80GB each. For Mask2Former losses we employ $\lambda_{\rm CE}=2.0$, $\lambda_{\rm BCE}=5.0$ and $\lambda_{\rm DICE}=5.0$ following the authors protocol \cite{cheng2022masked}. For a fair comparison with previous works, we used two different settings for the synthetic-to-real and real-to-real experiments. The synthetic-to-real experiments for semantic segmentation were conducted with a crop size of 512 $\times$ 512 and a batch size 16. Only 5k fine-tuning iterations were necessary in this setting to avoid overfitting to the synthetic source domain. 
\setlength{\textfloatsep}{0pt}
\setlength{\intextsep}{4pt}
\setlength{\abovecaptionskip}{2pt}
\begin{table}
  \centering
  \caption{\textbf{Transfer Learning Domain Generalization Performance} of supervised training on $\gtavtrain$ and $\synthiatrain$.* denotes that the \network{ViT-L-16} was used as the encoder. $\circ$ denotes that \network{ViT-L-14} was used as the encoder. \faLock \ denotes frozen encoder weights.}
  \include{tables/m2f_transfer_learning}
  \label{tab:m2f_transfer} 
\end{table}
The real-to-real experiments used a larger crop size of 1024 $\times$ 1024 and batch size 8 to better compare with HGFormer \cite{ding2023hgformer} and all trained on 20k iterations. We refer to both setups as \textbf{VLTSeg}. For object detection all trainings were conducted with a crop size of 1024 $\times$ 1024 and batch size 14. We refer to these setups as \textbf{VLTDet}. For evaluation, we always select the last checkpoint of a training in line with previous works \cite{schwonberg2023augmentation, niemeijer2024generalization}.\\
We use the same basic augmentations for all segmentation experiments: random resizing, random cropping, horizontal flip and color jitter. For object detection we use Mosaic, RandomAffine, Mixup and RandomFlip as default augmentations with RandAugment added for VLTDet. 
As the optimizer, we employ AdamW \cite{loshchilov2018decoupled} for all our experiments as it is a common standard for training vision transformer models \cite{Hoyer2022daformer, Termoehlen2023arxiv, cheng2022masked} with a default learning rate of 0.0001, and the backbone learning rate is established at one-tenth of the default rate for segmentation and decayed per layer with a rate of 0.7 for object detection. A learning rate schedule with a warm-up and decay was employed. More extensive implementations details are provided in the supplement.

\section{Results}
\label{sec:results}
In the following, we first analyze the performance of our investigated transfer learning set-up and compare it with SOTA approaches afterwards. Afterwards, we analyze important fine-tuning aspects like the combination with other DG methods, layer freezing and different encoder complexities.

\subsection{Pre-Training Comparison}
\label{transfer_learning}
Our fine-tuning results for semantic segmentation training on a synthetic source datasets are shown in \cref{tab:m2f_transfer} where we compare the three major pre-training paradigms: ImageNet supervised pre-training, vision self-supervised pre-training, and vision-language pre-training.\\
First, we observe that the recent self-supervised vision pre-trained methods provide a stronger generalization than supervised pre-training since DEiT \cite{touvron2022deit} outperforms supervised pre-training by +5.7\% mIoU. Surprisingly, SAM which is pre-trained on a segmentation tasks performs slightly worse than DEiT but still outperforms ImageNet supervised training. However, SAM shows a weaker generalization than vision-language pre-training.\\ 
Notably, the CLIP \cite{radford2021learning} initialization performs even stronger on the DG mean than DEiT by +1.2\% mIoU, indicating that vision-language pre-training benefits from the larger pre-training datasets and the task-agnostic image-text pair training. When employing the EVA-CLIP \cite{EVA-CLIP} initialization, we observe a further increase of +8.2\% mIoU over CLIP. Similar effects can be observed when training on SYNTHIA as the source dataset.\\
Vision-language pre-training datasets contain a multitude of samples compared to self-supervised vision datasets, e.g. SAM with 11 million images has only a fraction of the 2 billion image-text pairs used for EVA-CLIP \cite{EVA-CLIP} pre-training. We consider this an inherent advantage of vision-language pre-training since collecting these datasets is easier, and the dataset sizes are significantly larger \cite{radford2021learning}. We attribute the clear differences between CLIP and EVA02-CLIP to three factors: architecture adjustments in EVA-02-CLIP, masked image modeling initialization and the larger size of data set. Possible ablations studies on this are left for future work.

\subsection{Domain Generalized Perception Benchmarks} 

In Table \ref{tab:sota_gta5}, we compare VLTSeg to other state-of-the-art methods for synthetic-to-real generalization. Our simple pre-training setting outperforms the best vision DG method HRDA \cite{hoyer2022hrda} by 7.3\% mIoU and most of the significantly more complex vision-language DG methods like CLOUDS \cite{benigmim2024collaborating} and performs competitive to the more complex approach Rein \cite{wei2024stronger} which needs an additional network module to reach this performance. This highlights the strong generalization
\begin{wraptable}{l}{0.5\columnwidth}
  \caption{\textbf{Domain generalization performance} of state-of-the-art approaches. Training was performed on the synthetic GTA5 ($\src\!=\!\gtavtrain$). Prior work results are cited from the respective paper.}
  \include{tables/sota_ext}
  \label{tab:sota_gta5} 
\end{wraptable}
 capabilities of vision-language pre-trained models without including any additional domain generalization methods. We also conducted experiments on SYNTHIA and with a ResNet-101 \cite{he2016deep} backbone (see supplement), which confirms these observations.\\
In addition, we compare our fine-tuning scheme using CLIP \cite{radford2021learning} for ResNet and EVA-CLIP \cite{EVA-CLIP} for ViT to state-of-the-art DG object detection methods following the Single-DGOD \cite{wu2022single} benchmark in Table \ref{tab:sdgod_objectdetection}. We include the two vision-language methods, CLIP The Gap \cite{vidit2023clip} and PODA \cite{fahes2023poda} and three methods with ImageNet initialization, namely S-DGOD \cite{wu2022single}, G-NAS \cite{wu2024g} and PDDOC \cite{li2024prompt}. Our fine-tuning performs similar in-domain and has a clear advantage over all approaches with ImageNet initialization for out-of-domain by e.g. 3.4\% mAP in average over G-NAS \cite{wu2024g} with a ResNet-101 encoder. Even though both vision-language methods use feature space augmentations, our fine-tuning method also achieves significant gains. Compared to CLIP The Gap \cite{vidit2023clip}, we outperform by up to 9.2 \% in-domain and out-of-domain with improvements up to 7.7 \% in mAP. Moreover, VLTDet performs competitive with PODA \cite{fahes2023poda} with a ResNet-101 backbone with an average of 36.9\% mAP compared to 37.1\% mAP for PODA. With the EVA02-CLIP backbone VLTDet reaches 39.1\% mAP and outperforms all other methods. On the challenging night rainy conditions it outperforms PODA \cite{fahes2023poda} as the second best approach by 4.5\% mAP. 
\begin{table}
    \centering
    \caption{\textbf{Comparison of VLTDet on real-to-real object detection benchmark $\mathrm{mAP}_{50}$ of S-DGOD\cite{wu2022single} }}
    
    \resizebox{0.9\columnwidth}{!}{
    \begin{tabular}{ccc|c|cccc|c}
    \toprule
      \textbf{Method} & Encoder & Init   & Day Clear  &Night Clear& Dusk Rainy & Night Rainy & Day Foggy & Average  \\ \hline 
      S-DGOD \cite{wu2022single}& R-101 & IN1k & 56.1 & 36.6&28.2 & 16.6 &33.5&28.7\\
      G-NAS \cite{wu2024g} & R-101 & IN1k & 58.4 & \textbf{45.0} & 35.1 & 17.4 & 36.4 & 33.5 \\
      PDDOC \cite{li2024prompt} & R-101 & IN1k & 53.6 & 38.5 & 33.7 & 19.2 & 39.1 & 32.6 \\
      CLIP The GAP \cite{vidit2023clip}& R-101 & CLIP  & 51.3 & 36.9  & 32.3 &18.7 &38.5&31.6\\
      PODA \cite{fahes2023poda} & R-101 & CLIP & - & 43.4 & 40.2 & 20.5 & \textbf{44.4}&37.1 \\
       
      \midrule
      \rowcolor{textcolor!20} VLTDet & R-101 & CLIP & \textbf{60.5} & 44.6 & 38.4 & 22.1 & 42.3 & 36.9 \\
      \rowcolor{textcolor!20} VLTDet & ViT-L-14 & EVA02-CLIP & 56.6 & 44.4 & \textbf{43.6} & \textbf{26.6} & 41.8 & \textbf{39.1} \\
\bottomrule
\end{tabular}}
\label{tab:sdgod_objectdetection}
\end{table}
To further analyze the effectiveness of vision-language pre-training we also benchmarked VLTSeg for in- and out-of-domain performance on real datasets as shown in Table \ref{tab:testset_performance}. Vision-language pre-training outperforms the previous domain adaptation(!) SOTA significantly by +8.36\% mIoU on the ACDC test set even though we are not accessing any ACDC samples during training different to CISS \cite{sakaridis2023condition}. It also outperforms all existing DG approaches like Rein \cite{wei2024stronger} which translates into first place on the official leaderboard. For the in-domain performance on $\cstest$, we obtain 86.4\% mIoU, marking best performance on the official leaderboard. Note that this performance is achieved without an extended training dataset \cite{borse2021inverseform, wang2023internimage, tao2020hierarchical} and less than half the parameters and only $40k$ training iterations compared to the previous SOTA InternImage \cite{wang2023internimage}. These results confirm the superior generalization capabilities of vision-language models in our simple fine-tuning approach compared to significantly more complex approaches. 
\begin{table}
  \centering
  \renewcommand{\arraystretch}{1.3}
  \caption{\textbf{Test set Performance of VLTSeg on Cityscapes and ACDC} compared with previous state-of-the-art methods. Detailed settings in the appendix.}
 \centering
    \resizebox{0.9\columnwidth}{!}{
        \begin{tabular}{ccccc||ccc}
        \toprule
            \multicolumn{5}{c||}{\textbf{Cityscapes} ($\cstrain$) $\rightarrow$ \textbf{Cityscapes} ($\cstest$)} & \multicolumn{3}{c}{\textbf{Cityscapes} ($\cstrain$) $\rightarrow$ \textbf{ACDC} ($\acdctest$)}\\
            \hline
            \textbf{Method} & \textbf{Rank}&\textbf{Params}&\textbf{Iter.} &\textbf{mIoU in \%} & \textbf{Method} &\textbf{UDA/DG}& \textbf{mIoU in \%}\\
            \hline
            ViT-Adapter-L \cite{chen2022vision}&5&571M&80k& 85.2 & HRDA \cite{hoyer2022hrda}& UDA & 67.96 \\
            InverseForm \cite{borse2021inverseform} &3&-&-& 85.6 & CISS \cite{sakaridis2023condition}& UDA & 69.55 \\ 
            \cline{6-8}
            HS3 \cite{borse2021hs3} &4&-&-& 85.8& PromptFormer \cite{gong2023prompting}& DG & 62.0 \\
            InternImage \cite{wang2023internimage}&2&1.2B&80k & 86.1& Rein \cite{wei2024stronger}&DG &  77.56  \\
            \cellcolor{textcolor!20}VLTSeg &\cellcolor{textcolor!20}1&\cellcolor{textcolor!20}304M &\cellcolor{textcolor!20}40k& \cellcolor{textcolor!20}\textbf{86.4} & \cellcolor{textcolor!20}VLTSeg ($1024^2$)&\cellcolor{textcolor!20}DG &\cellcolor{textcolor!20}\cellcolor{textcolor!20} \textbf{77.91}\\
            \bottomrule
    \end{tabular}
    }
    
  \label{tab:testset_performance} 
\end{table}

\subsection{Combination with existing DG methods}
\begin{table}
  \centering
  \renewcommand{\arraystretch}{0.8}
  \setlength{\tabcolsep}{.1em}
  \caption{\textbf{Vision-language pre-training in combination with different domain generalization methods.} Training was performed on the synthetic GTA5 dataset ($\src\!=\!\gtavtrain$) for segmentation and UrbanSyn ($\src\!=\!\mathcal{D}^\mathrm{US}_\mathrm{train}$) for object detection.}
  \scriptsize
  \include{tables/dg_ablation}
  \label{tab:dg_ablation} 
\end{table}
So far, most DG methods for segmentation utilize ImageNet pre-training. To analyze the effect of vision-language pre-training in combination with recent DG methods we tested several DG methods along with vision-language pre-training as shown in \cref{tab:dg_ablation}. For a fair comparison, we also evaluated those methods with an ImageNet pre-trained \network{ViT-L-16} encoder. The same evaluation is conducted for object detection to verify the applicability of the results to this task.\\
We can draw three major observations from this analysis. First, for both segmentation and object detection the vision-language pre-training shows a significantly better performance compared to ImageNet pre-training. Second, the additional gain by domain generalization methods for vision-language pre-training for segmentation is small, e.g. TLDR \cite{kim2023texture} adds 0.6\% mIoU. However, for ImageNet pre-training this gain is mostly higher with RandAugment \cite{Cubuk2022} providing 3.0\% mIoU improvement on the DG mean. Third, we notice that this differs for object detection where e.g. RandAugment also provides an improvement for vision-language pre-training. We reason that the vision-language pre-training already provides highly generalized representations so existing DG methods only provide a small gain.

\subsection{Real-to-real domain generalization}
\begin{table}
  \centering
  \renewcommand{\arraystretch}{0.8}
  \setlength{\tabcolsep}{.1em}
  \caption{\textbf{Real-to-real generalization} for segmentation and object detection. Training and evaluation conducted on different real-world datasets. The \colorbox{tu74}{gray} boxes show the oracle performance. rPD in \%.}
  \scriptsize
  \include{tables/r2r_ablation}
  \label{tab:r2r_ablation} 
\end{table}
For real-to-real domain generalization, we designed a cross-correlation experiment as shown in Table \ref{tab:r2r_ablation}. We observe that VLTSeg shows a significantly better domain generalization across all datasets, both in terms of absolute mIoU and our rPD metric. The gains are significant in comparison with the SegFormer architecture  \cite{Xie2022segformer}, which is commonly used in domain adaptation and generalization and therefore included as a baseline. When training on Cityscapes, VLTSeg improves the rPD by +8.3\% and by +14.7\% mIoU on ACDC. Similarly, when training on Mapillary, we improve significantly on ACDC by +8.5\% mIoU. \\
We fine-tuned a ViT-backbone initialized with SAM \cite{kirillov2023segment} as it is the largest segmentation vision pre-training dataset currently available. The generalization performance of SAM pre-trained weights is dataset-dependent but overall not significantly higher than for the ImageNet pre-trained SegFormer \cite{Xie2022segformer}. VLTSeg in contrast reaches a significantly better generalization than SAM, which shows that vision-language pre-training generalizes better than large-scale vision pre-training.\\
For real-to-real generalization for object detection we obtain similar results. For both training on Cityscapes and ACDC significant improvements of +7.3\% and +10.2\% rPD can be observed indicating a better generalization of vision-language pre-training. For training on BDD the improvement is smaller and ImageNet pre-training performs slightly better when evaluating on ACDC. This is most likely related to the much larger training dataset of 70k images (e.g. Cityscapes 2975 images) which causes a diminishing influence of the pre-training as previous work have already shown \cite{he2019rethinking, zoph2020rethinking}. Surprisingly, the SAM pre-training \cite{kirillov2023segment} performs different and achieves a better absolute performance on Cityscapes and BDD but on ACDC it performs signifcantly worse than VLTDet. Also the rPD of SAM is lower for all datasets.

\subsection{Analysis}
\begin{figure}
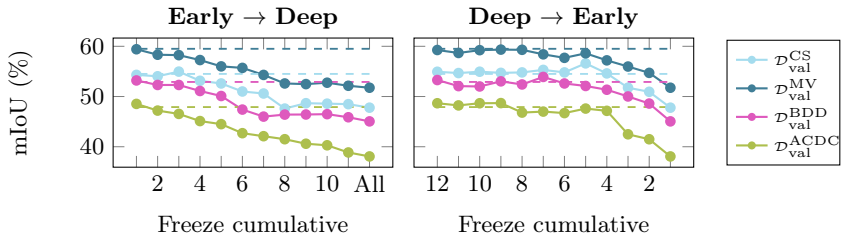

    \centering
    \include{tables/finetuning_ablation}
    \caption{\textbf{Ablation of freezing encoder layers} for the syn-to-real DG performance. Training on $\gtavtrain$) with \network{EVA-02-ViT-B-16}. Dashed lines indicate full fine-tuning.}
    \label{fig:frozen_layers}
\end{figure}
\begin{figure}
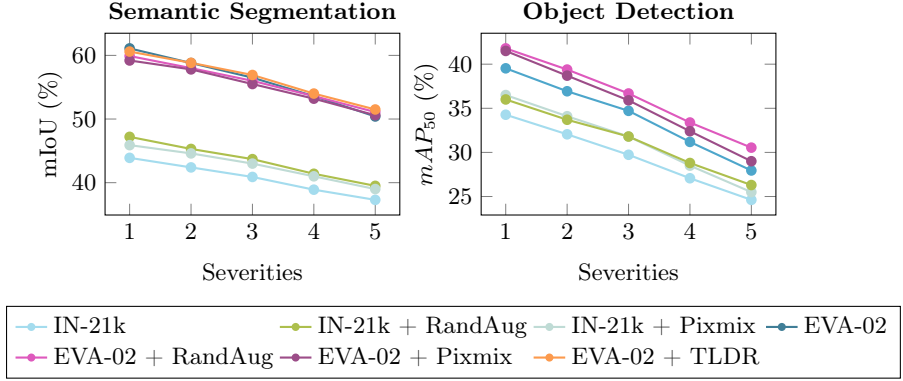

    \centering
    \include{tables/robustness_lineplot}
    \caption{\textbf{Robustness over increasing severities} of corruptions from \cite{michaelis2019benchmarking} for segmentation and object detection. Training on $\gtavtrain$ with \network{EVA-02-L-16} and \network{ViT-L-16}.}
    \label{fig:robustness}
\end{figure}

\textbf{Layer Freezing} We study the effect of layer freezing in \cref{fig:frozen_layers} to discover the best setting for fine-tuning the CLIP weights. We perform freezing cumulatively, starting either from the beginning or the end of the network. Fully fine-tuning results in the best performance while freezing the last layers performs competitively. Interestingly, only freezing the early layers degrades performance significantly indicating their importance for adapting to the perception task.\\
\textbf{Robustness Analysis} In \cref{fig:robustness}, we show the EVA-CLIP and ImageNet-trained models' robustness with and without augmentations against common natural corruptions. We employ the benchmark corruptions from  \cite{michaelis2019benchmarking} imitating real-world shifts like blurs, noise, or weather in 5 different severities. While EVA-CLIP is more robust, augmentations improve the robustness of object detection for both ImageNet and EVA-CLIP initialization, with RandAugment being the best policy. For semantic segmentation however, the robustness is only significantly enhanced with additional methods for ImageNet pre-training.\\ 
\textbf{Encoder Complexity}
We also researched the impact of the encoder complexity on the domain generalization performance as shown in \cref{tab:encoder_ablation_m2f}. 
\begin{wraptable}{r}{0.5\columnwidth}
\caption{\textbf{Domain generalization of different encoder initializations and complexities.} Training on GTA5. * denotes training over 20k iterations.}
\include{tables/encoder_ablation_m2f_decoder}
\label{tab:encoder_ablation_m2f}
\end{wraptable}
We observe that for CLIP and EVA02-CLIP the generalization gets better with an increasing encoder complexity increasing by 16.5\% mIoU from \network{EVA-02-S-16} to \network{EVA-02-L-14}. 
A larger complexity enables the encoder to better utilize the large vision-language pre-training datasets for generalized representations.\\
\textbf{Limitations} Our standard encoder is the \network{EVA-02-L-14} with 304M parameters which is 3.7$\times$ the parameters of the SegFormer-MiT-B5 encoder \cite{Xie2022segformer} which was used for previous state-of-the-art works. Therefore, our transfer learning requires more training resources in terms of GPU memory and time and is, similar to other DG approaches, not directly applicable to real-time settings.

\section{Conclusion}
\label{sec:conclusion}
We demonstrate that simple fine-tuning with vision-language representations for domain generalized segmentation and object detection performs competitively or superior to much more complex approaches. Hereby, we provide a strong but simple baseline, which we extensively evaluate on various DG benchmarks. We compare SOTA vision and vision-language pre-training strategies. Our results underpin the strong vision-language generalization and evolve the established strategy for domain generalization of initializing models with ImageNet pre-training to the utilization of vision-language transfer learning.

\bibliographystyle{splncs04}
\bibliography{main}
\include{supplement}
\end{document}

%% file: mathematical_formulations.tex
\newcommand{\src}[0]{{\mathcal{D}^\mathrm{S}}}
\newcommand{\tgt}[0]{{\mathcal{D}^\mathrm{T}}}

\newcommand{\network}[1]{{\texttt{#1}}}

\newcommand{\cstrain}[0]{{\mathcal{D}^\mathrm{CS}_\mathrm{train}}}

\newcommand{\cstest}[0]{{\mathcal{D}^\mathrm{CS}_\mathrm{test}}}

\newcommand{\csteststar}[0]{{\mathcal{D}^\mathrm{CS}_\mathrm{val}}}

\newcommand{\bddtrain}[0]{{\mathcal{D}^\mathrm{BDD}_\mathrm{train}}}

\newcommand{\bddteststar}[0]{{\mathcal{D}^\mathrm{BDD}_\mathrm{val}}}

\newcommand{\mvtrain}[0]{{\mathcal{D}^\mathrm{MV}_\mathrm{train}}}

\newcommand{\mvteststar}[0]{{\mathcal{D}^\mathrm{MV}_\mathrm{val}}}

\newcommand{\acdctrain}[0]{{\mathcal{D}^\mathrm{ACDC}_\mathrm{train}}}
\newcommand{\acdcval}[0]{{\mathcal{D}^\mathrm{ACDC}_\mathrm{val}}}
\newcommand{\acdctest}[0]{{\mathcal{D}^\mathrm{ACDC}_\mathrm{test}}}

\newcommand{\acdcteststar}[0]{{\mathcal{D}^\mathrm{ACDC}_\mathrm{val}}}

\newcommand{\gtavtrain}[0]{{\mathcal{D}^\mathrm{GTA5}_\mathrm{train}}}

\newcommand{\synthiatrain}[0]{{\mathcal{D}^\mathrm{SYN}_\mathrm{train}}}

%% file: ter-colors.tex
\definecolor{tu0}{rgb}{0.7451, 0.1176, 0.2353}

\definecolor{tu1}{rgb}{1.0000, 0.8039, 0.0000}
\definecolor{tu11}{rgb}{1.0000, 0.8627, 0.3020}
\definecolor{tu12}{rgb}{1.0000, 0.9020, 0.4980}
\definecolor{tu13}{rgb}{1.0000, 0.9412, 0.6980}
\definecolor{tu14}{rgb}{1.0000, 0.9608, 0.8000}

\definecolor{tu2}{rgb}{0.9804, 0.4314, 0.0000}
\definecolor{tu21}{rgb}{0.9882, 0.6039, 0.3020}
\definecolor{tu22}{rgb}{0.9882, 0.7137, 0.4980}
\definecolor{tu23}{rgb}{0.9922, 0.8275, 0.6980}
\definecolor{tu24}{rgb}{0.9961, 0.8863, 0.8000}

\definecolor{tu3}{rgb}{0.6902, 0.0000, 0.2745}
\definecolor{tu31}{rgb}{0.7529, 0.2000, 0.4196}
\definecolor{tu32}{rgb}{0.8431, 0.4980, 0.6353}
\definecolor{tu33}{rgb}{0.9216, 0.7490, 0.8196}
\definecolor{tu34}{rgb}{0.9529, 0.8510, 0.8902}

\definecolor{tu4}{rgb}{0.4863, 0.8039, 0.9020}
\definecolor{tu41}{rgb}{0.6431, 0.8627, 0.9333}
\definecolor{tu42}{rgb}{0.7412, 0.9020, 0.9490}
\definecolor{tu43}{rgb}{0.8431, 0.9412, 0.9686}
\definecolor{tu44}{rgb}{0.8980, 0.9608, 0.9804}

\definecolor{tu5}{rgb}{0.0000, 0.5020, 0.7059}
\definecolor{tu51}{rgb}{0.3020, 0.6510, 0.7961}
\definecolor{tu52}{rgb}{0.5490, 0.7765, 0.8667}
\definecolor{tu53}{rgb}{0.7490, 0.8745, 0.9255}
\definecolor{tu54}{rgb}{0.8510, 0.9255, 0.9569}

\definecolor{tu6}{rgb}{0.0000, 0.3255, 0.4549}
\definecolor{tu61}{rgb}{0.2510, 0.4941, 0.5922}
\definecolor{tu62}{rgb}{0.5490, 0.6941, 0.7529}
\definecolor{tu63}{rgb}{0.7490, 0.8314, 0.8627}
\definecolor{tu64}{rgb}{0.8510, 0.8980, 0.9176}

\definecolor{tu7}{rgb}{0.0314, 0.0314, 0.0314}
\definecolor{tu71}{rgb}{0.3725, 0.3725, 0.3725}
\definecolor{tu72}{rgb}{0.5882, 0.5882, 0.5882}
\definecolor{tu73}{rgb}{0.7529, 0.7529, 0.7529}
\definecolor{tu74}{rgb}{0.8667, 0.8667, 0.8667}

\definecolor{tu8}{rgb}{0.7765, 0.9333, 0.0000}
\definecolor{tu81}{rgb}{0.8431, 0.9529, 0.3020}
\definecolor{tu82}{rgb}{0.8863, 0.9647, 0.4980}
\definecolor{tu83}{rgb}{0.9333, 0.9804, 0.6980}
\definecolor{tu84}{rgb}{0.9569, 0.9882, 0.8000}

\definecolor{tu9}{rgb}{0.5373, 0.6431, 0.0000}
\definecolor{tu91}{rgb}{0.6784, 0.7490, 0.3020}
\definecolor{tu92}{rgb}{0.7686, 0.8196, 0.4980}
\definecolor{tu93}{rgb}{0.8588, 0.8941, 0.6980}
\definecolor{tu94}{rgb}{0.9059, 0.9294, 0.8000}

\definecolor{tu10}{rgb}{0.0000, 0.4431, 0.3373}
\definecolor{tu101}{rgb}{0.3020, 0.6118, 0.5373}
\definecolor{tu102}{rgb}{0.5490, 0.7490, 0.7020}
\definecolor{tu103}{rgb}{0.7490, 0.8588, 0.8353}
\definecolor{tu104}{rgb}{0.8549, 0.9176, 0.9059}

\definecolor{tu110}{rgb}{0.8000, 0.0000, 0.6000}
\definecolor{tu111}{rgb}{0.8706, 0.3490, 0.7412}
\definecolor{tu112}{rgb}{0.9216, 0.6000, 0.8392}
\definecolor{tu113}{rgb}{0.9608, 0.8000, 0.9216}
\definecolor{tu114}{rgb}{0.9804, 0.8980, 0.9608}

\definecolor{tu120}{rgb}{0.4627, 0.0000, 0.4627}
\definecolor{tu121}{rgb}{0.5961, 0.2510, 0.5961}
\definecolor{tu122}{rgb}{0.7294, 0.4980, 0.7294}
\definecolor{tu123}{rgb}{0.8392, 0.6980, 0.8392}
\definecolor{tu124}{rgb}{0.9216, 0.8510, 0.9216}

\definecolor{tu130}{rgb}{0.4627, 0.0000, 0.3294}
\definecolor{tu131}{rgb}{0.6118, 0.3020, 0.5333}
\definecolor{tu132}{rgb}{0.7569, 0.5490, 0.6980}
\definecolor{tu133}{rgb}{0.8667, 0.7490, 0.8314}
\definecolor{tu134}{rgb}{0.9216, 0.8510, 0.9020}

\definecolor{con_blue}{HTML}{004488}
\definecolor{con_yellow}{HTML}{DDAA33}
\definecolor{con_red}{HTML}{BB5566}
\definecolor{con_black}{HTML}{000000}
\definecolor{con_gray}{HTML}{AAAAAA}

\definecolor{vib_orange}{HTML}{EE7733}
\definecolor{vib_darkblue}{HTML}{0077BB}
\definecolor{vib_lightblue}{HTML}{33BBEE}
\definecolor{vib_magenta}{HTML}{EE3377}
\definecolor{vib_red}{HTML}{CC3311}
\definecolor{vib_green}{HTML}{009988}
\definecolor{vib_gray}{HTML}{BBBBBB}
\definecolor{vib_black}{HTML}{000000}

%% file: mathmacros.tex
\newcommand{\VEC}[1]{\mathbf{#1}}          %

\newcommand{\putindex}[3]{\vtop{\hbox{\hspace{#3} $#1$}
            \hbox{\raise 6mm \hbox{$\scriptscriptstyle #2$}}}}

\newcommand{\gradx}[0]{\vtop{\hbox{\rm grad}
            \hbox{\raise 2.5mm \hbox{\rm \hspace{2mm} \footnotesize x}}}}

\newcommand{\grady}[0]{\vtop{\hbox{\rm grad}
            \hbox{\raise 2.5mm \hbox{\rm \hspace{2mm} \footnotesize y}}}}

\newcommand{\grad}[1]{\vtop{\hbox{\rm grad}
            \hbox{\raise 2.5mm \hbox{#1}}}}

\newcommand{\btb}{     \begin{tabbing}             }
\newcommand{\bte}{     \end{tabbing}               }

%% file: tables/m2f_transfer_learning.tex
\extrarowheight=\aboverulesep
    \addtolength{\extrarowheight}{\belowrulesep}
    \aboverulesep=0pt
    \belowrulesep=0pt
    \resizebox{0.85\columnwidth}{!}{%
    \begin{tabular}{ccccc|ccccc}
    \toprule
    & \multicolumn{4}{c|}{\textbf{Pre-Training}}& \multicolumn{5}{c}{\textbf{mIoU in \%}} \\
    \cline{6-10}
    &Method&Data&Sup.&Self-Sup.& $\csteststar$ & $\bddteststar$&$\mvteststar$ &$\acdcteststar$ &\raisebox{-4.0pt}{\shortstack{\textbf{DG} \\\textbf{mean}}}  \rule[-1.2ex]{0mm}{3.65ex}\\
    \toprule
     
    \parbox[t]{4mm}{\multirow{8}{*}{\rotatebox[origin=c]{90}{\small{{$\src$: \textbf{GTA5}}}}}} &SegFormer\cite{Xie2022segformer}&ImgNet-1K&\checkmark&\ding{55} &46.6 &45.6&50.1&36.4&44.7\\
    
    &Supervised*&ImgNet-21K&\checkmark&\ding{55}& 49.3 & 47.0 & 52.2& 43.5 & 48.0\\
    
      &MoCov3* \cite{chenempirical}&ImgNet-1K&\ding{55}&\checkmark& 49.7 & 46.2 & 52.4 & 39.1 & 46.9\\
    
    &DeiT3* \cite{touvron2022deit}&ImgNet-21K&\checkmark &\checkmark & 53.7 & 52.6 & 59.3 & 49.0 & 53.7\\
    &SAM* \cite{kirillov2023segment} & SA-1B &\checkmark&\ding{55}& 53.2  & 50.3 &  58.8 &  45.5 & 52.0 \\
    \cline{2-10}
    &LDM \cite{rombach2021highresolution}& LAION-5B&\checkmark&\ding{55}& 49.2 & - & - & - & - \\
   & CLIP \faLock $^\circ$ \cite{radford2021learning} \cellcolor{textcolor!20}   &WIT \cellcolor{textcolor!20}   &\checkmark \cellcolor{textcolor!20}   & \ding{55} \cellcolor{textcolor!20}  & 53.2 \cellcolor{textcolor!20}   & 49.8 \cellcolor{textcolor!20}  & 57.1\cellcolor{textcolor!20}   & 45.0 \cellcolor{textcolor!20}  & 51.2 \cellcolor{textcolor!20}   \\   
   &CLIP  $^\circ$ \cite{radford2021learning} \cellcolor{textcolor!20}   &WIT \cellcolor{textcolor!20}   &\checkmark \cellcolor{textcolor!20}   &\ding{55}    \cellcolor{textcolor!20}  & 55.6 \cellcolor{textcolor!20}  & 52.5 \cellcolor{textcolor!20}  & 59.9 \cellcolor{textcolor!20}  &51.5 \cellcolor{textcolor!20}   & 54.9 \cellcolor{textcolor!20}  \\
   &EVA-02-CLIP \faLock $^\circ$ \cite{EVA-CLIP} \cellcolor{textcolor!20}   &Merged-2B \cellcolor{textcolor!20}   &\checkmark \cellcolor{textcolor!20}   &\ding{55}  \cellcolor{textcolor!20}  & 55.2  \cellcolor{textcolor!20}  & 51.3  \cellcolor{textcolor!20}  & 57.4 \cellcolor{textcolor!20}   & 47.3  \cellcolor{textcolor!20}  & 52.8 \cellcolor{textcolor!20}    \\
   &EVA-02-CLIP$^\circ$ \cite{EVA-CLIP} \cellcolor{textcolor!20}  &Merged-2B \cellcolor{textcolor!20}   &\checkmark \cellcolor{textcolor!20}  &\ding{55} \cellcolor{textcolor!20}  &  \textbf{65.3}  \cellcolor{textcolor!20}  & \textbf{58.3} \cellcolor{textcolor!20}  & \textbf{66.0} \cellcolor{textcolor!20}  & \textbf{62.6} \cellcolor{textcolor!20}  &  \textbf{63.1} \cellcolor{textcolor!20}  \\
    \hline
    \parbox[t]{4mm}{\multirow{7}{*}{\rotatebox[origin=c]{90}{\small{{$\src$: \textbf{SYNTHIA}}}}}} &SegFormer\cite{Xie2022segformer}&ImgNet-1K&\checkmark&\ding{55}&41.1 &36.2&42.4&32.6&38.1\\
     &Supervised*&ImgNet-21K&\checkmark&\ding{55}& 44.3 & 37.1  & 43.1 & 34.8 & 39.8\\
      &MoCov3*\cite{chenempirical}&ImgNet-1K&\ding{55}&\checkmark & 40.2 & 35.4 & 41.5 & 31.7 & 37.2 \\
     &DeiT3*\cite{touvron2022deit}&ImgNet-21K&\checkmark &\checkmark & 47.8 & 39.1 & 45.4 & 34.7 & 41.8 \\
     &SAM* \cite{kirillov2023segment} & SA-1B &\checkmark&\ding{55}& 51.6 & 40.4 &  50.1 &  40.1 & 45.6 \\
     \cline{2-10}
    & CLIP \faLock   \cellcolor{textcolor!20}$^\circ$ \cite{radford2021learning}   \cellcolor{textcolor!20}& WIT   \cellcolor{textcolor!20}& \checkmark   \cellcolor{textcolor!20}& \ding{55}  \cellcolor{textcolor!20} & 46.1   \cellcolor{textcolor!20}& 41.8  \cellcolor{textcolor!20} & 45.8  \cellcolor{textcolor!20} & 35.1  \cellcolor{textcolor!20} & 42.2   \cellcolor{textcolor!20}\\
    & CLIP $^\circ$ \cite{radford2021learning} \cellcolor{textcolor!20}&WIT \cellcolor{textcolor!20}&\checkmark \cellcolor{textcolor!20}&\ding{55} \cellcolor{textcolor!20}& 51.1 \cellcolor{textcolor!20} & 44.7 \cellcolor{textcolor!20}& 50.6 \cellcolor{textcolor!20} & 40.7 \cellcolor{textcolor!20}& 46.8\cellcolor{textcolor!20} \\
    &EVA-02-CLIP \faLock  $^\circ$  \cite{EVA-CLIP}  \cellcolor{textcolor!20} &Merged-2B  \cellcolor{textcolor!20}&\checkmark  \cellcolor{textcolor!20}& \ding{55}  \cellcolor{textcolor!20}& 48.3  \cellcolor{textcolor!20}  & 42.6  \cellcolor{textcolor!20}& 46.4  \cellcolor{textcolor!20}& 37.1  \cellcolor{textcolor!20} & 43.6  \cellcolor{textcolor!20}\\
    &EVA-02-CLIP $^\circ$ \cite{EVA-CLIP}  \cellcolor{textcolor!20}&Merged-2B  \cellcolor{textcolor!20}&\checkmark  \cellcolor{textcolor!20} &\ding{55}  \cellcolor{textcolor!20} & \textbf{56.8}  \cellcolor{textcolor!20}& \textbf{51.9}  \cellcolor{textcolor!20}& \textbf{55.1}  \cellcolor{textcolor!20}& \textbf{48.5} \cellcolor{textcolor!20} & \textbf{53.1}  \cellcolor{textcolor!20}\\
    \bottomrule
    \end{tabular}}

%% file: tables/sota_ext.tex
\extrarowheight=\aboverulesep
    \addtolength{\extrarowheight}{\belowrulesep}
    \aboverulesep=0pt
    \belowrulesep=0pt
    \setlength\tabcolsep{0pt}
    \resizebox{0.5\columnwidth}{!}{
    \begin{tabular}{@{}ccccccc@{}}
        \toprule
        \multirow{1}{*}{\rot{\makecell[c]{\textbf{}}}} & \textbf{DG Method} & \raisebox{-4.0pt}{\shortstack{\textbf{Enc.} \\\textbf{Params}}}  \rule[-1.2ex]{0mm}{3.65ex}&\multicolumn{4}{c}{\textbf{mIoU (\%) on}} \\
        \cmidrule{4-7}
        &&& $\csteststar$ & $\bddteststar$ & $\mvteststar$ & \raisebox{-4.0pt}{\shortstack{\textbf{DG} \\\textbf{mean}}}  \rule[-1.2ex]{0mm}{3.65ex} \\

        \midrule

         \cellcolor{tu22} &Baseline \cite{Xie2022segformer}& 81.4M& 46.6&45.6& 50.1   &47.4\\
         \cellcolor{tu22}&ReVT \cite{Termoehlen2023arxiv}& 81.4M  & 50.0&48.0& 52.8   &50.3\\
         \cellcolor{tu22}&SHADE \cite{zhao2022style}    &   81.4M         & 53.3& 48.2& 55.0&52.2\\
         \cellcolor{tu22}&IBAFormer \cite{sun2023ibaformer}& 81.4M & 56.3& 49.8& 58.3&54.8 \\
        \cellcolor{tu22}&CMFormer \cite{bi2023learning}& 197M  & 55.3& 49.9& 60.1&55.1\\
        {\multirow{-5}{*}{\rotatebox[origin=c]{90}{\cellcolor{tu22}\footnotesize \textbf{Vision DG}}}}&HRDA \cite{hoyer2023domain}  &  81.4M             & 57.4& 49.1& 61.2&55.9\\
        \cline{1-7}
       \cellcolor{tu12}&DGinStyle \cite{jia2023dginstyle}&81.4M & 58.6 & 52.3 & 62.5 & 57.8 \\
        \cellcolor{tu12}&DIDEX \cite{niemeijer2024generalization}&81.4M & 62.0 & 54.3 & 62.0 & 59.7 \\
         \cellcolor{tu12}&PromptFormer \cite{gong2023prompting}&-  & 52.0& -& -&- \\
        \cellcolor{tu12}& CLOUDS \cite{benigmim2024collaborating}&198M & 60.2 & 57.4 & \textbf{67.0} & 61.5 \\
        \cellcolor{tu12}&Rein \cite{wei2024stronger}&307M & \textbf{65.3} & \textbf{60.5} & 64.9 & \textbf{63.6} \\
        
         {\multirow{-6}{*}{\rotatebox[origin=c]{90}{\cellcolor{tu12}\footnotesize \textbf{Vision-Language DG}}}}& VLTSeg  \cellcolor{textcolor!20}& 304M  \cellcolor{textcolor!20} & \textbf{65.3}  \cellcolor{textcolor!20}& 58.3 \cellcolor{textcolor!20} & 66.0 \cellcolor{textcolor!20} & 63.2 \cellcolor{textcolor!20}\\

        \bottomrule
    \end{tabular}}

%% file: tables/dg_ablation.tex
\extrarowheight=\aboverulesep
    \addtolength{\extrarowheight}{\belowrulesep}
    \aboverulesep=0pt
    \belowrulesep=0pt
    \begin{tabular}{@{}lcccccc|ccccc@{}}
        \toprule
        &&\multicolumn{5}{c|}{\textbf{Segmentation}}&\multicolumn{5}{c}{\textbf{Object Detection}}\\
         \textbf{DG Method}&\textbf{Init} &\multicolumn{4}{c}{\textbf{mIoU (\%) on}}&&\multicolumn{4}{c}{\textbf{$\textrm{mAP}_{50}$ (\%) on}}& \\
        \cmidrule{3-12}
        && $\csteststar$ & $\bddteststar$ & $\mvteststar$ & \raisebox{-4.0pt}{\shortstack{\textbf{DG} \\\textbf{mean}}}  \rule[-1.2ex]{0mm}{3.65ex}&$\Delta$&$\csteststar$ & $\bddteststar$ & $\acdcteststar$ & \raisebox{-4.0pt}{\shortstack{\textbf{DG} \\\textbf{mean}}}  \rule[-1.2ex]{0mm}{3.65ex}&$\Delta$ \\
        \midrule

        \multirow{2}{*}{Baseline}&IN-21k&46.4&47.2&52.1&48.6&- & 38.0 & 22.4 & 21.9 & 27.4&-\\
        & EVA-02-CLIP \cellcolor{Gray!20}&  65.3 \cellcolor{Gray!20}&58.3 \cellcolor{Gray!20}&66.0 \cellcolor{Gray!20}& 63.2 \cellcolor{Gray!20}&- \cellcolor{Gray!20}&43.1 \cellcolor{Gray!20}& 27.1 \cellcolor{Gray!20}& 29.0 \cellcolor{Gray!20} & 33.1 \cellcolor{Gray!20}&- \cellcolor{Gray!20}\\
        \hline
        \multirow{2}{*}{RandAugment \cite{Cubuk2022}}&IN-21k & 50.9&49.2& 54.6   &51.6& \textcolor{Green}{+3.0}& 40.8 & 22.6 & 21.6 & 28.3 &\textcolor{Green}{+0.9} \\
        &EVA-02-CLIP \cellcolor{Gray!20} &  63.5 \cellcolor{Gray!20}&58.0 \cellcolor{Gray!20} &66.0 \cellcolor{Gray!20} & \cellcolor{Gray!20} 62.5 &\textcolor{Red}{-0.7} \cellcolor{Gray!20}& 47.3 \cellcolor{Gray!20} & 29.4 \cellcolor{Gray!20} & 30.6 \cellcolor{Gray!20}& \textbf{35.8} \cellcolor{Gray!20}&\textcolor{Green}{+2.7 \cellcolor{Gray!20}}\\
        \hline
       
        \multirow{2}{*}{PixMix \cite{Hendrycks2022pixmix}} &IN-21k & 49.3&46.4& 54.0  &49.9&\textcolor{Green}{+1.3}& 41.8 & 22.8 & 23.0 & 29.2&\textcolor{Green}{+1.8} \\
        & EVA-02-CLIP  \cellcolor{Gray!20}& 63.1 \cellcolor{Gray!20}& 57.7 \cellcolor{Gray!20}& 63.8 \cellcolor{Gray!20}&61.5 \cellcolor{Gray!20} &\textcolor{Red}{-1.7} \cellcolor{Gray!20}& 46.7 \cellcolor{Gray!20}&  28.0 \cellcolor{Gray!20}& 29.9 \cellcolor{Gray!20} & 34.9 \cellcolor{Gray!20} & \textcolor{Green}{+1.8} \cellcolor{Gray!20}\\
        \hline
        \multirow{2}{*}{ReVT\cite{Termoehlen2023arxiv}} &IN-21k  & 48.2&46.6& 51.9   &48.9&\textcolor{Green}{+0.3}&39.3&20.7&21.5&27.2&\textcolor{Red}{-0.2}\\
        & \cellcolor{Gray!20} EVA-02-CLIP&63.7 \cellcolor{Gray!20} & 60.7 \cellcolor{Gray!20} & 66.6 \cellcolor{Gray!20} &63.5 \cellcolor{Gray!20}&\textcolor{Green}{+0.3} \cellcolor{Gray!20}&42.8 \cellcolor{Gray!20}&23.5 \cellcolor{Gray!20}&26.2 \cellcolor{Gray!20}&30.8 \cellcolor{Gray!20}&\textcolor{Red}{-2.3} \cellcolor{Gray!20}\\
        \hline
        \multirow{2}{*}{PASTA \cite{2023iccv_PASTA}} &IN-21k  & 49.2&49.4&  53.9  &50.8
&\textcolor{Green}{+2.2}& 38.8 & 22.4 & 22.9 & 28.0& \textcolor{Green}{+0.6} \\
        
        & \cellcolor{Gray!20} EVA-02-CLIP&  \cellcolor{Gray!20} 62.9   \cellcolor{Gray!20} &58.4  \cellcolor{Gray!20}& 65.8  \cellcolor{Gray!20} &62.4  \cellcolor{Gray!20}&\textcolor{Red}{-0.8}  \cellcolor{Gray!20}& 43.8  \cellcolor{Gray!20}& 27.4  \cellcolor{Gray!20}& 29.2 \cellcolor{Gray!20} & 33.5 \cellcolor{Gray!20} &\textcolor{Green}{+0.4}  \cellcolor{Gray!20}\\
        \hline
         \multirow{2}{*}{TLDR \cite{kim2023texture}} &IN-21k  & 49.2&46.9&52.2   &49.4 &\textcolor{Green}{+0.8} & -& - & - & -&-\\
        &\cellcolor{Gray!20} EVA-02-CLIP&  \cellcolor{Gray!20} 64.5 \cellcolor{Gray!20}&61.0 \cellcolor{Gray!20}& 66.0 \cellcolor{Gray!20}  &\textbf{63.8}\cellcolor{Gray!20}& \textcolor{Green}{+0.6} \cellcolor{Gray!20}& -\cellcolor{Gray!20}& - \cellcolor{Gray!20}& - \cellcolor{Gray!20}& -\cellcolor{Gray!20}&-\cellcolor{Gray!20}\\

        \bottomrule
    \end{tabular}

%% file: tables/r2r_ablation.tex
\extrarowheight=\aboverulesep
    \addtolength{\extrarowheight}{\belowrulesep}
    \aboverulesep=0pt
    \belowrulesep=0pt
    \begin{tabular}{@{}lcccccc|ccccc@{}}
        \toprule
        &&\multicolumn{5}{c|}{\textbf{Segmentation}}&\multicolumn{5}{c}{\textbf{Object Detection}}\\
         \textbf{Source}&\textbf{Method} &\multicolumn{4}{c}{\textbf{mIoU (\%) on}}& \textbf{rPD}&\textbf{Method}&\multicolumn{3}{c}{\textbf{$\textrm{mAP}_{50}$ (\%) on}}& \textbf{rPD}  \\
        \cmidrule{3-6}
        \cmidrule{9-11}
        && $\csteststar$ & $\bddteststar$ & $\mvteststar$ & $\acdcteststar$& & &$\csteststar$ & $\bddteststar$ & $\acdcteststar$ & \\
        \midrule

        \multirow{4}{*}{$\cstrain$}&SegFormer \cite{Xie2022segformer}& \cellcolor{tu74} 80.8 & 58.0 & 69.1 & 58.6& 76.6 & IN21k &\cellcolor{tu74} 52.8 & 30.8 & 35.6& 62.9\\
        &SAM\cite{kirillov2023segment}+M2F&\cellcolor{tu74} 82.8 & 57.1 & 71.0 & 60.5 & 75.9&SAM \cite{kirillov2023segment}&\cellcolor{tu74}\textbf{61.2}&37.2 &\textbf{45.1} &67.2\\
        &  \cellcolor{textcolor!20}VLTSeg&\cellcolor{tu74} \textbf{85.0} &  \cellcolor{textcolor!20}\textbf{66.0} &  \cellcolor{textcolor!20}\textbf{77.2} &  \cellcolor{textcolor!20}\textbf{73.3} & \textbf{84.9}  \cellcolor{textcolor!20}& VLTDet \cellcolor{textcolor!20}&\cellcolor{tu74} 59.7 & \cellcolor{textcolor!20} \textbf{39.6} &  \cellcolor{textcolor!20}44.3 & \cellcolor{textcolor!20} \textbf{70.2} \\
        \hline
        \multirow{4}{*}{$\bddtrain$}&SegFormer \cite{Xie2022segformer} &62.4 &\cellcolor{tu74} 64.4 & 63.0& 54.2 &93.0 &IN21k  & 45.2  & \cellcolor{tu74} 51.2 & 35.7 & 79.0\\
        &SAM\cite{kirillov2023segment}+M2F& 68.8 & \cellcolor{tu74}69.3 & 69.1 & 57.9 & 94.2& SAM\cite{kirillov2023segment} & \textbf{47.5} &\cellcolor{tu74}\textbf{53.8}&\textbf{37.1}&78.6\\
        &\cellcolor{textcolor!20} VLTSeg& \textbf{78.9}  \cellcolor{textcolor!20}&\cellcolor{tu74} \textbf{72.5}  & \textbf{75.9}  \cellcolor{textcolor!20}& \textbf{71.3} \cellcolor{textcolor!20} & \textbf{104.0}  \cellcolor{textcolor!20}&VLTDet  \cellcolor{textcolor!20}& 46.6  \cellcolor{textcolor!20}&\cellcolor{tu74} 49.2  & 34.5  \cellcolor{textcolor!20}& \textbf{82.4} \cellcolor{textcolor!20}\\
        \hline
        \multirow{4}{*}{$\acdctrain$}&SegFormer \cite{Xie2022segformer}& 66.3 & 54.1 & 64.9 &\cellcolor{tu74} 75.0 & 82.4 &IN21k & 22.4 & 22.5 &\cellcolor{tu74} 32.8 & 68.4\\
        &SAM\cite{kirillov2023segment}+M2F& 68.9 & 57.0 & 68.1 & \cellcolor{tu74} 77.4 & 83.5 & SAM \cite{kirillov2023segment}& 33.9&25.3&47.8&61.9 \\
        & \cellcolor{textcolor!20}VLTSeg& \textbf{78.8}  \cellcolor{textcolor!20} & \textbf{65.0}  \cellcolor{textcolor!20}& \textbf{74.1}  \cellcolor{textcolor!20}&\cellcolor{tu74} \textbf{80.0} & \textbf{90.8 \cellcolor{textcolor!20}}&VLTDet  \cellcolor{textcolor!20}& \textbf{44.9} \cellcolor{textcolor!20} & \textbf{31.0}  \cellcolor{textcolor!20}& \cellcolor{tu74} \textbf{48.3} & \textbf{78.6} \cellcolor{textcolor!20}\\
        \hline
        \multirow{4}{*}{$\mvtrain$}& SegFormer \cite{Xie2022segformer}  & 76.1 & 63.0 &\cellcolor{tu74} 78.9 & 65.7& 86.5 &-  &- &-  &- &-\\
         &SAM\cite{kirillov2023segment}+M2F& 76.4 & 62.8 &\cellcolor{tu74} 78.9 & 63.4 & 85.6&-&-&- &- &-\\
        & \cellcolor{textcolor!20} VLTSeg& \textbf{81.6}  \cellcolor{textcolor!20}& \textbf{69.2}  \cellcolor{textcolor!20}&\cellcolor{tu74} \textbf{83.8} & \textbf{74.2} \cellcolor{textcolor!20} & \textbf{89.5} \cellcolor{textcolor!20}&-  \cellcolor{textcolor!20}&-  \cellcolor{textcolor!20} &-   \cellcolor{textcolor!20}&-  \cellcolor{textcolor!20}& - \cellcolor{textcolor!20}\\
        \hline

        \bottomrule
    \end{tabular}

%% file: tables/finetuning_ablation.tex
\begin{tikzpicture}
\begin{axis}[
    width=53mm,
height=33mm,
at={(0.0cm,1.1cm)},
       ylabel={mIoU (\%)},
        xlabel={ \footnotesize Freeze cumulative},
        title = {\footnotesize \textbf{Early} $\rightarrow$ \textbf{Deep}}, 
        title style={yshift=-1.5ex},
        xtick={1,2,3,4,5,6,7,8,9,10,11,12},
        xticklabels={,2,,4,,6,,8,,10,,All},
        label style={font=\footnotesize},
        tick label style={font=\footnotesize}, 
        legend style={anchor=east, at={(2.6,0.5)}, text width=0.7cm, legend image post style={scale=0.7}},
        legend cell align=left,
        legend columns=1
    ]

    \addplot[line width=0.8pt,color=tu41, mark=*, mark options={scale=0.8}] coordinates {
        (1, 54.31)
        (2, 54.03) 
        (3, 54.96) 
        (4, 53.13)
        (5, 52.6)
        (6, 51.0)
        (7, 50.6)
        (8, 47.6) 
        (9, 48.67) 
        (10, 48.58) 
        (11, 48.48)
        (12, 47.78)
    };
    \addlegendentry{\tiny{$\csteststar$}}
    
    \addplot[line width=0.8pt,color=tu61, mark=*, mark options={scale=0.8}] coordinates {
        (1, 59.41)
        (2, 58.32) 
        (3, 58.24) 
        (4, 57.26)
        (5, 56.0)
        (6, 55.7)
        (7, 54.3)
        (8, 52.6) 
        (9, 52.47) 
        (10, 52.75) 
        (11, 52.14)
        (12, 51.75)
    };
    \addlegendentry{\tiny{$\mvteststar$}}
    
    \addplot[line width=0.8pt,color=tu111, mark=*, mark options={scale=0.8}] coordinates {
        (1, 53.2)
        (2, 52.3) 
        (3, 52.3) 
        (4, 51.1)
        (5, 50.1)
        (6, 47.4)
        (7, 46.0)
        (8, 46.4) 
        (9, 46.42) 
        (10, 46.49) 
        (11, 45.87)
        (12, 45.05)
    };
    \addlegendentry{\tiny{$\bddteststar$}}
    
    \addplot[line width=0.8pt,color=tu91, mark=*, mark options={scale=0.8}] coordinates {
        (1, 48.53)
        (2, 47.2) 
        (3, 46.54) 
        (4, 45.11)
        (5, 44.5)
        (6, 42.7)
        (7, 42.1)
        (8, 41.5) 
        (9, 40.6) 
        (10, 40.29) 
        (11, 38.86)
        (12, 38.07)
    };
    \addlegendentry{\tiny{$\acdcteststar$}}
    \addplot[tu91, dashed, line width=0.7pt] coordinates {(12,47.9) (1,47.9)};
    \addplot[tu41, dashed, line width=0.7pt] coordinates {(12,54.5) (1,54.5)};
    \addplot[tu61, dashed, line width=0.7pt] coordinates {(12,59.5) (1,59.5)};
    \addplot[tu111, dashed, line width=0.7pt] coordinates {(12,52.9) (1,52.9)};

    \end{axis}

     \begin{axis}[
    width=53mm,
height=33mm,
at={(4.0cm,1.1cm)},
        xlabel={\footnotesize Freeze cumulative},
        yticklabel={\empty}, 
        title={\footnotesize \textbf{Deep} $\rightarrow$ \textbf{Early}},
        title style={yshift=-1.5ex},
        xtick={12,11,10,9,8,7,6,5,4,3,2,1},
        xticklabels={12,,10,,8,,6,,4,,2,},
        label style={font=\footnotesize},
        tick label style={font=\footnotesize}, 
        legend style={at={(-0.6,-1.2)}, anchor=south west},
        legend cell align=left,
        legend columns=2, 
        x dir=reverse
    ]
    \addplot[line width=0.8pt,color=tu41, mark=*, mark options={scale=0.8}] coordinates {
        (1, 47.78)
        (2, 50.94) 
        (3, 51.75) 
        (4, 54.6)
        (5, 56.6)
        (6, 54.8)
        (7, 55.3)
        (8, 54.8) 
        (9, 54.71) 
        (10, 54.94) 
        (11, 54.67)
        (12, 54.93)
    };
    \addplot[tu41, dashed, line width=0.7pt] coordinates {(12,54.5) (1,54.5)};
    \addplot[line width=0.8pt,color=tu61, mark=*, mark options={scale=0.8}] coordinates {
        (1, 51.75)
        (2, 54.71) 
        (3, 55.97) 
        (4, 57.2)
        (5, 58.6)
        (6, 57.7)
        (7, 58.4)
        (8, 59.3) 
        (9, 59.34) 
        (10, 59.23) 
        (11, 58.67)
        (12, 59.26)
    };
    \addplot[tu61, dashed, line width=0.7pt] coordinates {(12,59.5) (1,59.5)};
    \addplot[line width=0.8pt,color=tu111, mark=*, mark options={scale=0.8}] coordinates {
        (1, 45.05)
        (2, 48.6) 
        (3, 50.01) 
        (4, 51.35)
        (5, 52.1)
        (6, 52.6)
        (7, 53.9)
        (8, 52.4) 
        (9, 53.02) 
        (10, 52.0) 
        (11, 52.07)
        (12, 53.28)
    };
    \addplot[tu111, dashed, line width=0.7pt] coordinates {(12,52.9) (1,52.9)};
    \addplot[line width=0.8pt,color=tu91, mark=*, mark options={scale=0.8}] coordinates {
       (1, 38.07)
        (2, 41.49) 
        (3, 42.48) 
        (4, 47.15)
        (5, 47.6)
        (6, 46.7)
        (7, 47.0)
        (8, 46.8) 
        (9, 48.69) 
        (10, 48.66) 
        (11, 48.23)
        (12, 48.66)
    };
    \addplot[tu91, dashed, line width=0.7pt] coordinates {(12,47.9) (1,47.9)};
    
    \end{axis}

\end{tikzpicture}

%% file: tables/robustness_lineplot.tex
\begin{tikzpicture}
\begin{axis}[
    width=55mm,
height=40mm,
at={(0.0cm,1.1cm)},
       ylabel={mIoU (\%)},
        xlabel={ \footnotesize Severities},
        title = {\footnotesize \textbf{Semantic Segmentation}}, 
        title style={yshift=-1.5ex},
        xtick={1,2,3,4,5},
        xticklabels={1,2,3,4,5},
        label style={font=\footnotesize},
        tick label style={font=\footnotesize}, 
        legend style={anchor=east, at={(2.7,-0.7)}, legend image post style={scale=0.7}},
        y label style = {yshift=-0.5cm}, 
        legend cell align=left,
        legend columns=4, 
    ]

    \addplot[line width=0.8pt,color=tu41, mark=*, mark options={scale=0.8}] coordinates {
        (1, 43.9)
        (2, 42.4) 
        (3, 40.9) 
        (4, 38.9)
        (5, 37.3)
    };
    \addlegendentry{\footnotesize IN-21k}

    \addplot[line width=0.8pt,color=tu91, mark=*, mark options={scale=0.8}] coordinates {
        (1, 47.2)
        (2, 45.3) 
        (3, 43.7) 
        (4, 41.4)
        (5, 39.5)
    };
    \addlegendentry{\footnotesize IN-21k + RandAug}

     \addplot[line width=0.8pt,color=tu103, mark=*, mark options={scale=0.8}] coordinates {
        (1, 45.9)
        (2, 44.6) 
        (3, 43.0) 
        (4, 41.0)
        (5, 39.0)
    };
    \addlegendentry{\footnotesize IN-21k + Pixmix}

    \addplot[line width=0.8pt,color=tu61, mark=*, mark options={scale=0.8}] coordinates {
        (1, 61.1)
        (2, 58.8) 
        (3, 56.5) 
        (4, 53.6)
        (5, 50.4)
    };
    \addlegendentry{\footnotesize EVA-02}

      \addplot[line width=0.8pt,color=tu111, mark=*, mark options={scale=0.8}] coordinates {
        (1, 59.9)
        (2, 58.0) 
        (3, 56.0) 
        (4, 53.6)
        (5, 51.1)
    };
    \addlegendentry{\footnotesize EVA-02 + RandAug}

      \addplot[line width=0.8pt,color=tu131, mark=*, mark options={scale=0.8}] coordinates {
        (1, 59.2)
        (2, 57.8) 
        (3, 55.5) 
        (4, 53.2)
        (5, 50.6)
    };
    \addlegendentry{\footnotesize EVA-02 + Pixmix}

       \addplot[line width=0.8pt,color=tu21, mark=*, mark options={scale=0.8}] coordinates {
        (1, 60.6)
        (2, 58.83) 
        (3, 56.92) 
        (4, 54.0)
        (5, 51.5)
    };
  
  \legend{
IN-21k,
IN-21k + RandAug,
IN-21k + Pixmix,
EVA-02, 
EVA-02 + RandAug, 
EVA-02 + Pixmix, 
EVA-02 + TLDR
  }

    \end{axis}

     \begin{axis}[
    width=55mm,
height=40mm,
at={(5cm,1.1cm)},
        xlabel={\footnotesize Severities},
        ylabel={$mAP_{50}$ (\%)},
        title={\footnotesize \textbf{Object Detection}},
        title style={yshift=-1.5ex},
        xtick={1,2,3,4,5},
        xticklabels={1,2,3,4,5},
        label style={font=\footnotesize},
        tick label style={font=\footnotesize}, 
        y label style = {yshift=-0.5cm}, 
        legend cell align=left,
        legend columns=1
    ]
    \addplot[line width=0.8pt,color=tu41, mark=*, mark options={scale=0.8}] coordinates {
        (1, 34.27)
        (2, 32.05) 
        (3, 29.73) 
        (4, 27.07)
        (5, 24.62)
    };
    \addplot[line width=0.8pt,color=tu111, mark=*, mark options={scale=0.8}] coordinates {
        (1, 41.79)
        (2, 39.39) 
        (3, 36.68) 
        (4, 33.39)
        (5, 30.54)
    };

     \addplot[line width=0.8pt,color=tu51, mark=*, mark options={scale=0.8}] coordinates {
        (1, 39.53)
        (2, 36.93) 
        (3, 34.71) 
        (4, 31.19)
        (5, 27.95)
    };

    \addplot[line width=0.8pt,color=tu103, mark=*, mark options={scale=0.8}] coordinates {
        (1, 36.5)
        (2, 34.1) 
        (3, 31.8) 
        (4, 28.5)
        (5, 25.5)
    };

      \addplot[line width=0.8pt,color=tu131, mark=*, mark options={scale=0.8}] coordinates {
        (1, 41.5)
        (2, 38.7) 
        (3, 35.9) 
        (4, 32.4)
        (5, 29.0)
    };

     \addplot[line width=0.8pt,color=tu91, mark=*, mark options={scale=0.8}] coordinates {
        (1, 36.0)
        (2, 33.7) 
        (3, 31.8) 
        (4, 28.8)
        (5, 26.3)
    };

    \end{axis}

\end{tikzpicture}

%% file: tables/encoder_ablation_m2f_decoder.tex
\extrarowheight=\aboverulesep
    \addtolength{\extrarowheight}{\belowrulesep}
    \aboverulesep=0pt
    \belowrulesep=0pt
    \resizebox{0.5\columnwidth}{!}{%
    \begin{tabular}{cc|ccccc}
    \textbf{Init} & \textbf{Encoder} & \multicolumn{5}{c}{\textbf{mIoU in \%}}\\
    \hline
   & & $\csteststar$ & $\bddteststar$&$\mvteststar$ &$\acdcteststar$ &\raisebox{-4.0pt}{\shortstack{\textbf{DG} \\\textbf{mean}}}  \rule[-1.2ex]{0mm}{3.65ex}\\
    \cline{3-7}
    CLIP&\network{ViT-B-16} &  50.0  & 45.0
  & 54.3 & 42.5  & 48.0 \\
    \hline
    CLIP&\network{ViT-L-14}& 55.6
   & 52.5& 59.9&51.5 & 54.9 \\
\hline
    EVA-02-CLIP*&\network{EVA-02-S-16}&  45.9 & 47.5 & 51.7 & 41.3 & 46.6 \\
    \hline
     EVA-02-CLIP&\network{EVA-02-B-16}&  54.9 & 50.3 & 57.2 & 46.6 & 52.3 \\
\hline
    
    EVA-02-CLIP &\network{EVA-02-L-14}& 65.3 & 58.3 & 66.0 &  62.6 & 63.1\\
    \end{tabular}}

%% file: supplement.tex
\clearpage
\setcounter{page}{1}
\setcounter{section}{0}
\begin{center}
    \textbf{\Large Strong but simple: A Baseline for Domain Generalized Dense Perception by CLIP-based Transfer Learning}\\
        Supplementary Material \\
\end{center}

\section{Theoretical Motivation}
\label{sec:theoretical_motivation}
To motivate our fine-tuning setting mathematically, we consider the domain shift from $\mathcal{D}^S$ to $\mathcal{D}^T$ as a bijective map $\mathcal{D}^S\ni \mathbf{x}\mapsto \phi(\mathbf{x})\in\mathcal{D}^T$. Note that learning such maps is a common computer vision task \cite{Zhou2017}. Let $\mathcal{D}^{S}\cup\mathcal{D}^T \ni \mathsf{x}\mapsto T(\mathbf{x})\in \mathcal{T}$ be the map which provides the text description to the image $\mathbf{x}$. We assume that, with high probability, the text description does not refer to the domain and thus remains valid in the new domain. E.g., if a scene in sunshine $\mathbf{x}$ is transferred to the rainy domain by $\phi(\textbf{x})$, the text description only changes if there is an explicit mention of weather in either of the scenes, which is assumed to be rare. Mathematically, this is expressed as $T(\mathbf{x})=T(\phi(\mathbf{x}))$ for $\mathbf{x}\in\mathcal{D}^S $ with probability $p\gg (1-p)\geq 0$. Furthermore, we  assume perfect image-to-text feature alignment, i.e. $\VEC{M}_E^V(\mathbf{x})=\VEC{M}_E^L(T(\mathbf{x}))$ for $\mathbf{x}\in\mathcal{D}^S\cup\mathcal{D}^T$. Then we have:
\begin{lemma}\label{lem:Lem1}
Under the above assumptions, the encoder-decoder network $\VEC{M}=\VEC{M}^V_D\circ \VEC{M}_E^V$ is domain robust, i.e. provides the same output to $\mathbf{x}$ and $\phi(\mathbf{x})$, with probability not smaller than $p$.   
\end{lemma}
\begin{proof}
With probability not less than $p$, we have for $\mathbf{x}\in\mathcal{D}^S$ 
\[
\VEC{M}_E^V(\phi(\mathbf{x}))=\VEC{M}_E^L(T(\phi(\mathbf{x})))=\VEC{M}_E^L(T(\mathbf{x})) =\VEC{M}_E^V(\mathbf{x}).
\]
Application of $\VEC{M}^V_D$ to both sides completes the proof.
\end{proof}
Let us remark that the task of the decoder network $\VEC{M}_D^V$ plays no role in the proof of Lemma \ref{lem:Lem1} which is equally valid for semantic segmentation, object detection or other downstream tasks. The mechanics of this lemma is simply based by an alignment of the invariances of the network in the sense of \cite{dosovitskiy2016inverting,rombach2022invertible} with the direction of the domain shift via feature alignment with an auxiliary modality (language, in our case) with invariance under the given shift already at the level of the data itself. Our practical implementation of this idealized description builds upon large-scale vision-language pre-trainings as done by CLIP \cite{radford2021learning}, or EVA-CLIP \cite{fang2023eva} with image-to-text feature alignment as a training objective. These produce highly generalized encoder representations for $\VEC{M}_E^V$. We utilize this pre-trained vision encoder in a simple, transfer learning-only setting for our investigations. In the following, we will introduce the details.
\section{ResNet-101 \& Synthia Experiments}
In \cref{tab:sota_resnet}, the domain generalization performance of our VLTSeg approach is shown with a \network{ResNet-101} backbone. When training on GTA5, we perform competitively with previous SOTA in the DG mean, demonstrating the effectiveness of fine-tuning for convolutional neural networks. It has to be considered that the EVA-02 \cite{EVA-CLIP} pre-trained initialization was not available for the \network{ResNet-101} backbone, so the CLIP \cite{radford2021learning} initialization was used. Moreover, it is important to note that some recent approaches, like DIDEX \cite{niemeijer2024generalization} or CLOUDS \cite{benigmim2024collaborating}, rely on other foundation models like stable diffusion for data generation/augmentation or label refinement by SAM \cite{kirillov2023segment}. That creates a significant advantage during training compared to approaches that do not incorporate other foundation models.\\
We also performed experiments on the Synthia \cite{ros2016synthia} dataset to verify the effectiveness on another synthetic dataset. As shown in \cref{tab:sota_synthia} VLTSeg outperforms most of the state-of-the-art approaches and shows only a slightly lower generalization compared to DIDEX \cite{niemeijer2024generalization} which uses additional generated data from a stable diffusion model.
\begin{table}
  \centering
  \renewcommand{\arraystretch}{.9}
  \setlength{\tabcolsep}{.3em}
  \caption{\textbf{Domain generalization performance} in comparison with state-of-the-art approaches. Training was performed on the synthetic GTA5 ($\src\!=\!\gtavtrain$) dataset. All our experiments employed a CLIP \cite{radford2021learning} pre-trained \network{ResNet-101} \cite{he2016deep} backbone (therefore denoted as VLTSeg-R) with a Mask2Former \cite{cheng2022masked} head. Prior work results are cited from the respective paper, only the values for works marked with $\circ$ are taken from \cite{Peng2022semanticaware}. 
  }
  \input{tables/resnet_sota}
  \label{tab:sota_resnet} 
\end{table}
\section{Decoder Architecture}
We also examine the impact of different decoder architectures on the domain generalization performance, as shown in \cref{tab:decoder_ablation}. The DG mean of the model with the Mask2Former \cite{cheng2022masked} decoder is 4.1{\%} higher than the model a Semantic FPN head and also 1.5{\%} better than the ASPP-based decoder from DAFormer \cite{Hoyer2022daformer}.
\begin{table}

  \centering
  \renewcommand{\arraystretch}{.9}
  \setlength{\tabcolsep}{.3em}
  \caption{\textbf{Domain generalization performance} in comparison with state-of-the-art approaches. Training was performed on the synthetic SYNTHIA and UrbanSyn dataset dataset. Prior work results are cited from the respective paper. 
  }
  \input{tables/sota_ext_synthia}
  \label{tab:sota_synthia} 
\end{table}
VLTSeg with a Mask2Former \cite{cheng2022masked} decoder performs consistently better across all four real-world datasets than the other decoder architectures. That implies that Mask2Former can leverage the provided visual embeddings from vision-language pre-training more effectively and better focus on domain-invariant features.
\begin{table}
\parbox[t]{.48\linewidth}{\centering

\setlength{\tabcolsep}{1pt}
\scriptsize
 \caption{\textbf{Ablation study of different decoder architectures} and their domain generalization performance (mIoU (\%)). Training was performed on the synthetic GTA5 ($\src\!=\!\gtavtrain$) dataset. Evaluation is performed on the four shown real-world datasets.}
\input{tables/decoder_ablation}
  \label{tab:decoder_ablation} 
}
\hfill
\parbox[t]{.48\linewidth}{
\centering
\setlength{\tabcolsep}{1pt}
\renewcommand{\arraystretch}{.7}
\scriptsize
\caption{\textbf{Computational complexity of the vision encoders} that were employed in our experiments. GFLOPS and Parameters are computed using \network{MMPreTrain} \cite{2023mmpretrain}, numbers are rounded to integers.}
\input{tables/params_flops}
\label{tab:params_flops}}
\end{table}

\section{Sensitivity Analysis of the Training Configuration}
In the analysis with EVA-02-B-CLIP we observe that a change of the learning rate and the optimizer affect the results significantly. Changing the optimizer to SGD or Adam leads to a collapsed training most likely because of the randomly initialized Mask2Former decoder. While a different batch size does not affect results, as expected, a higher learning rate diminishes DG performance more than a lower one.
\begin{figure}
    \centering
    \includestandalone[width=\columnwidth]{fig/sens_analysis}
\end{figure}
\section{Qualitative Results}

\begin{figure}
    \centering
    \begin{tikzpicture}
        \node[](pic1){\includegraphics[width=0.3\textwidth]{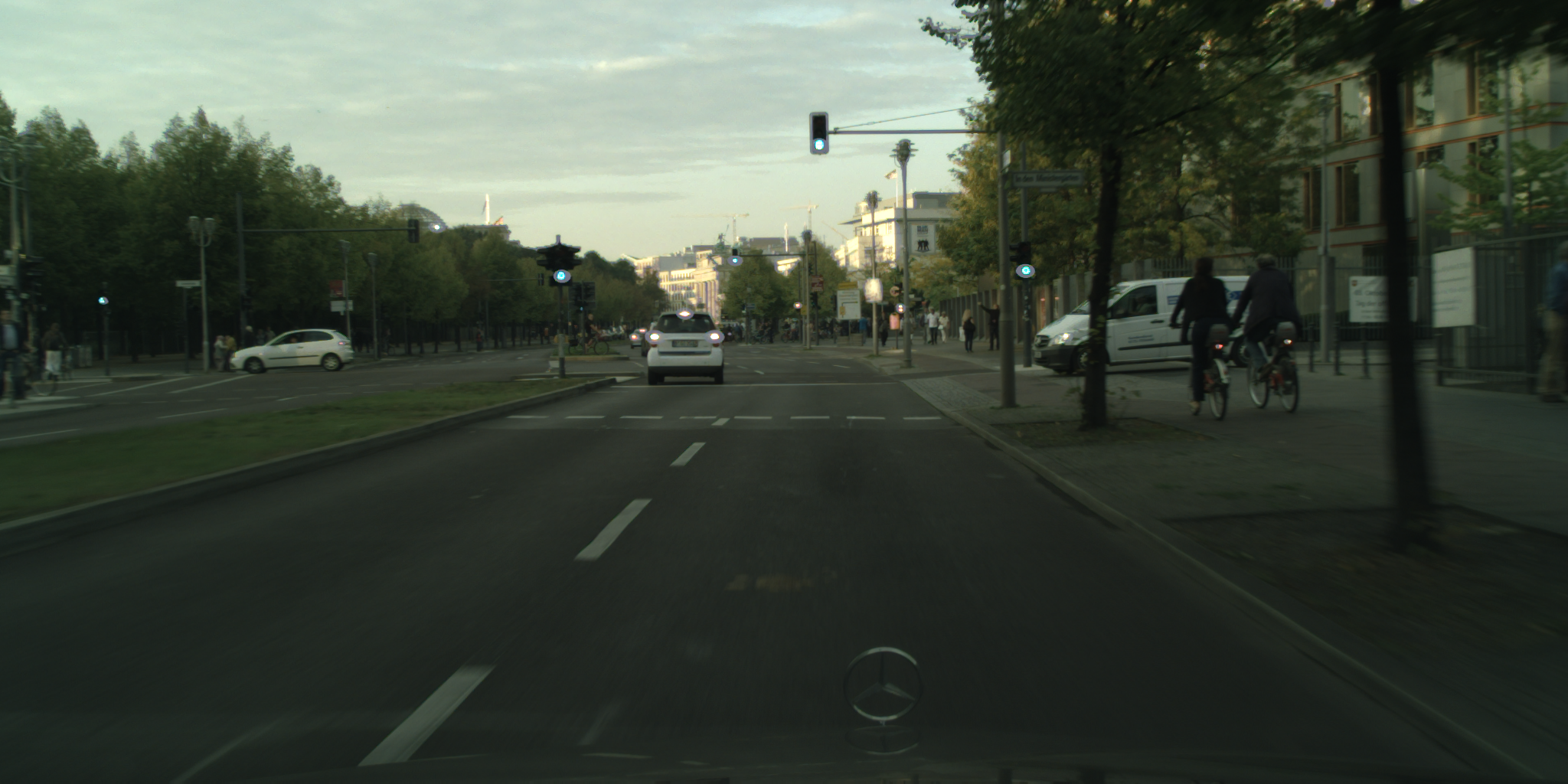}};
        \node[below of = pic1, node distance = 2cm](label1){\includegraphics[width=0.3\textwidth]{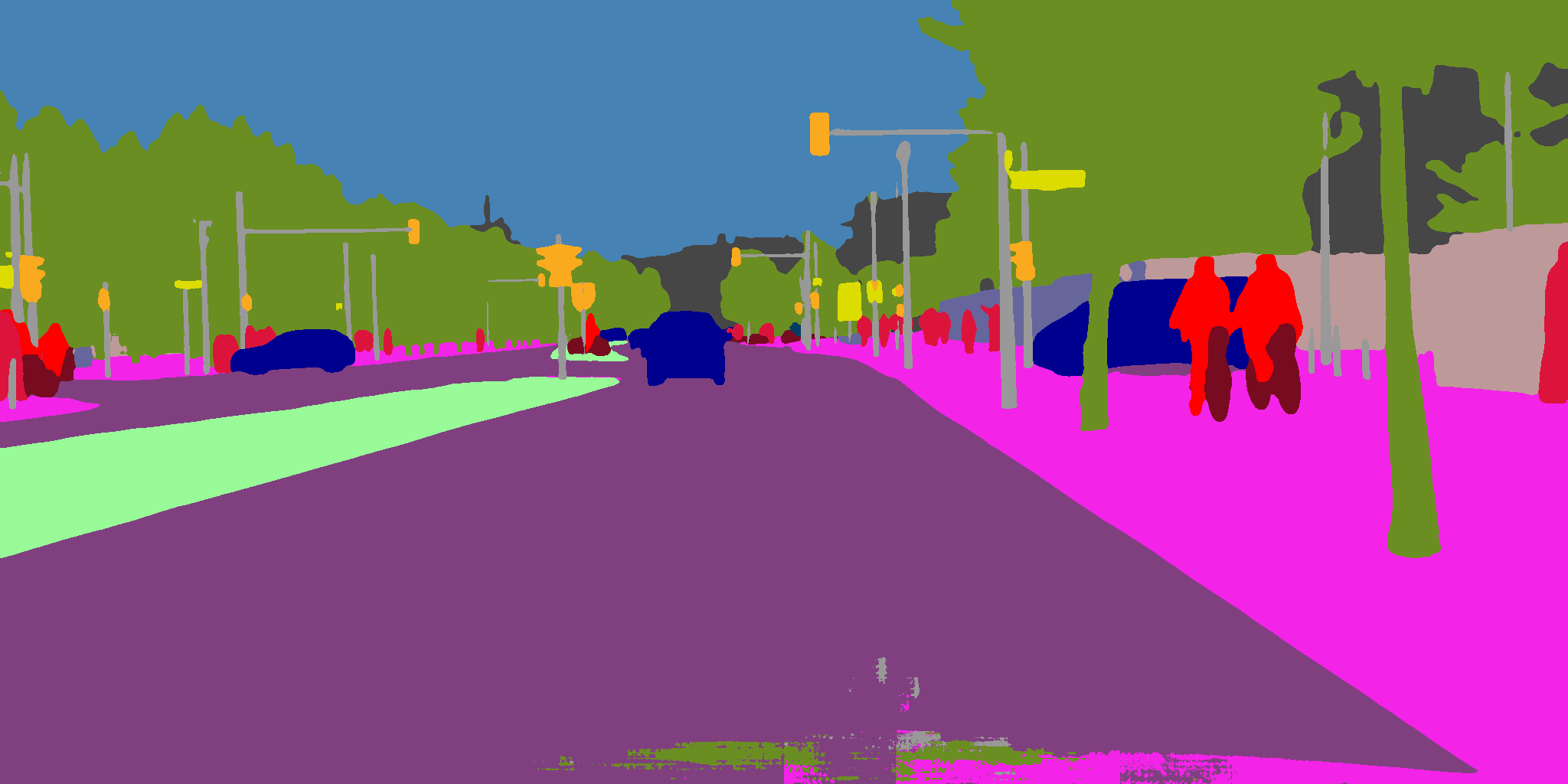}};
        \node[right of = pic1, node distance=4cm](pic2){\includegraphics[width=0.3\textwidth]{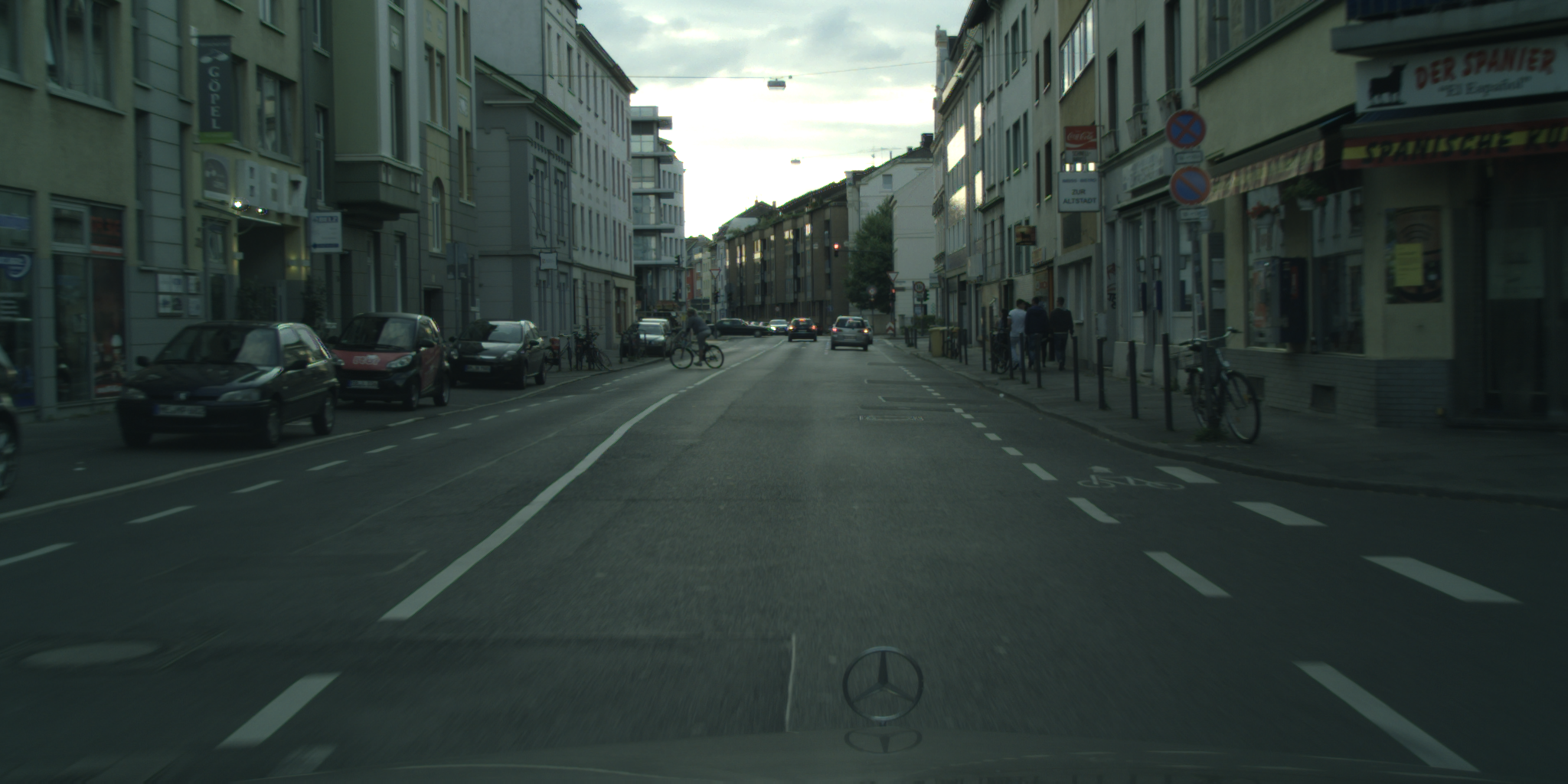}};
        \node[below of = pic2, node distance = 2cm](label2){\includegraphics[width=0.3\textwidth]{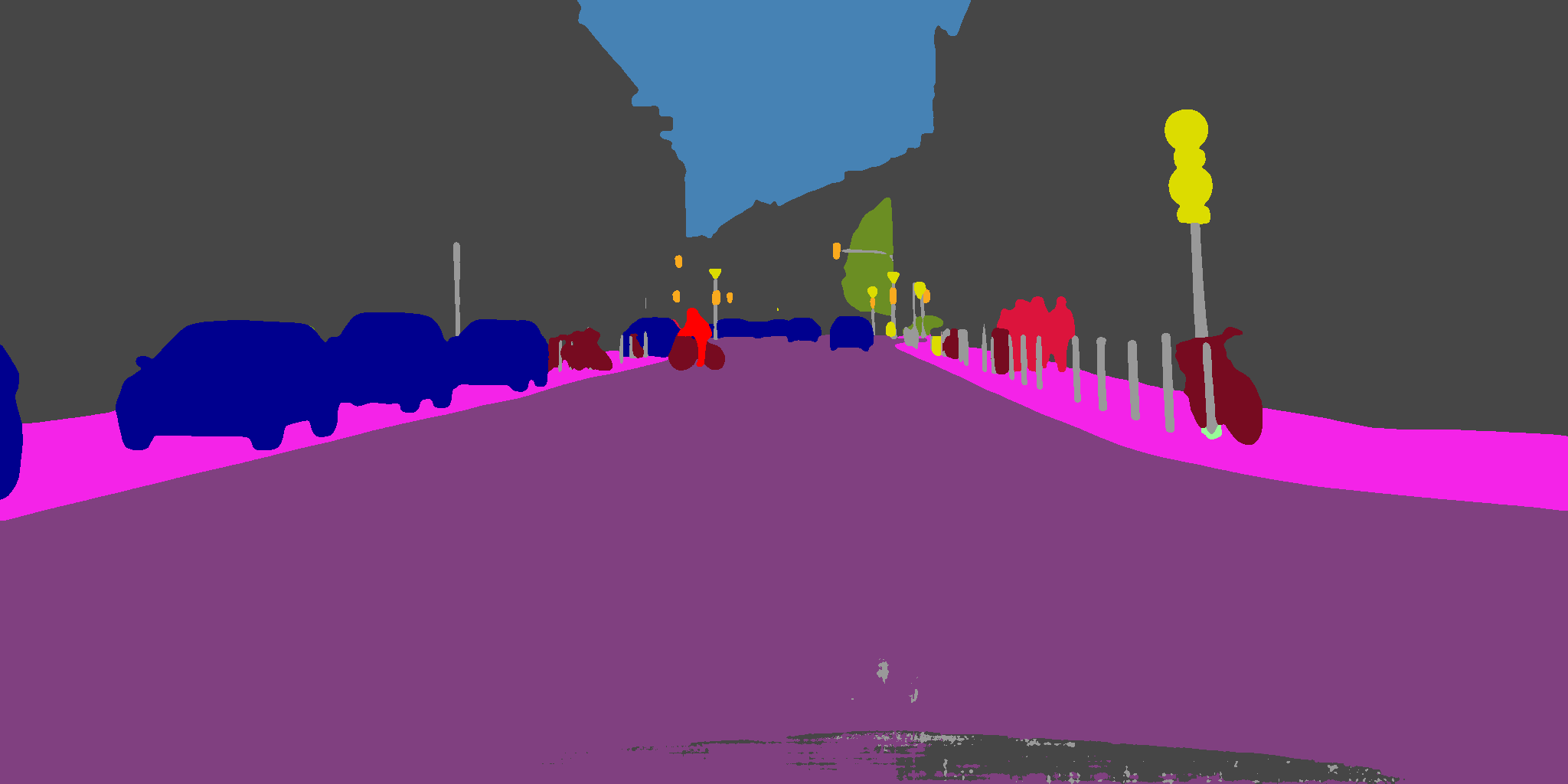}};
         \node[right of = pic2, node distance=4cm](pic3){\includegraphics[width=0.3\textwidth]{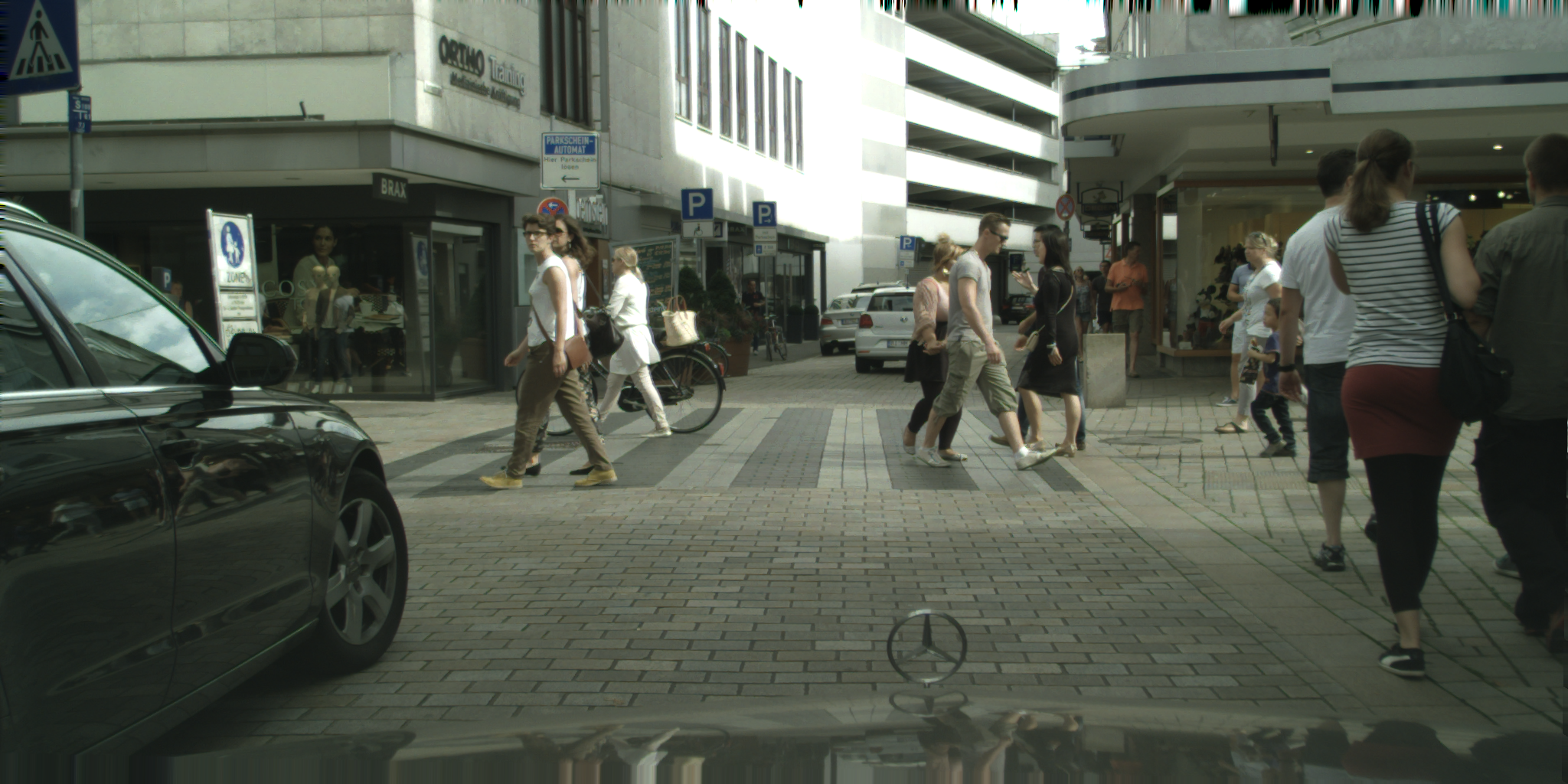}};
        \node[below of = pic3, node distance = 2cm](label3){\includegraphics[width=0.3\textwidth]{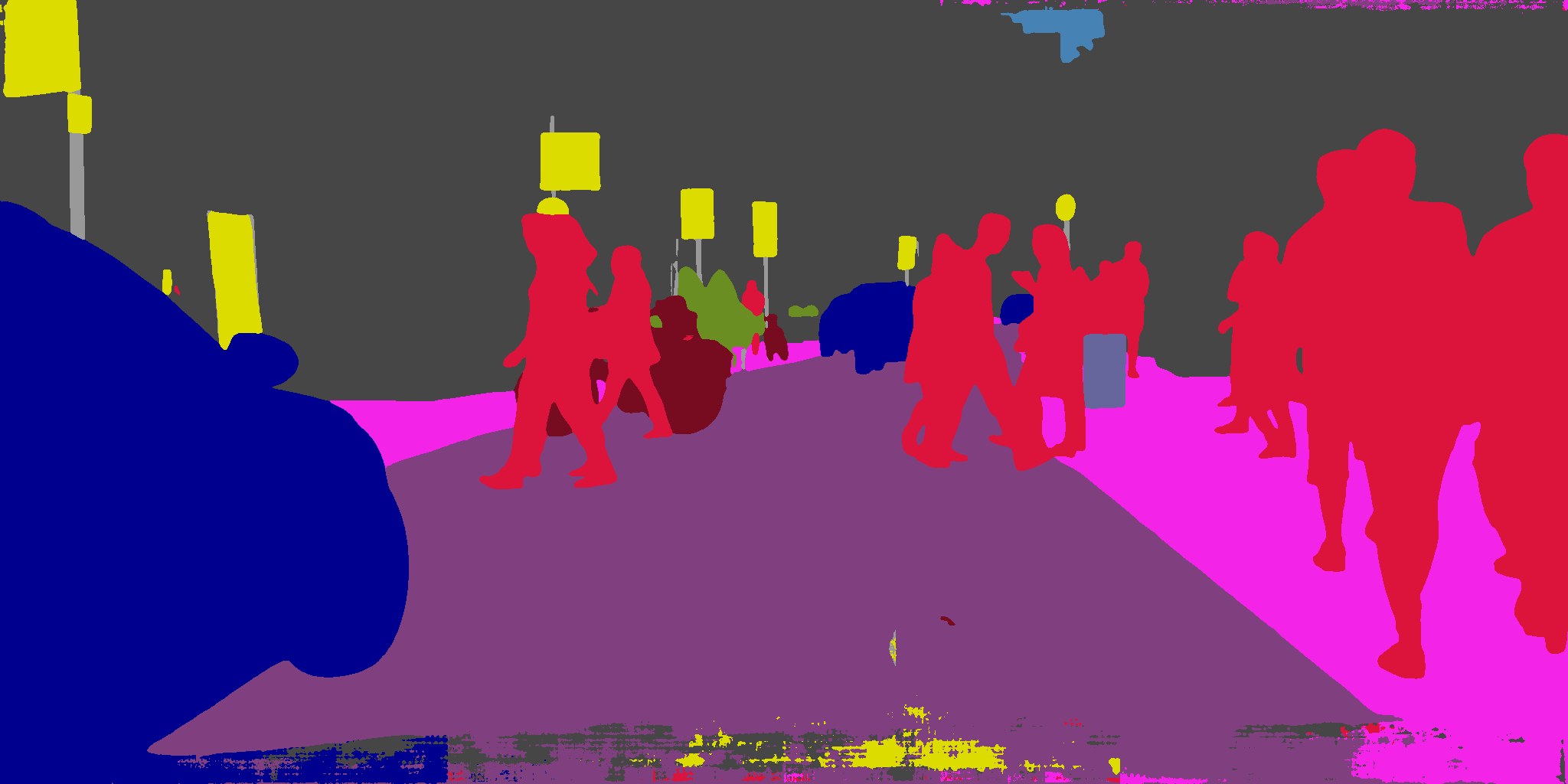}};
    \end{tikzpicture}
    
    \caption{\textbf{Predictions on Cityscapes test set $\cstest$.} Training and evaluation was conducted as described in \cref{sec:testset_eval}}
    \label{fig:cityscapes_sota}
\end{figure}

\fakeparagraph{Cityscapes}
We show predictions on the Cityscapes test set $\cstest$ in \cref{fig:cityscapes_sota} which visualizes the state-of-the-art segmentation quality of our VLTSeg approach when training supervised on Cityscapes.  

\begin{figure*}
    \centering
    \begin{tikzpicture}
        \node[](pic1){\includegraphics[width=0.3\textwidth]{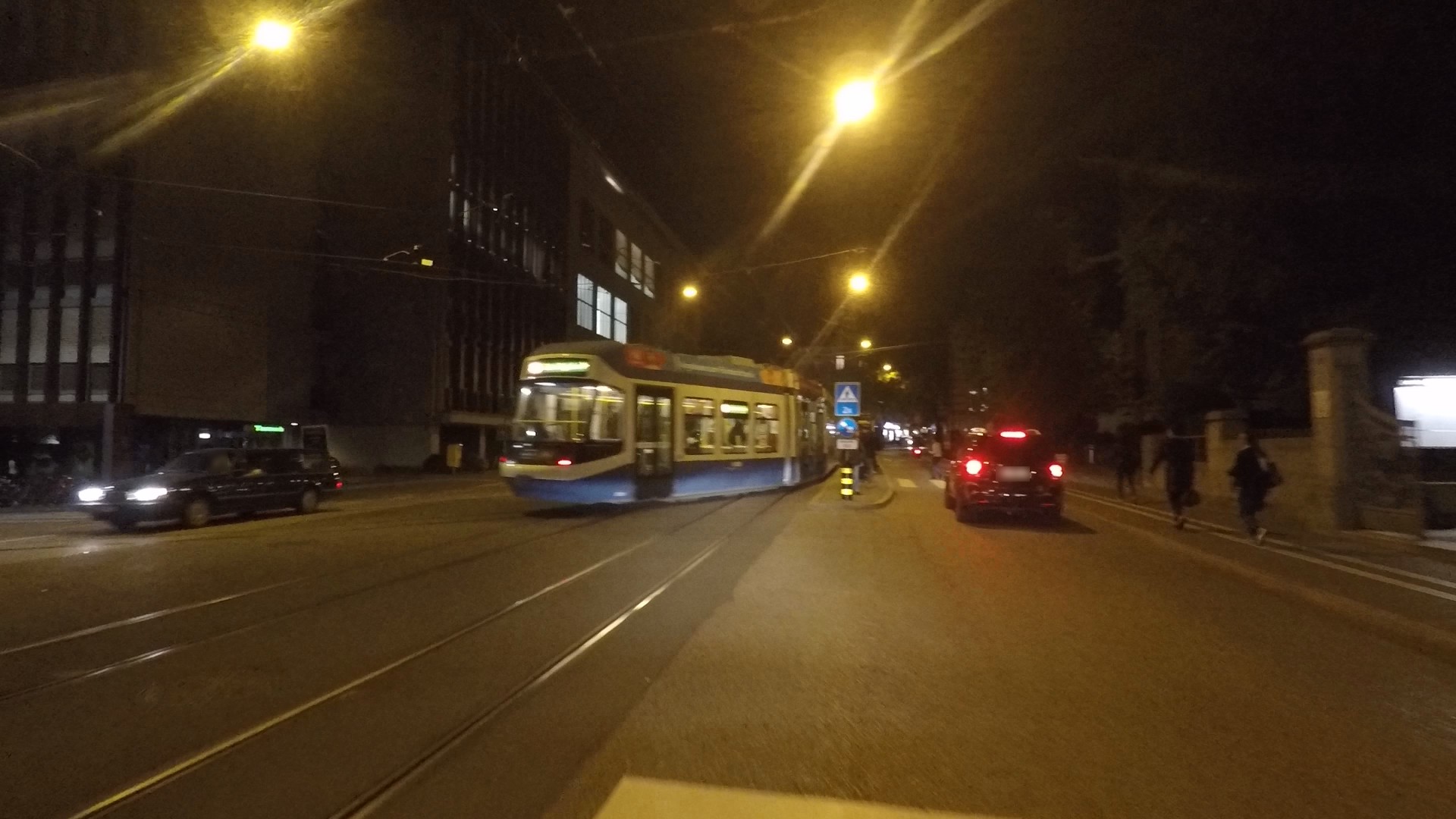}};
        \node[below of = pic1, node distance = 2.4cm](label1){\includegraphics[width=0.3\textwidth]{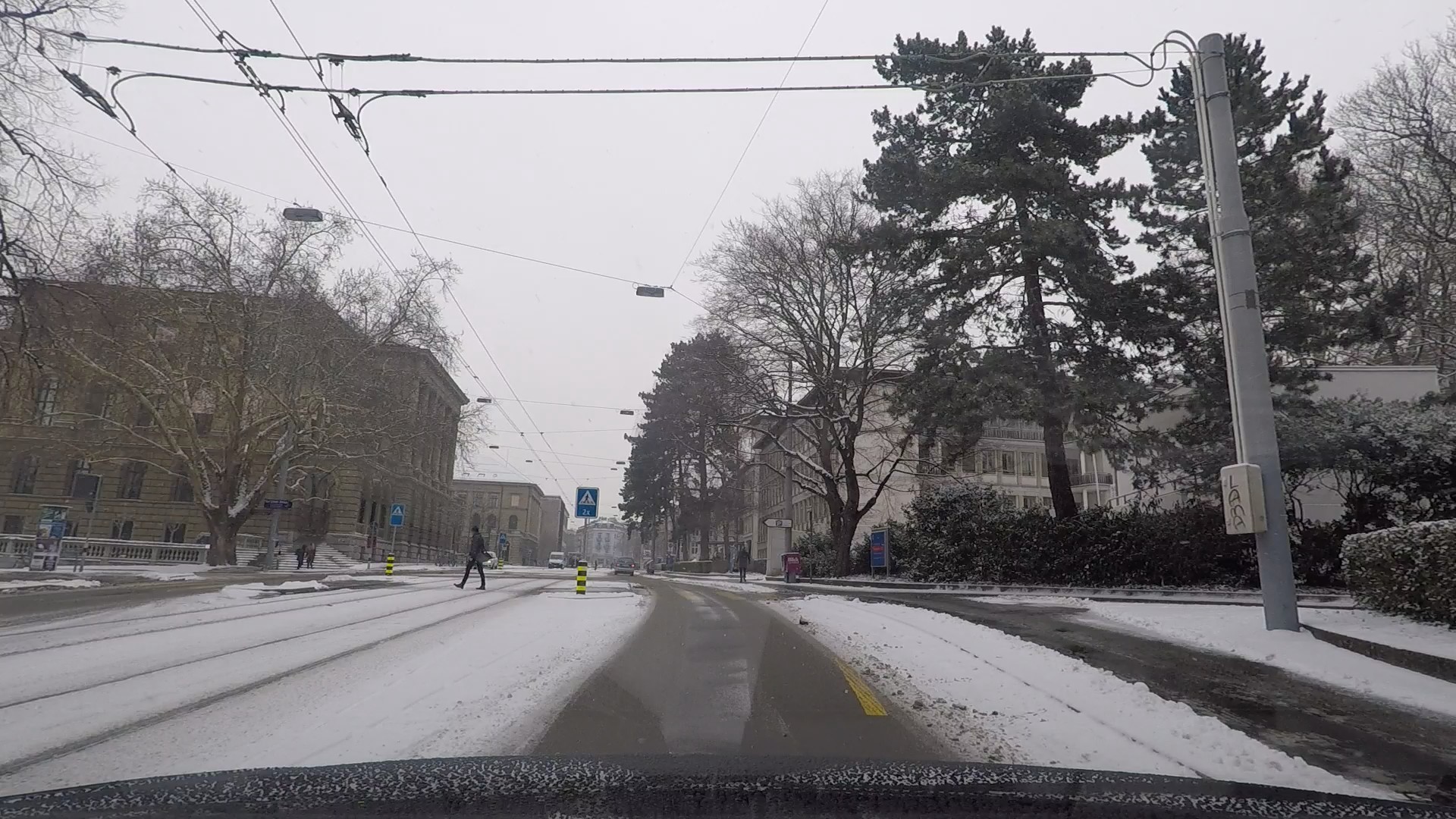}};
        \node[right of = pic1, node distance=4cm](pic2){\includegraphics[width=0.3\textwidth]{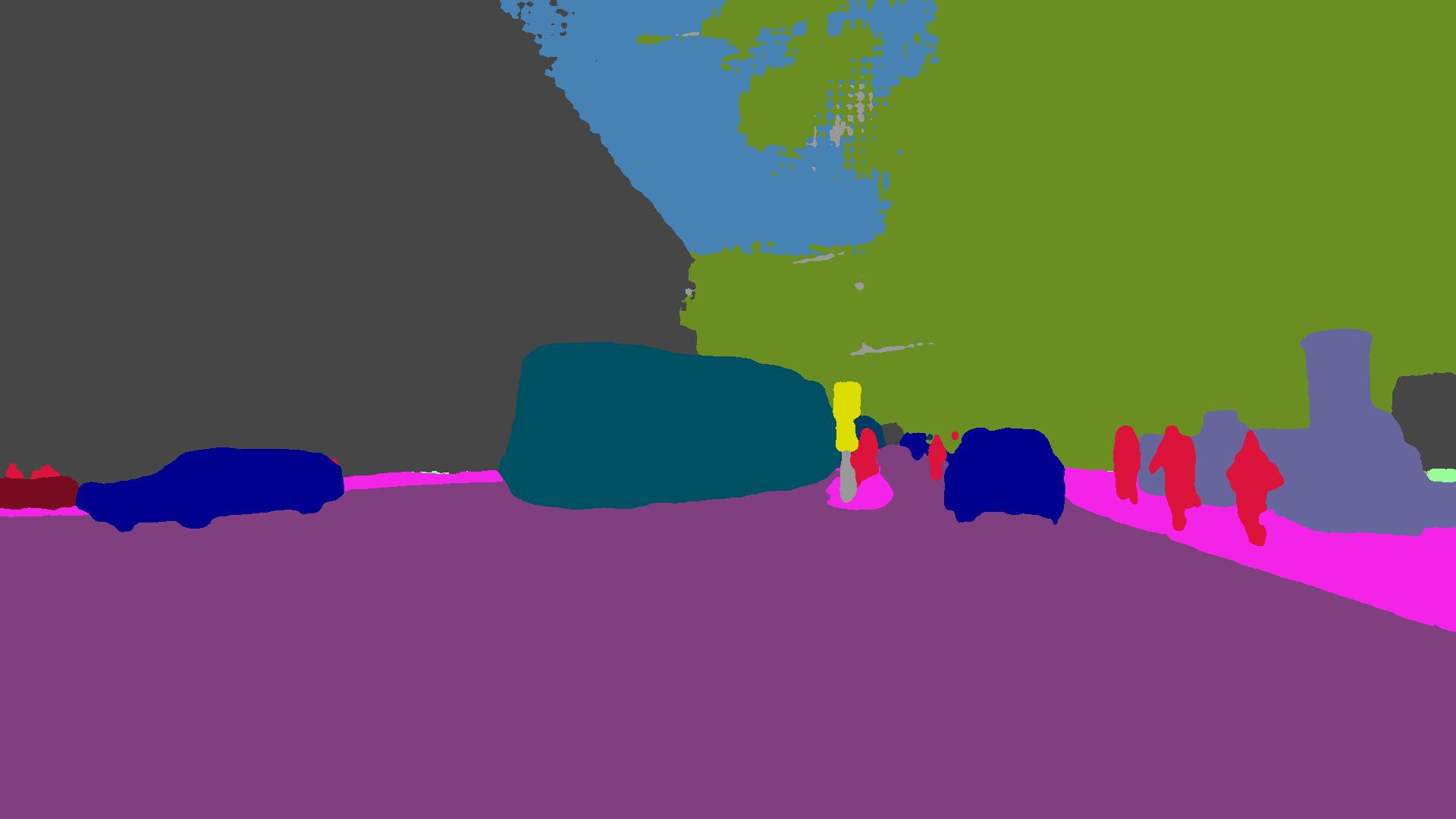}};
        \node[below of = pic2, node distance = 2.4cm](label2){\includegraphics[width=0.3\textwidth]{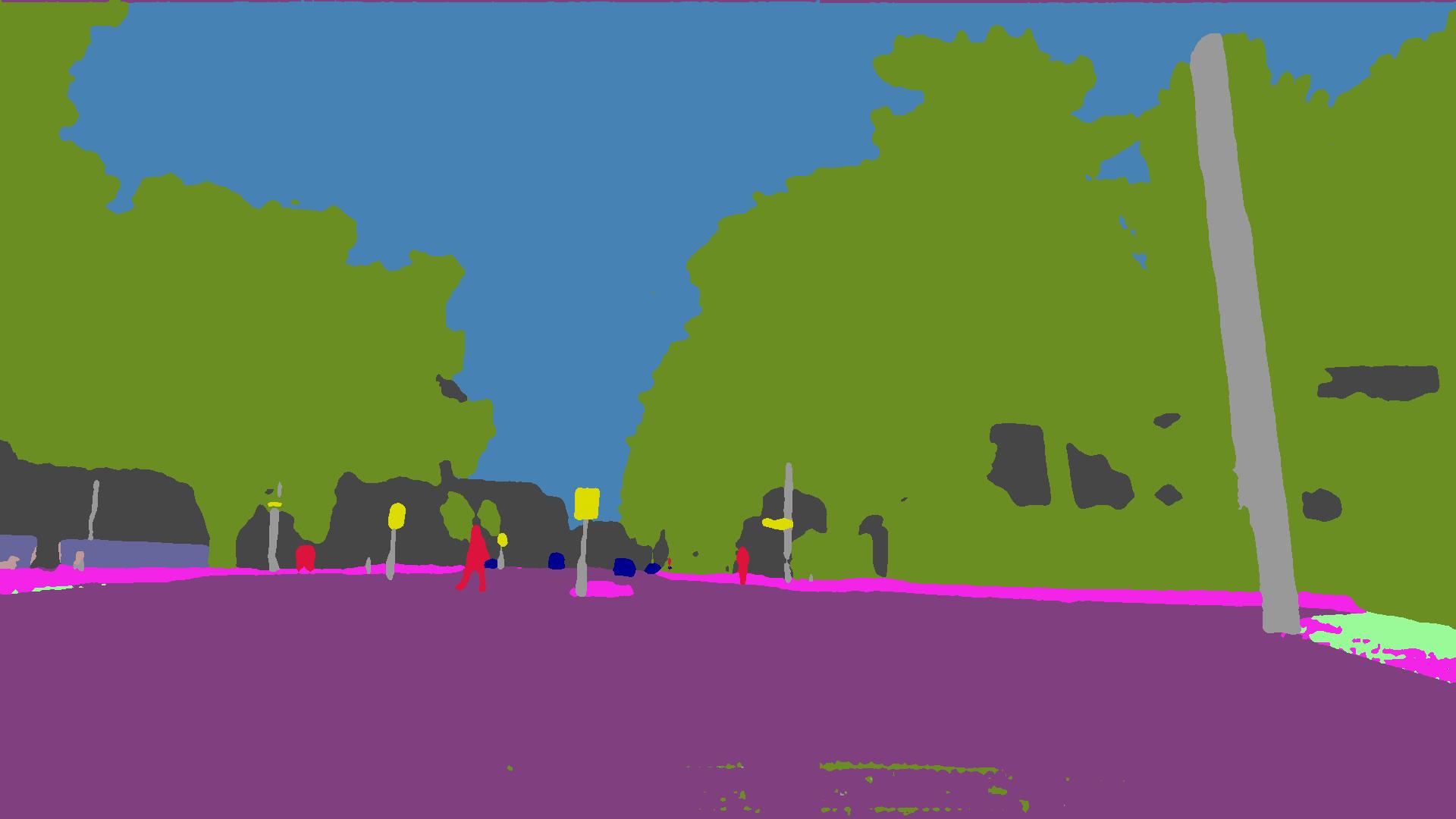}};
         \node[right of = pic2, node distance=4cm](pic3){\includegraphics[width=0.3\textwidth]{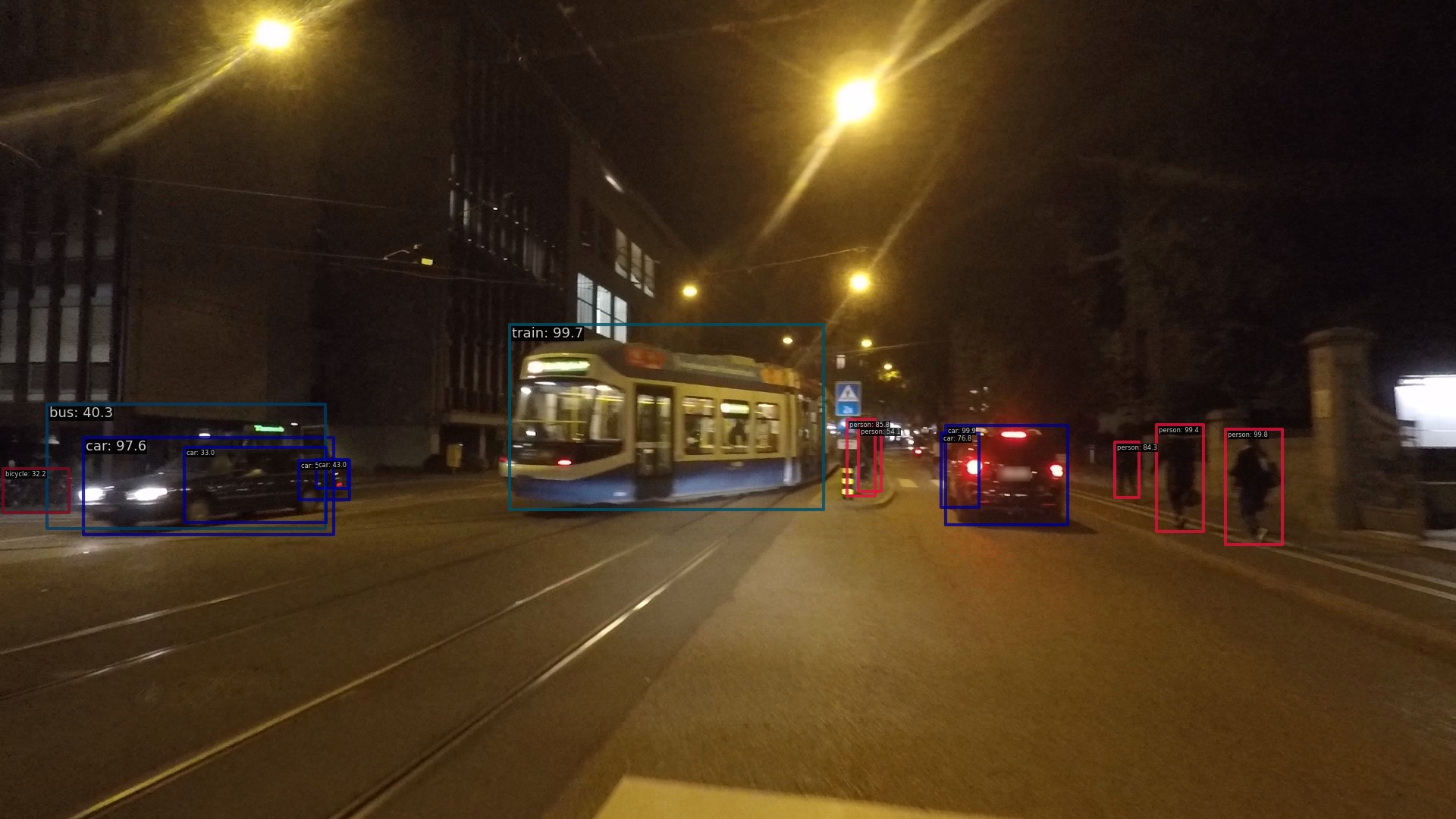}};
        \node[below of = pic3, node distance = 2.4cm](label3){\includegraphics[width=0.3\textwidth]{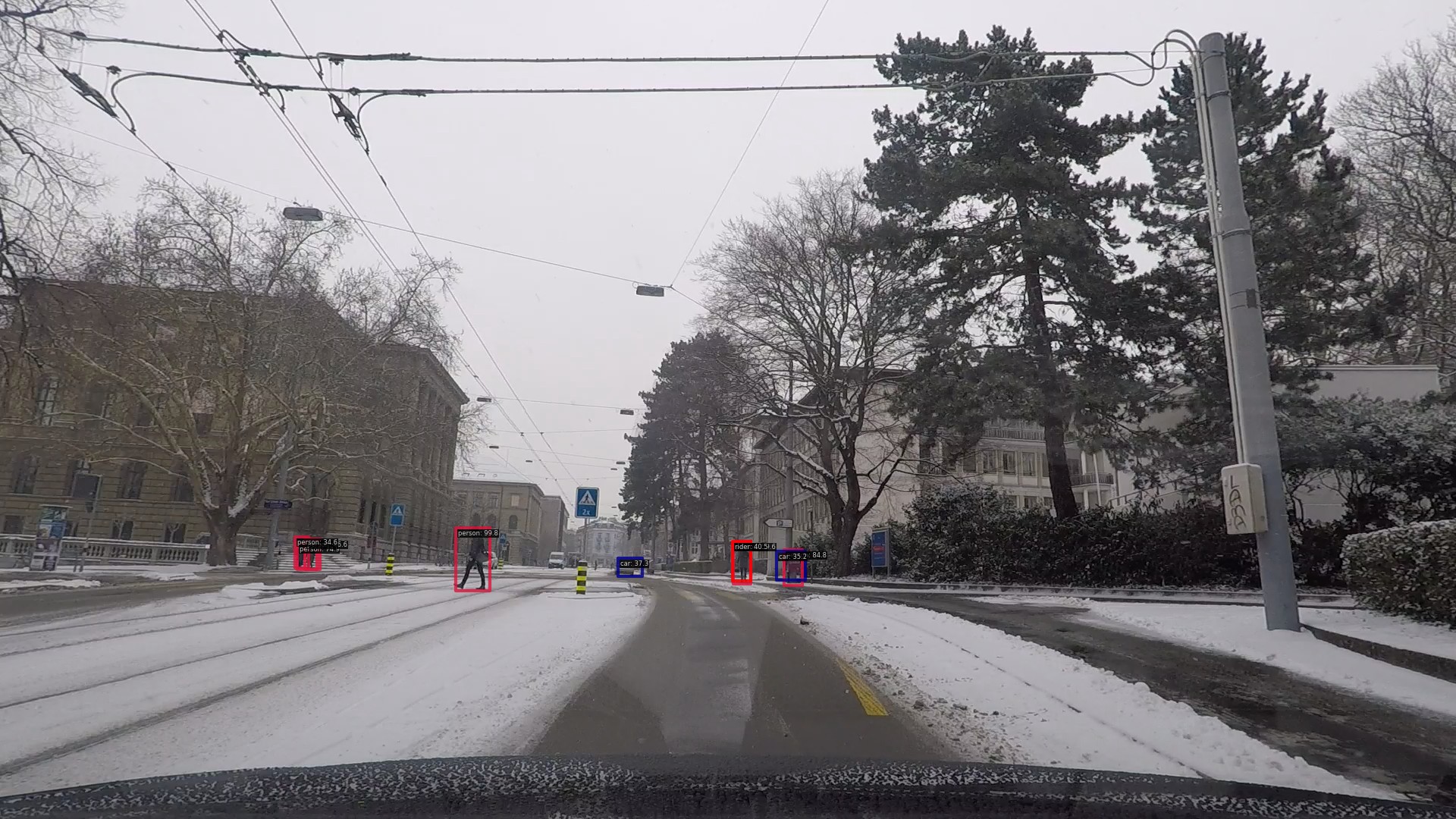}};
        \node[below of = label1, node distance = 2.4cm](img3){\includegraphics[width=0.3\textwidth]{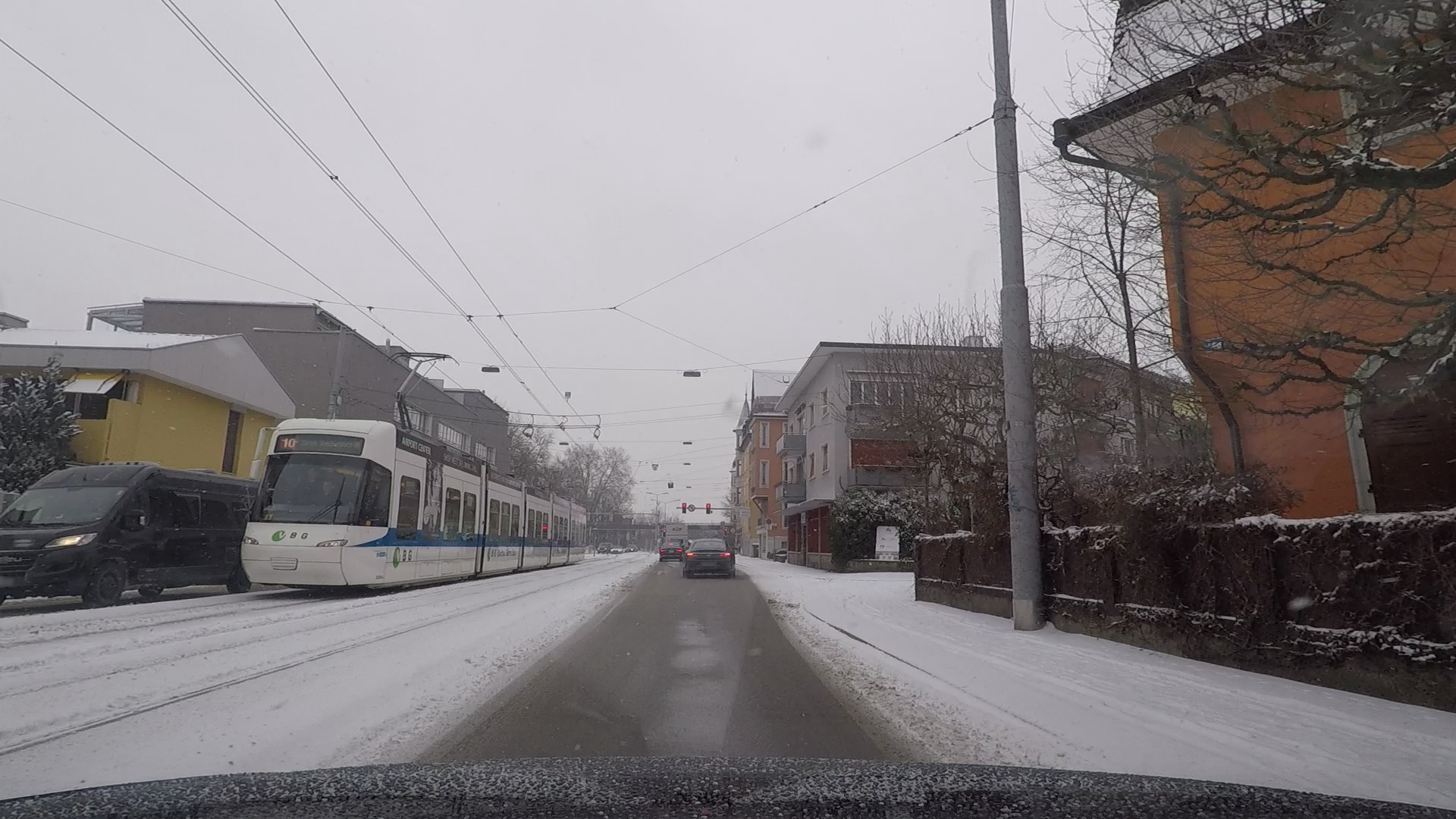}};
        \node[right of = img3, node distance = 4cm](seg3){\includegraphics[width=0.3\textwidth]{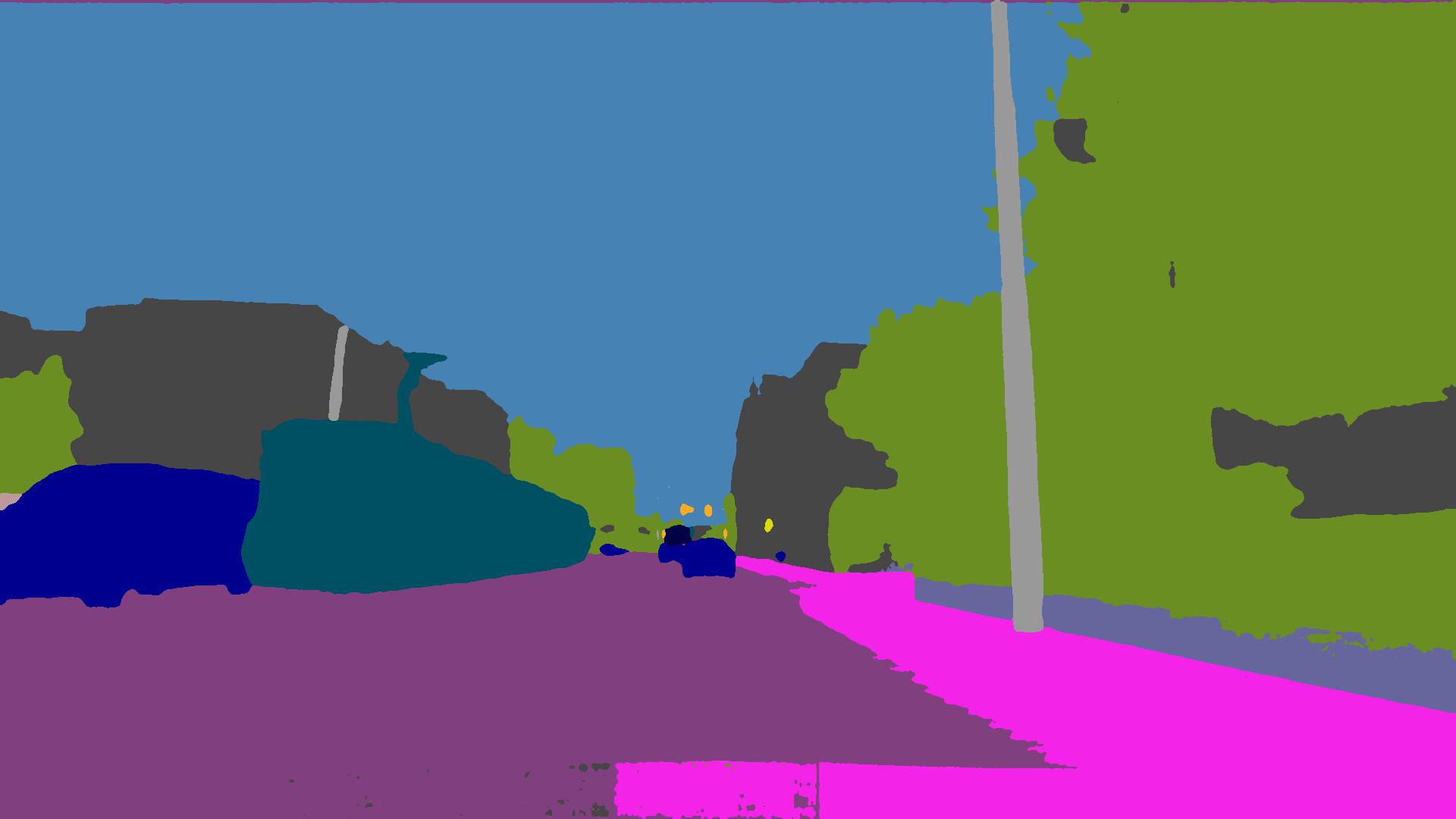}};
        \node[right of = seg3, node distance = 4cm](bbox3){\includegraphics[width=0.3\textwidth]{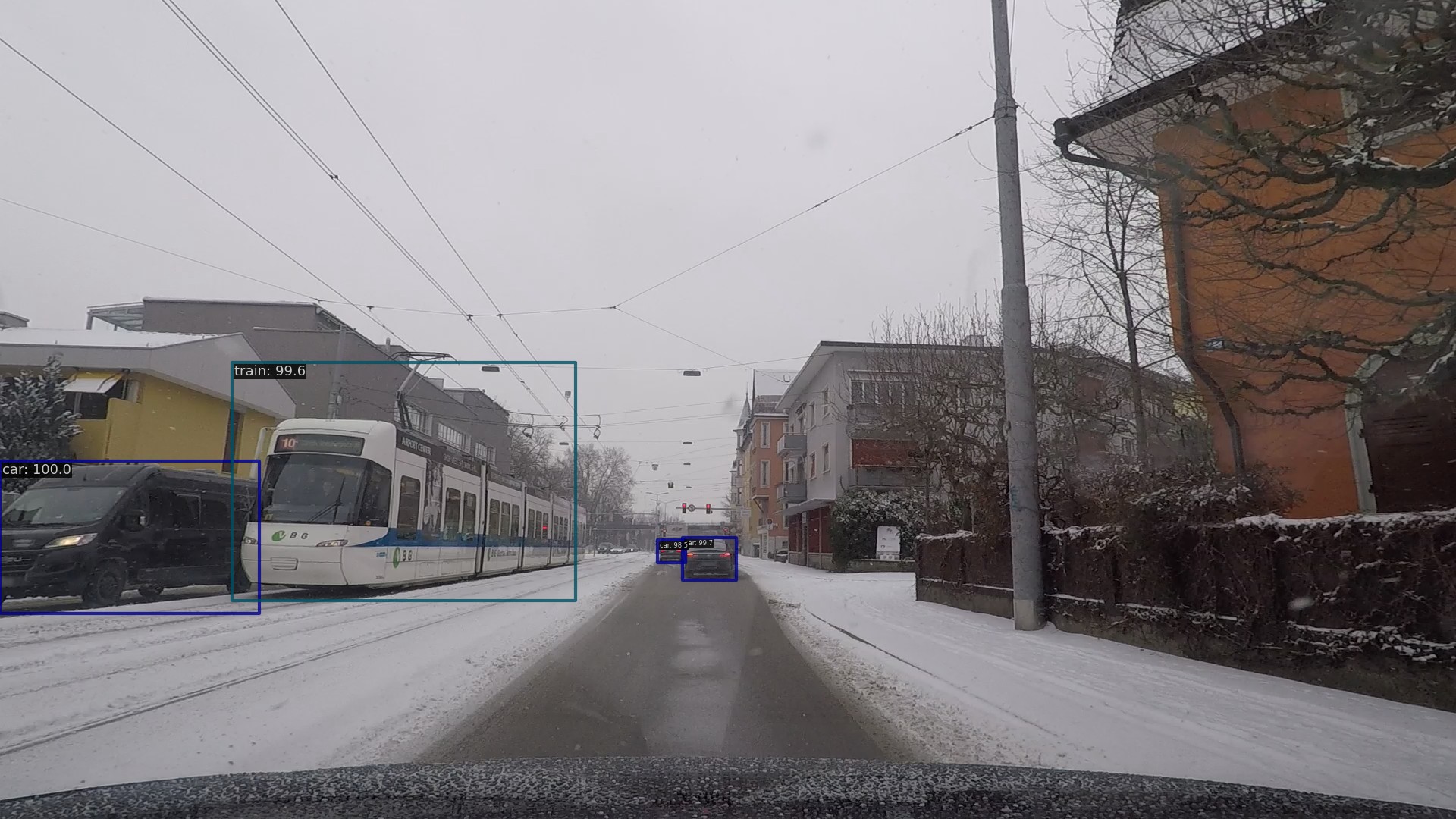}};
        \node[below of = img3, node distance = 2.4cm](img4){\includegraphics[width=0.3\textwidth]{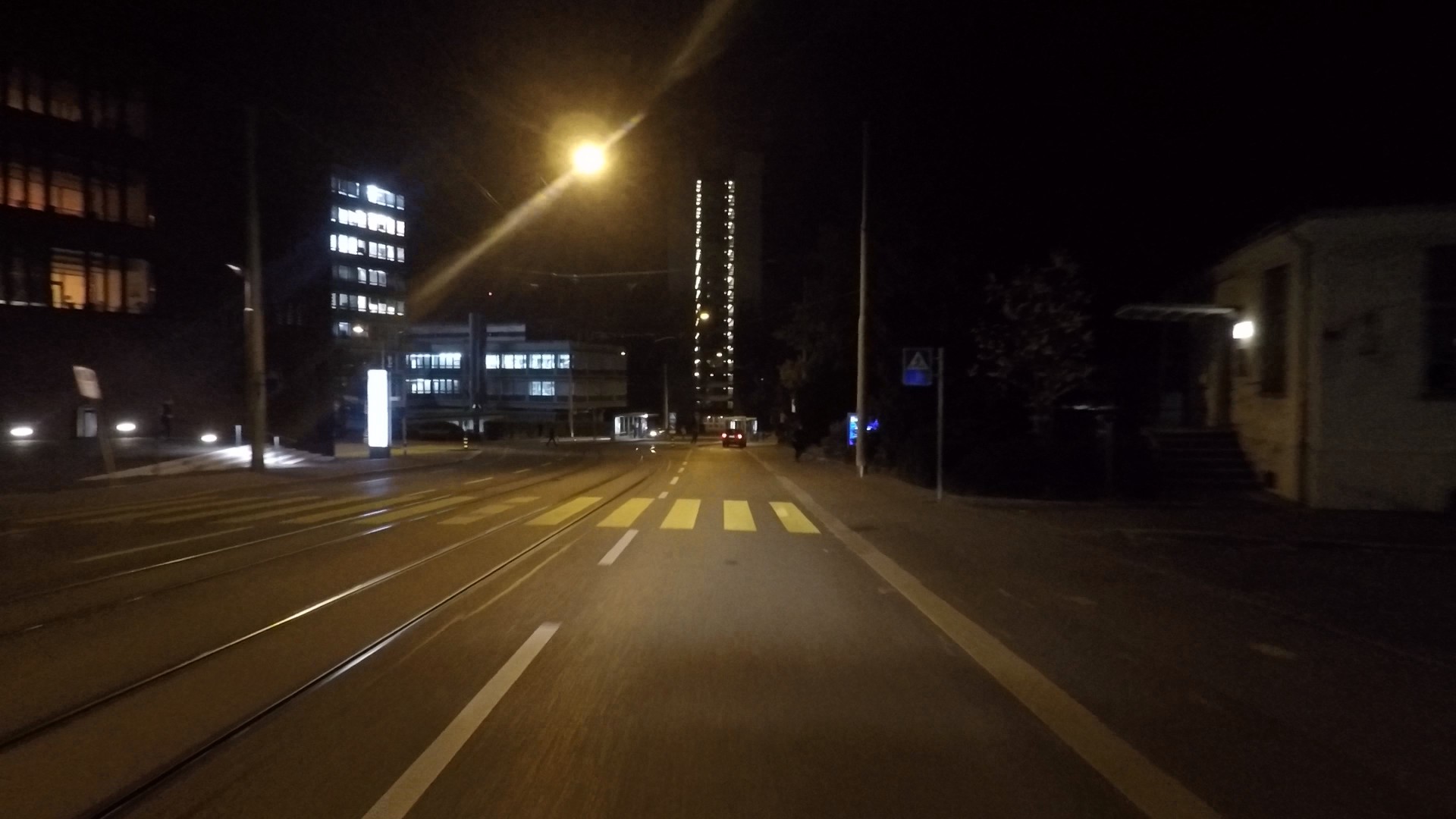}};
        \node[right of = img4, node distance = 4cm](seg4){\includegraphics[width=0.3\textwidth]{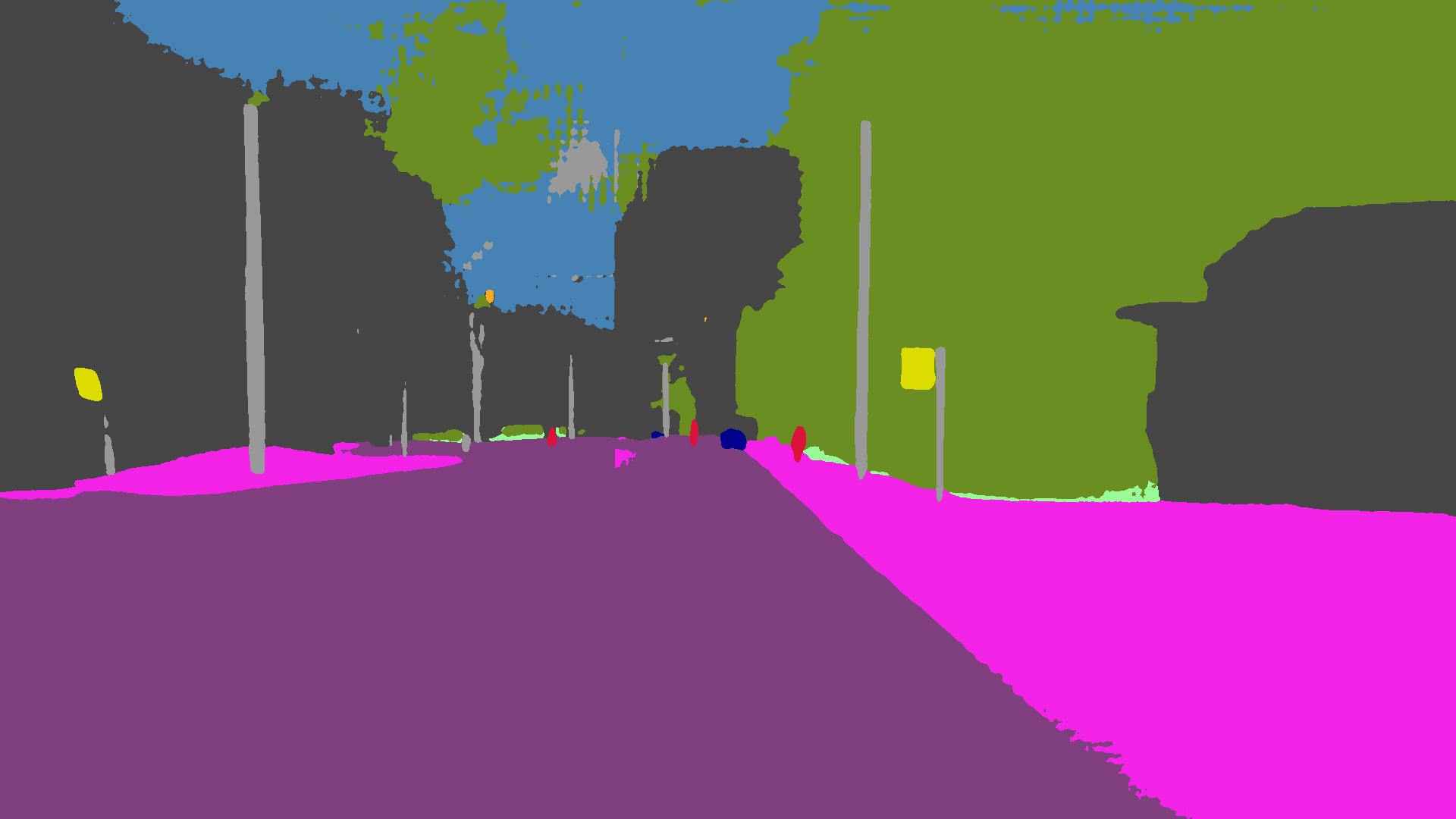}};
        \node[right of = seg4, node distance = 4cm](bbox4){\includegraphics[width=0.3\textwidth]{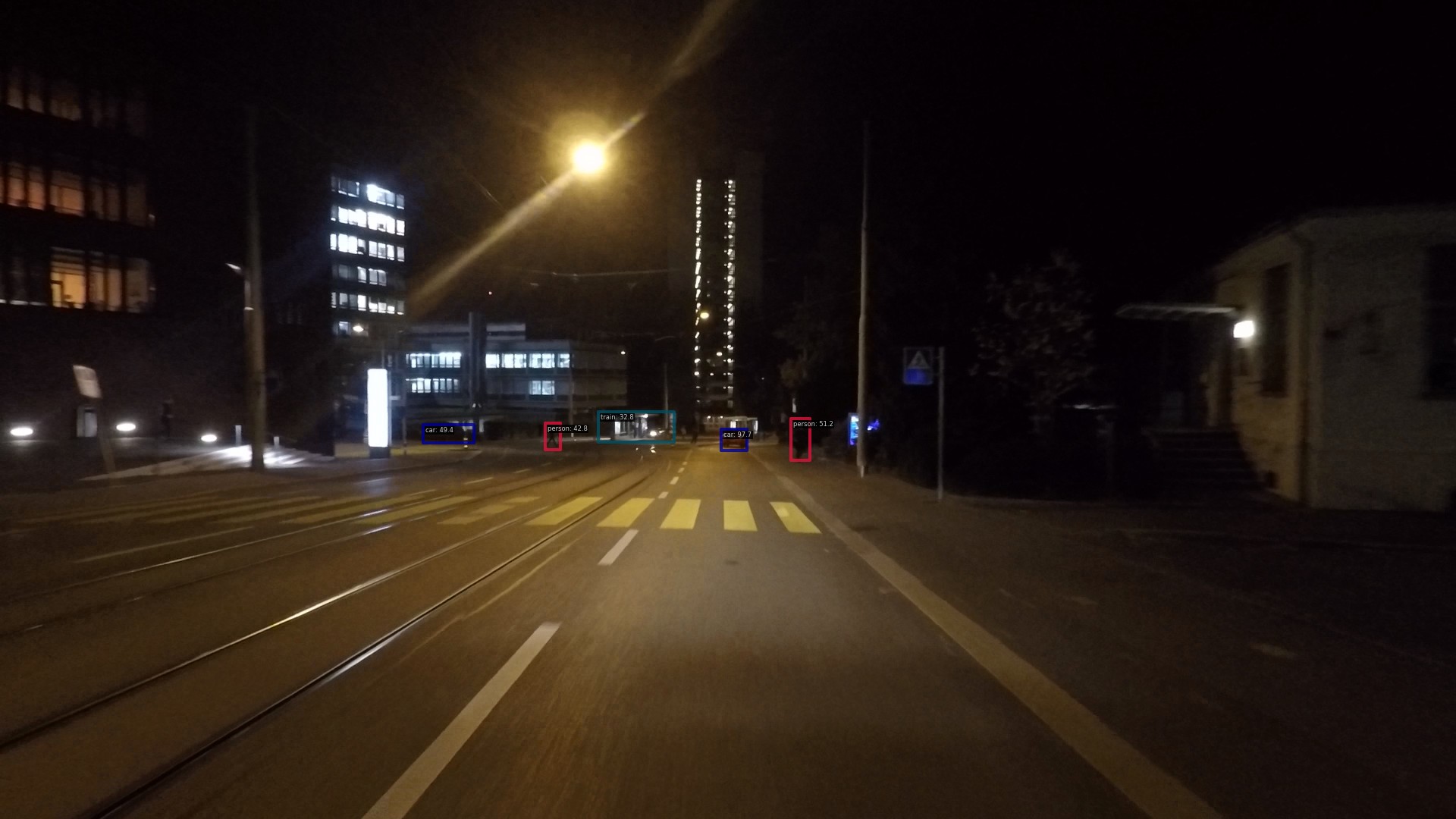}};
        \node[below of = img4, node distance = 2.4cm](img5){\includegraphics[width=0.3\textwidth]{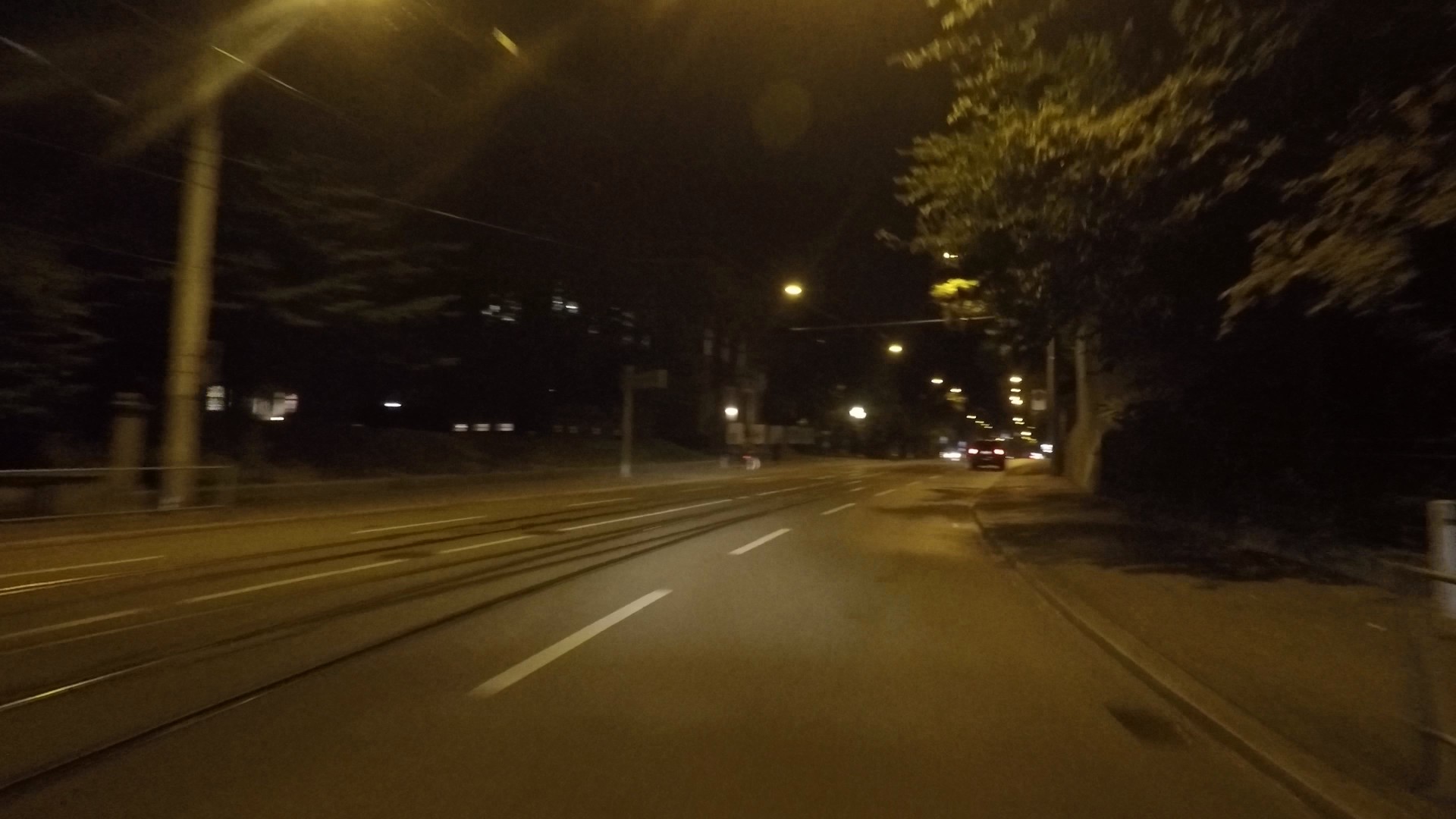}};
        \node[right of = img5, node distance = 4cm](seg5){\includegraphics[width=0.3\textwidth]{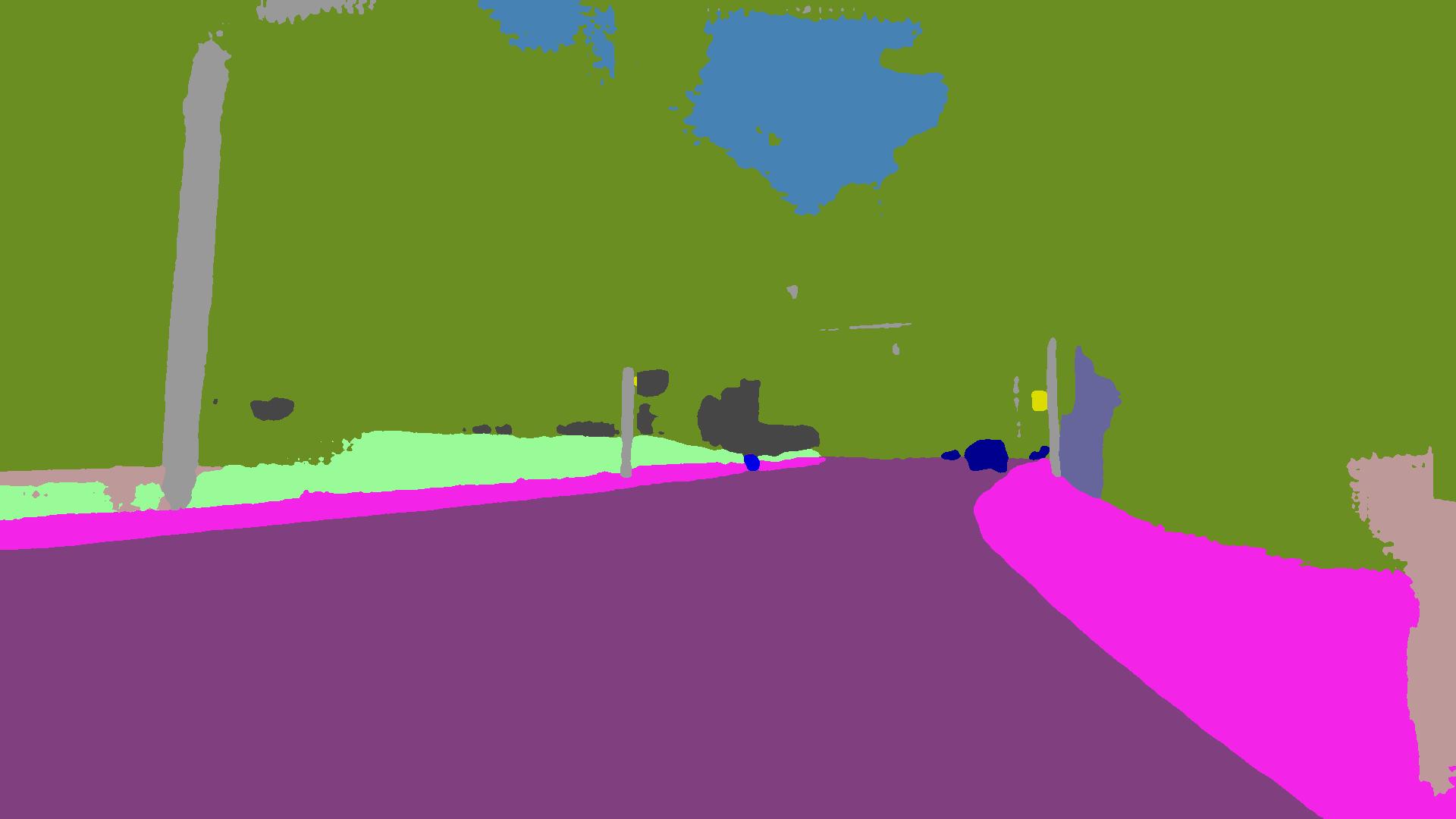}};
        \node[right of = seg5, node distance = 4cm](bbox5){\includegraphics[width=0.3\textwidth]{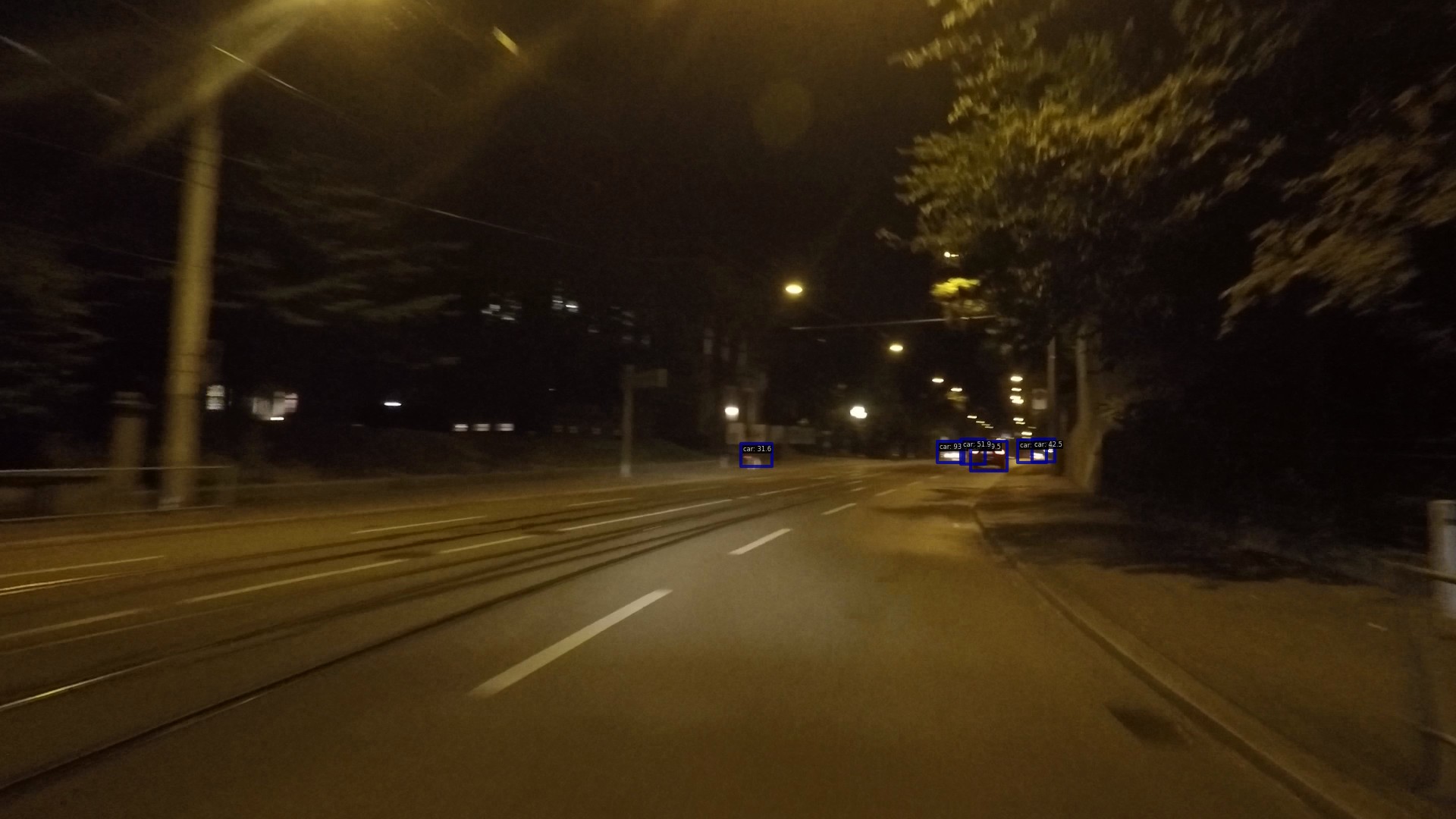}};
        \node[below of = img5, node distance = 2.4cm](img6){\includegraphics[width=0.3\textwidth]{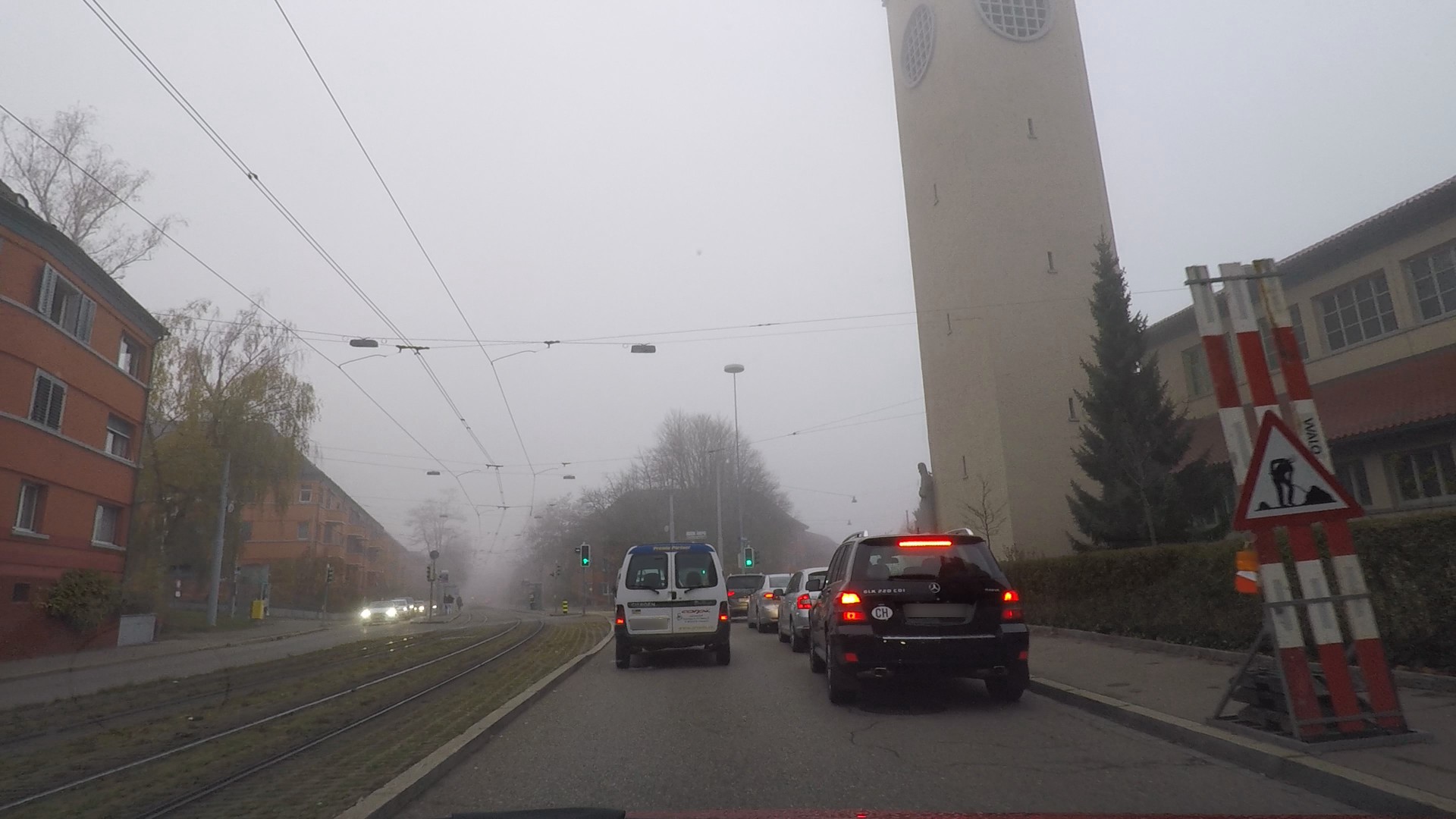}};
        \node[right of = img6, node distance = 4cm](seg6){\includegraphics[width=0.3\textwidth]{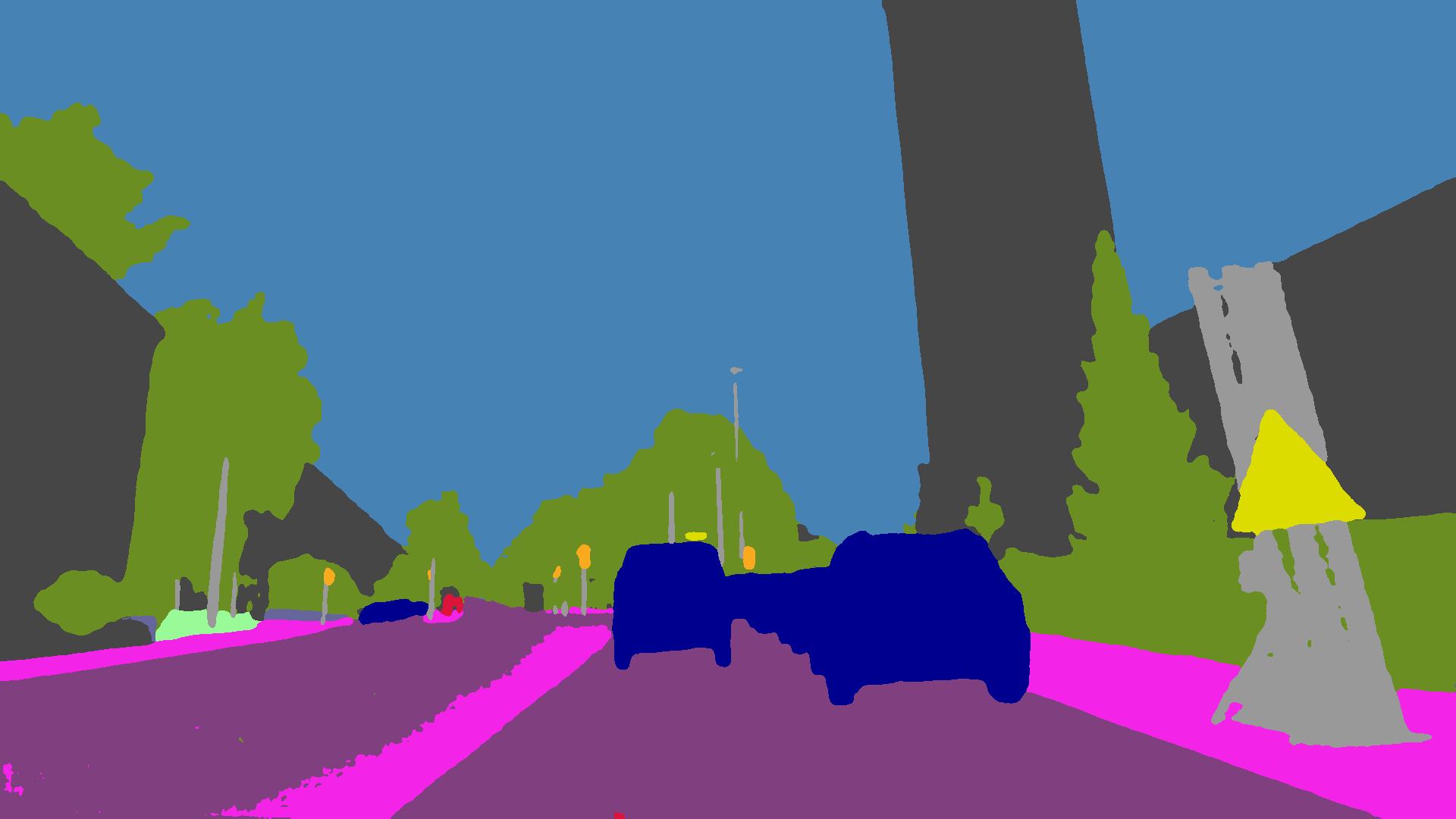}};
        \node[right of = seg6, node distance = 4cm](bbox6){\includegraphics[width=0.3\textwidth]{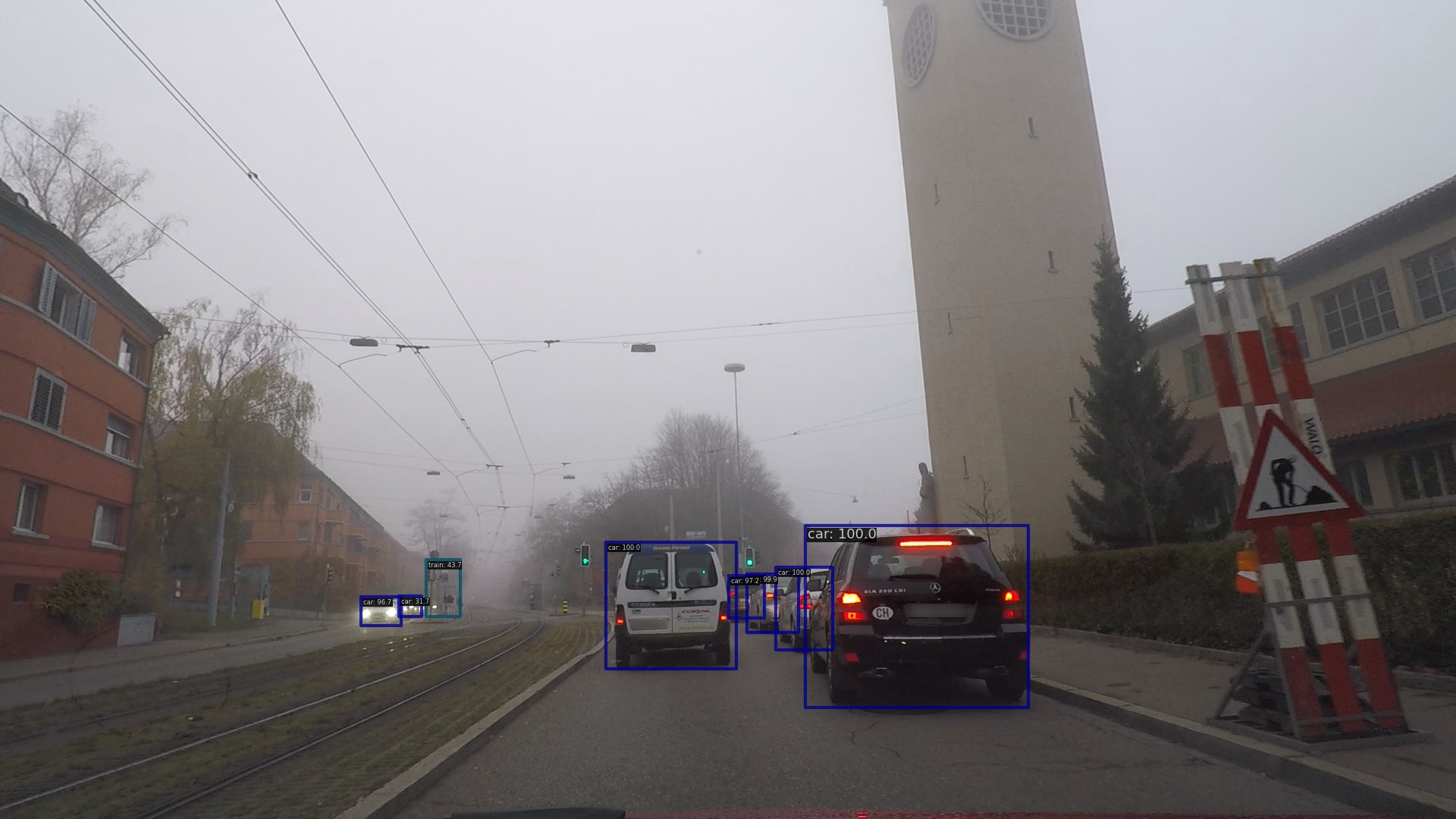}};
    \node[above of = pic1, node distance=1.3cm]{\textbf{RGB Input}};
    \node[above of = pic2, node distance=1.3cm]{\textbf{VLTSeg Prediction}};
    \node[above of = pic3, node distance=1.3cm]{\textbf{VLTDet Prediction}};

    \end{tikzpicture}
    
    \caption{\textbf{Predictions on the ACDC val set $\acdcval$.} Training on $\cstrain$ and evaluation was conducted as described in \cref{sec:testset_eval}. Best viewed digital for the predictions.}
    \label{fig:acdc_sota}
\end{figure*}

\fakeparagraph{ACDC}
We show predictions on the ACDC val set $\acdcval$ in \cref{fig:acdc_sota}. Even though our model has neven seen this domain before during training we can observe that it provides high-quality segmentation maps across challenging adverse weather conditions.

\section{Implementation Details}
\subsection{Training settings}
We provide an extensive list of our hyperparameters and detailed settings in \cref{tab:experimental_settings} to make our experiments reproducible.
For the Mask2Former \cite{cheng2022masked} decoder we used the default settings as provided by the authors. For the \network{EVA-02-S-16} model we interpolated the text encoder output projection in bilinear mode to match the vision encoder output shape of 384. Similar to DenseCLIP \cite{rao2022denseclip} we used FPN projection layers between Encoder and Decoder.
\begin{table}
\centering
  \caption{\textbf{Detailed Experimental Settings}}
  \input{tables/experimental_settings}

  \label{tab:experimental_settings} 
\end{table}

\subsection{Test set evaluation}
\label{sec:testset_eval}
For the evaluation on the Cityscapes \cite{Cordts2016} and ACDC \cite{sakaridis2021acdc} test set we followed the corresponding common practice.\\
For Cityscapes test set evaluation, we first trained VLTSeg on Mapillary for 20k iterations as also done by previous state-of-the-art approaches \cite{wang2023internimage,borse2021hs3, borse2021inverseform}. Afterwards, 40k fine-tuning iterations on the official $\cstrain$ dataset with 2975 images were conducted. In contrast to the other works, no additional data like the coarse annotations were used. Both trainings used 1024 $\times$ 1024 resolution. We used multi-scale evaluation with $[1.0, 1.25, 1.5, 1.75, 2.0, 2.25]$ image ratios and random flip as test-time augmentation during inference. Test set evaluation was done on the full 2048$\times$ 1024 resolution.For the ACDC test set evaluation, we trained for 20k iterations on 1024 $\times$ 1024 crops only on Cityscapes $\cstrain$ without using any ACDC data. The test set inference was similar to Cityscapes with a multi-scale evaluation with $[1.0, 1.25, 1.5, 1.75]$ image ratios and on the full 1920 $\times$ 1080 resolution.

\section{Computational complexity of vision encoders}
We show the GFLOPS and parameters of the vision encoders employed in our experiments in \cref{tab:params_flops}.  We used three variants of the \network{ViT} with base, large, and SAM and a patch size of 16. Moreover, we employed three complexities of EVA: small, base, and large. Due to the architectural modifications, EVA has more GFLOPS than ViT at a similar number of parameters.

\section{Feature Space Analysis}
We conducted a feature space analysis for the other real-world datasets for Cityscapes \cite{cordts2016cityscapes} before and after fine-tuning on the synthetic GTA5 \cite{richter2016playing} dataset. The results are shown in \cref{fig:umap_analysis}.\\
We observe that real-world embeddings are well divided for the real-world target dataset after the synthetic source-only training. That implies that fine-tuning improves the feature space of vision-text alignment by separating classes more clearly and offering strong generalization capabilities across several different real-world domains.

\begin{figure}
\vspace{3mm}
  \centering
  \includegraphics[width=0.9\textwidth]{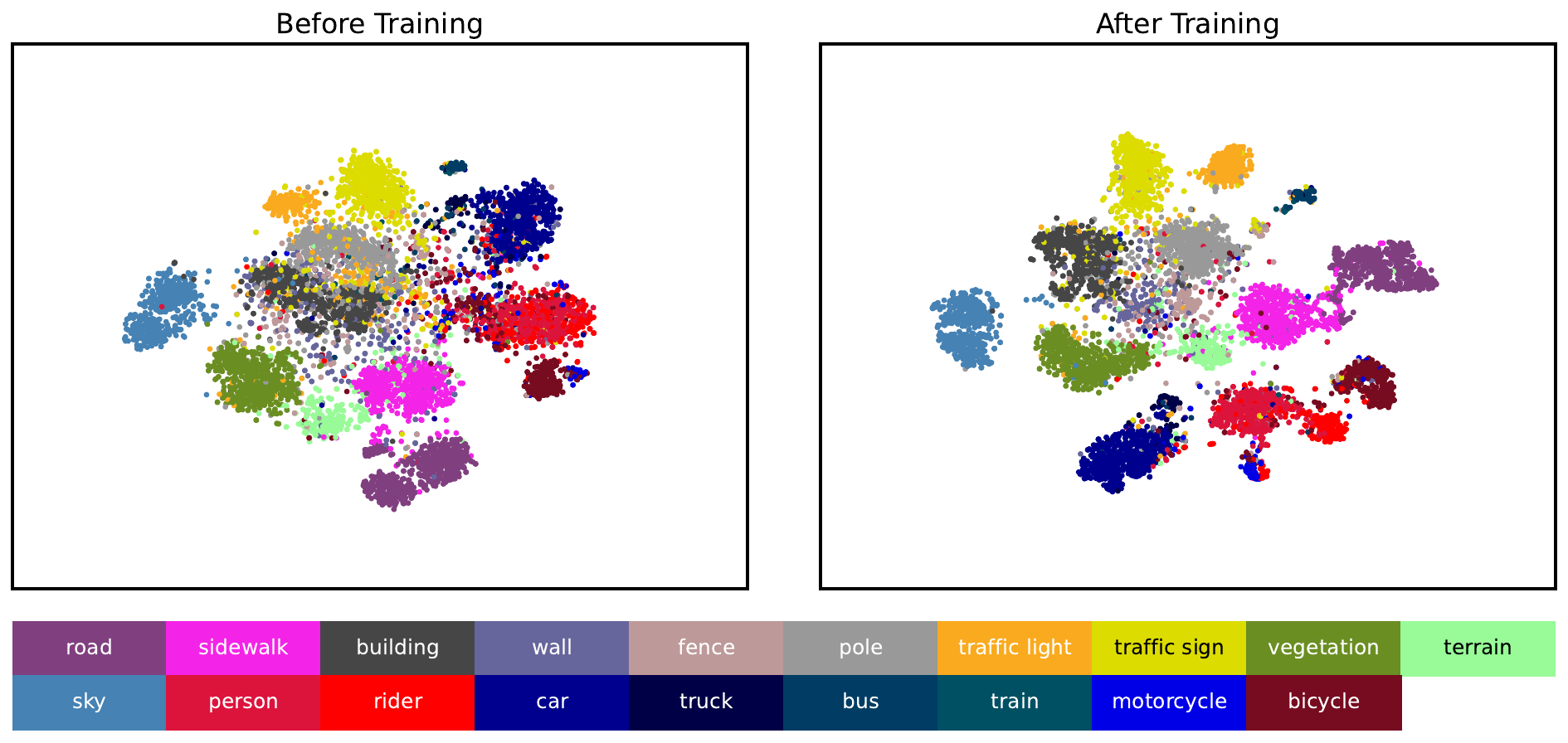}
  \label{fig:umap_analysis}

  \caption{\textbf{t-SNE feature space analysis on the real-world dataset Cityscapes.} We sampled 500 images from the real-world validation set and extracted the visual embeddings of our best performing VLTSeg network. From the plot, we can see that real target class clusters are well separated (right image) after our VLTSeg training on source synthetic GTA5 dataset. Best viewed digitally.}
  \label{fig:umap_analysis}
\end{figure}

%% file: tables/resnet_sota.tex
\extrarowheight=\aboverulesep
    \addtolength{\extrarowheight}{\belowrulesep}
    \aboverulesep=0pt
    \belowrulesep=0pt
    \begin{tabular}{@{}clcccc@{}}
        \toprule
        \multirow{1}{*}{\rot{\makecell[c]{\textbf{}}}} & \textbf{DG Method} &\multicolumn{4}{c}{\textbf{mIoU (\%) on}} \\
        \cmidrule{3-6}
        && $\csteststar$ & $\bddteststar$ & $\mvteststar$ & \raisebox{-4.0pt}{\shortstack{\textbf{DG} \\\textbf{mean}}}  \rule[-1.2ex]{0mm}{3.65ex} \\

        \midrule

        \parbox[t]{2mm}{\multirow{13}{*}{\rotatebox[origin=c]{90}{\textbf{\small{$\src$: GTA5}}}}} &Baseline \cite{Xie2022segformer} & 36.1 & 36.6 & 43.8& 38.8\\
        &IBN-Net$^\circ$~\cite{Pan2018}  & 37.7 & 36.7 & 36.8 & 37.1\\
        &RobustNet$^\circ$~\cite{Choi2021}&  37.3 & 38.7 & 38.1 &38.0\\
        &DRPC~\cite{Yue2019}&  42.5& 38.7& 38.1&39.8\\
        &SW~\cite{pan2019switchable}&  36.1 &36.6&32.6&35.1 \\
        &FSDR~\cite{Huang2021}&  44.8 &41.2&43.4&43.1 \\
        &SAN+SAW~\cite{Peng2022semanticaware}&  45.3 &41.2& 40.8 &42.4\\
       &WEDGE \cite{kim2021wedge}& 45.2&41.1&48.1&44.8 \\
       &GTR \cite{peng2021global}&43.7 &39.6&39.1&40.8 \\
        &SHADE \cite{zhao2022style}&46.7 &43.7&45.5&45.3 \\
        &WildNet$^\circ$~\cite{Lee2022wildnet}  &  {45.8} & {41.7} & {47.1} &44.9\\
        &TLDR~\cite{kim2023texture}  &  {47.6} & {44.9} & {48.8} & {47.1}\\
        &RICA~\cite{sun2023augment}  &  {48.0} & 45.2 & {46.3} & {46.5}\\
        &PASTA~\cite{2023iccv_PASTA}  &  {45.3} & 42.3 & {48.6} & {45.4}\\
        &FAMix \cite{fahes2024simple} & 49.5 & 46.4 & 52.0 & 49.3 \\
        &CLOUDS \cite{benigmim2024collaborating} & \textbf{55.7} & \textbf{49.3} &  \textbf{59.0} & \textbf{54.7}\\
        & DIDEX \cite{niemeijer2024generalization} & 52.4 & 40.9 & 49.2 & 47.5\\
        & DGinStyle \cite{jia2023dginstyle} & 46.9 & 42.8 & 50.2 & 46.6 \\

          &\cellcolor{textcolor!20} VLTSeg-R (Ours) & 49.5 \cellcolor{textcolor!20} & 40.0 \cellcolor{textcolor!20}& 52.4 \cellcolor{textcolor!20}& 47.3 \cellcolor{textcolor!20}\\

        \bottomrule
    \end{tabular}

%% file: tables/sota_ext_synthia.tex
\extrarowheight=\aboverulesep
    \addtolength{\extrarowheight}{\belowrulesep}
    \aboverulesep=0pt
    \belowrulesep=0pt
    \begin{tabular}{@{}cccccc@{}}
        \toprule
        \multirow{1}{*}{\rot{\makecell[c]{\textbf{}}}} & \textbf{DG Method} &\multicolumn{4}{c}{\textbf{mIoU (\%) on}} \\
        \cmidrule{3-6}
        && $\csteststar$ & $\bddteststar$ & $\mvteststar$ & \raisebox{-4.0pt}{\shortstack{\textbf{DG} \\\textbf{mean}}}  \rule[-1.2ex]{0mm}{3.65ex} \\

        \midrule

        \parbox[t]{2mm}{\multirow{6}{*}{\rotatebox[origin=c]{90}{\tiny{\textbf{$\src$: SYNTHIA}}}}} &Baseline \cite{Xie2022segformer}   & 41.4&36.2& 42.4  &40.0\\
        &ReVT \cite{Termoehlen2023arxiv}   & 46.3&40.3& 44.8   &43.8\\
        &CMFormer \cite{bi2023learning}   & 44.6& 33.4& 43.3&40.4\\
        &PromptFormer \cite{gong2023prompting}  & 49.3& -& -&- \\
        &IBAFormer \cite{sun2023ibaformer}  & 50.9& 44.7& 50.6&48.7 \\
        & CLOUDS \cite{benigmim2024collaborating} & 53.4 & 47.0 & 55.8 & 52.1 \\
        & DIDEX \cite{niemeijer2024generalization} & \textbf{59.8} & 47.4 & 59.5 & \textbf{55.6} \\
        
        &VLTSeg \cellcolor{textcolor!20}& 56.8\cellcolor{textcolor!20} & \textbf{51.9} \cellcolor{textcolor!20}& 55.1 \cellcolor{textcolor!20}&54.6 \cellcolor{textcolor!20}\\
        \midrule
        \parbox[t]{2mm}{\multirow{2}{*}{\rotatebox[origin=c]{90}{\tiny{\textbf{$\src$: US}}}}} &Baseline \cite{Xie2022segformer}   & 63.3&41.3& 58.8  &54.5\\
        &\cellcolor{textcolor!20} VLTSeg & \textbf{70.1} \cellcolor{textcolor!20}& \textbf{52.4} \cellcolor{textcolor!20}& \textbf{63.0}\cellcolor{textcolor!20} &\textbf{61.8} \cellcolor{textcolor!20}\\
        \bottomrule
    \end{tabular}

%% file: tables/decoder_ablation.tex
\extrarowheight=\aboverulesep
    \addtolength{\extrarowheight}{\belowrulesep}
    \aboverulesep=0pt
    \belowrulesep=0pt
    \resizebox{0.48\columnwidth}{!}{
    \begin{tabular}{c|ccccc}
    \textbf{Decoder} & \multicolumn{5}{c}{\textbf{mIoU in \%}}\\
    \hline
    & $\csteststar$ & $\bddteststar$&$\mvteststar$ &$\acdcteststar$ &\raisebox{-4.0pt}{\shortstack{\textbf{DG} \\\textbf{mean}}}  \rule[-1.2ex]{0mm}{3.65ex}  \\
    \cline{2-6}
    Semantic FPN \cite{loshchilov2018decoupled} & {60.5} & {57.5} & {62.2}&55.9 &{59.0} \\ 
     
    Segformer \cite{Xie2022segformer} & {58.2} & {55.0} & {61.4}& 56.0 &{57.7}\\
    DAFormer \cite{Hoyer2022daformer} & {64.0} & {57.9} & {65.0}& 59.3 &{61.6}\\
    Mask2Former \cite{cheng2022masked} &\textbf{65.3} & \textbf{58.3} & \textbf{66.0} & \textbf{62.6} & \textbf{63.1}\\
    \end{tabular}}

%% file: tables/params_flops.tex
\extrarowheight=\aboverulesep
    \addtolength{\extrarowheight}{\belowrulesep}
    \aboverulesep=0pt
    \belowrulesep=0pt
    \resizebox{0.48\columnwidth}{!}{
    \begin{tabular}{c|c |c}
    \textbf{Encoder} & \textbf{Parameters} &\textbf{GFLOPS}\\
    \hline
    \network{ViT-B-16} & 87 & 88\\
    \network{ViT-L-16} & 304 & 311\\
    \network{SAM-L-16} & 308 & 397\\
    \hline
    \network{EVA-02-S-16} & 22 & 32\\
    \network{EVA-02-B-16} & 86 & 107\\
    \network{EVA-02-L-14} & 303 & 508\\
    
    \bottomrule
    \end{tabular}}

%% file: fig/sens_analysis.tex
\definecolor{textcolor}{RGB}{70,170,34}

\pgfplotstableread[row sep=\\,col sep=&]{
   ablation& value1 & value2 & value3 & meta1 & meta2 & meta3 \\
    Batch Size    &  54.59 & 54.2 & 54.03 & 8 & 12 & 16  \\
    Learning Rate &   29.7  & 54.03  & 49.4  & 1e-5 & 1e-4& 1e-3  \\
    Optimizer   & 13.72 & 0.07 & 54.03 & SGD & Adam & AdamW \\
    }\mydata
\begin{tikzpicture}
    \begin{axis}[
            ybar=.3cm,
            width=1.0\columnwidth,
            height=0.4\columnwidth,
            enlarge x limits=0.3,
            symbolic x coords={Batch Size, Learning Rate, Optimizer},
            xtick=data,
            nodes near coords,
            nodes near coords align={vertical},
            bar width = 0.6cm,
            ymin=0,ymax=70,
            ylabel={DG Mean (\%)},
        ]
        \addplot+[nodes near coords style={font=\footnotesize}, point meta=explicit symbolic, color=textcolor!80]table[x=ablation,y=value1, meta=value1]{\mydata};
        \addplot+[nodes near coords style={font=\footnotesize},  point meta=explicit symbolic, color=textcolor!60]table[x=ablation,y=value2, meta=value2]{\mydata};
        \addplot+[nodes near coords style={font=\footnotesize}, point meta=explicit symbolic, color=textcolor!40]table[x=ablation,y=value3, meta=value3]{\mydata};
        
    \end{axis}

\node[] at (1.1,0.3){8};
\node[] at (2.0,0.3){12};
\node[] at (2.9,0.3){16};

\node[rotate=90] at (4.4,0.4){\footnotesize $10^{-3}$};
\node[rotate=90] at (5.3,0.4){\footnotesize $10^{-4}$};
\node[rotate=90] at (6.2,0.4){\footnotesize $10^{-5}$};

\node[rotate=90] at (7.7,1.4){\footnotesize SGD};
\node[rotate=90] at (8.5,1.4){\footnotesize Adam};
\node[rotate=90] at (9.5,1.4){\footnotesize AdamW};
\end{tikzpicture}

%% file: tables/experimental_settings.tex
  \resizebox{0.6\textwidth}{!}{
  \begin{tabular}{c|c|c|c}
  \toprule
    \multicolumn{2}{c|}{\textbf{Hyperparameter}} & \textbf{Synthetic-to-Real} & \textbf{Real-to-Real} \\
    \hline
    \multicolumn{2}{c|}{crop size} & 512 & 1024\\
    \multicolumn{2}{c|}{stride size} & 426 & 768\\
    \multicolumn{2}{c|}{iterations} & 5k & 20k\\
    \multicolumn{2}{c|}{batch size} & 16 & 8\\
    \hline
    \multirow{6}{*}{\textbf{Optimizer}} &type & \multicolumn{2}{c}{AdamW} \\
    &default lr & \multicolumn{2}{c}{1e-04} \\
    &backbone lr & \multicolumn{2}{c}{1e-05}\\
    &\text{weight\_decay} & \multicolumn{2}{c}{0.05}\\
    &eps &\multicolumn{2}{c}{1e-08}\\
     & betas&\multicolumn{2}{c}{(0.9, 0.999)}\\
    \hline
    \multirow{4}{*}{\text{\textbf{LR Schedule}}} &type & \multicolumn{2}{c}{PolyLR}\\
    & \text{eta\_min} &\multicolumn{2}{c}{0}\\
    &power & \multicolumn{2}{c}{0.9}\\
    &begin & \multicolumn{2}{c}{500}\\
    \hline
    \multirow{5}{*}{\textbf{Text}} &context length & \multicolumn{2}{c}{13} \\
     &embedding size & \multicolumn{2}{c}{768} \\
     &transformer heads & \multicolumn{2}{c}{12} \\
     &transformer width & \multicolumn{2}{c}{768} \\
     &transformer layers & \multicolumn{2}{c}{12} \\
    \hline
    \multirow{6}{*}{\textbf{Encoder}}
    & \text{in\_channels} &\multicolumn{2}{c}{3}\\
    &patch size & \multicolumn{2}{c}{14}\\
    &embedding size & \multicolumn{2}{c}{1024}\\
    &depth & \multicolumn{2}{c}{24} \\
    &indices & \multicolumn{2}{c}{[9, 14, 19, 23]}\\
    &\text{output\_dim} &\multicolumn{2}{c}{ 768 }\\

    \hline
     \multirow{8}{*}{ \textbf{Decoder}} &{\text{in\_channels}} &\multicolumn{2}{c}{[1024,1024,1024,1024]} \\
    &{stride} & \multicolumn{2}{c}{[4, 8, 16, 32]}\\
     &{\text{feat\_channels} }&\multicolumn{2}{c}{256}\\
     &{\text{out\_channels} }&\multicolumn{2}{c}{256}\\
     &{\text{num\_classes} }&\multicolumn{2}{c}{19}\\
     &{\text{num\_queries} }&\multicolumn{2}{c}{100}\\
     &{\text{num\_transformer\_feat\_level} }&\multicolumn{2}{c}{3}\\
    &{\text{positional\_encoding}} & \multicolumn{2}{c}{128} \\
\bottomrule
  \end{tabular}
  }

%% file: main.bbl
\begin{thebibliography}{100}
\providecommand{\url}[1]{\texttt{#1}}
\providecommand{\urlprefix}{URL }
\providecommand{\doi}[1]{https://doi.org/#1}

\bibitem{Alayrac2022}
Alayrac, J.B., Donahue, J., Luc, P., Miech, A., Barr, I., Hasson, Y., Lenc, K., Mensch, A., Millican, K., Reynolds, M., Ring, R., Rutherford, E., Cabi, S., Han, T., Gong, Z., Samangooei, S., Monteiro, M., Menick, J.L., Borgeaud, S., Brock, A., Nematzadeh, A., Sharifzadeh, S., Bi\'{n}kowski, M.a., Barreira, R., Vinyals, O., Zisserman, A., Simonyan, K.: Flamingo: a visual language model for few-shot learning. In: Proc.\ of NeurIPS (2022)

\bibitem{bao2022}
Bao, H., Wang, W., Dong, L., Wei, F.: Vl-beit: Generative vision-language pretraining (2022)

\bibitem{benigmim2024collaborating}
Benigmim, Y., Roy, S., Essid, S., Kalogeiton, V., Lathuili{\`e}re, S.: Collaborating foundation models for domain generalized semantic segmentation. In: Proceedings of the IEEE/CVF Conference on Computer Vision and Pattern Recognition. pp. 3108--3119 (2024)

\bibitem{bi2023learning}
Bi, Q., You, S., Gevers, T.: {Learning Content-enhanced Mask Transformer for Domain Generalized Urban-Scene Segmentation}. arXiv:2307.00371 pp. 1--18 (2023)

\bibitem{bi2024good}
Bi, Q., Zhou, B., Yi, J., Ji, W., Zhan, H., Xia, G.S.: Good: Towards domain generalized orientated object detection. arXiv preprint arXiv:2402.12765  (2024)

\bibitem{borse2021hs3}
Borse, S., Cai, H., Zhang, Y., Porikli, F.: Hs3: Learning with proper task complexity in hierarchically supervised semantic segmentation. arXiv preprint arXiv:2111.02333  (2021)

\bibitem{borse2021inverseform}
Borse, S., Wang, Y., Zhang, Y., Porikli, F.: Inverseform: A loss function for structured boundary-aware segmentation. In: Proc.\ of CVPR. pp. 5901--5911 (2021)

\bibitem{caron2021emerging}
Caron, M., Touvron, H., Misra, I., J{\'e}gou, H., Mairal, J., Bojanowski, P., Joulin, A.: Emerging properties in self-supervised vision transformers. In: Proc.\ of ICCV. pp. 9650--9660 (2021)

\bibitem{2023iccv_PASTA}
Chattopadhyay*, P., Sarangmath*, K., Vijaykumar, V., Hoffman, J.: Pasta: Proportional amplitude spectrum training augmentation for syn-to-real domain generalization. In: Proc. of ICCV (2023)

\bibitem{chen2020harmonizing}
Chen, C., Zheng, Z., Ding, X., Huang, Y., Dou, Q.: Harmonizing transferability and discriminability for adapting object detectors. In: Proceedings of the IEEE/CVF Conference on Computer Vision and Pattern Recognition. pp. 8869--8878 (2020)

\bibitem{chen2019mmdetection}
Chen, K., Wang, J., Pang, J., Cao, Y., Xiong, Y., Li, X., Sun, S., Feng, W., Liu, Z., Xu, J., et~al.: Mmdetection: Open mmlab detection toolbox and benchmark. arXiv preprint arXiv:1906.07155  (2019)

\bibitem{chen2020simple}
Chen, T., Kornblith, S., Norouzi, M., Hinton, G.: A simple framework for contrastive learning of visual representations. In: Proc.\ of ICML. pp. 1597--1607. PMLR (2020)

\bibitem{chenempirical}
Chen, X., Xie, S., He, K.: An empirical study of training self-supervised vision transformers. in 2021 ieee. In: Proc.\ of ICCV. pp. 9620--9629

\bibitem{chen2022vision}
Chen, Z., Duan, Y., Wang, W., He, J., Lu, T., Dai, J., Qiao, Y.: Vision transformer adapter for dense predictions. arXiv preprint arXiv:2205.08534  (2022)

\bibitem{cheng2022masked}
Cheng, B., Misra, I., Schwing, A.G., Kirillov, A., Girdhar, R.: Masked-attention mask transformer for universal image segmentation. In: Proc.\ of CVPR. pp. 1290--1299 (2022)

\bibitem{Choi2021}
Choi, S., Jung, S., Yun, H., Kim, J.T., Kim, S., Choo, J.: {RobustNet: Improving Domain Generalization in Urban-Scene Segmentation via Instance Selective Whitening}. In: Proc.\ of CVPR. pp. 11580--11590 (Jun 2021)

\bibitem{2023mmpretrain}
Contributors, M.: Openmmlab's pre-training toolbox and benchmark. \url{https://github.com/open-mmlab/mmpretrain} (2023)

\bibitem{contributors2020mmsegmentation}
Contributors, M.: Mmsegmentation: Openmmlab semantic segmentation toolbox and benchmark (2020)

\bibitem{cordts2016cityscapes}
Cordts, M., Omran, M., Ramos, S., Rehfeld, T., Enzweiler, M., Benenson, R., Franke, U., Roth, S., Schiele, B.: The cityscapes dataset for semantic urban scene understanding (2016)

\bibitem{Cordts2016}
Cordts, M., Omran, M., Ramos, S., Rehfeld, T., Enzweiler, M., Benenson, R., Franke, U., Roth, S., Schiele, B.: {The Cityscapes Dataset for Semantic Urban Scene Understanding}. In: Proc.\ of CVPR. pp. 3213--3223 (Jun 2016)

\bibitem{Cubuk2022}
Cubuk, E.D., Zoph, B., Shlens, J., Le, Q.: Randaugment: Practical automated data augmentation with a reduced search space. In: Proc. of NeurIPS. vol.~33 (2020)

\bibitem{ding2023hgformer}
Ding, J., Xue, N., Xia, G.S., Schiele, B., Dai, D.: Hgformer: Hierarchical grouping transformer for domain generalized semantic segmentation. In: Proc.\ of CVPR. pp. 15413--15423 (2023)

\bibitem{Dosovitskiy2021}
Dosovitskiy, A., Beyer, L., Kolesnikov, A., Weissenborn, D., Zhai, X., Unterthiner, T., Dehghani, M., Minderer, M., Heigold, G., Gelly, S., Uszkoreit, J., Houlsby, N.: {An Image is Worth 16x16 Words: Transformers for Image Recognition at Scale}. In: Proc.\ of {ICLR}. pp. 1--21 (May 2021)

\bibitem{dosovitskiy2016inverting}
Dosovitskiy, A., Brox, T.: Inverting visual representations with convolutional networks. In: Proceedings of the IEEE conference on computer vision and pattern recognition. pp. 4829--4837 (2016)

\bibitem{fahes2024simple}
Fahes, M., Vu, T.H., Bursuc, A., P{\'e}rez, P., de~Charette, R.: A simple recipe for language-guided domain generalized segmentation. In: Proceedings of the IEEE/CVF Conference on Computer Vision and Pattern Recognition. pp. 23428--23437 (2024)

\bibitem{fahes2023poda}
Fahes, M., Vu, T.H., Bursuc, A., Pérez, P., de~Charette, R.: P{\o}da: Prompt-driven zero-shot domain adaptation (2023)

\bibitem{fang2023eva}
Fang, Y., Sun, Q., Wang, X., Huang, T., Wang, X., Cao, Y.: Eva-02: A visual representation for neon genesis. arXiv preprint arXiv:2303.11331  (2023)

\bibitem{feng2021rethinking}
Feng, Y., Jiang, J., Tang, M., Jin, R., Gao, Y.: Rethinking supervised pre-training for better downstream transferring. In: Proc.\ of ICLR (2021)

\bibitem{geng2022multimodal}
Geng, X., Liu, H., Lee, L., Schuurmans, D., Levine, S., Abbeel, P.: Multimodal masked autoencoders learn transferable representations (2022)

\bibitem{goldblum2023battle}
Goldblum, M., Souri, H., Ni, R., Shu, M., Prabhu, V.U., Somepalli, G., Chattopadhyay, P., Ibrahim, M., Bardes, A., Hoffman, J., Chellappa, R., Wilson, A.G., Goldstein, T.: Battle of the backbones: A large-scale comparison of pretrained models across computer vision tasks. In: Proc.\ of NeurIPS Datasets and Benchmarks Track (2023), \url{https://openreview.net/forum?id=1yOnfDpkVe}

\bibitem{gomez2023all}
G{\'o}mez, J.L., Silva, M., Seoane, A., Borr{\'a}s, A., Noriega, M., Ros, G., Iglesias-Guitian, J.A., L{\'o}pez, A.M.: All for one, and one for all: Urbansyn dataset, the third musketeer of synthetic driving scenes. arXiv preprint arXiv:2312.12176  (2023)

\bibitem{gong2023prompting}
Gong, R., Danelljan, M., Sun, H., Mangas, J.D., Van~Gool, L.: {Prompting Diffusion Representations for Cross-Domain Semantic Segmentation}. arXiv:2307.02138 pp. 1--17 (2023)

\bibitem{grill2020bootstrap}
Grill, J.B., Strub, F., Altch{\'e}, F., Tallec, C., Richemond, P., Buchatskaya, E., Doersch, C., Avila~Pires, B., Guo, Z., Gheshlaghi~Azar, M., et~al.: Bootstrap your own latent-a new approach to self-supervised learning. In Proc.\ of NeurIPS  \textbf{33},  21271--21284 (2020)

\bibitem{he2020momentum}
He, K., Fan, H., Wu, Y., Xie, S., Girshick, R.: Momentum contrast for unsupervised visual representation learning. In: Proc.\ of CVPR. pp. 9729--9738 (2020)

\bibitem{he2019rethinking}
He, K., Girshick, R., Doll{\'a}r, P.: Rethinking imagenet pre-training. In: Proc.\ of ICCV. pp. 4918--4927 (2019)

\bibitem{he2016deep}
He, K., Zhang, X., Ren, S., Sun, J.: Deep residual learning for image recognition. In: Proc.\ of CVPR. pp. 770--778 (2016)

\bibitem{He_2023_CVPR}
He, W., Jamonnak, S., Gou, L., Ren, L.: Clip-s4: Language-guided self-supervised semantic segmentation. In: Proc.\ of CVPR. pp. 11207--11216 (June 2023)

\bibitem{Hendrycks2022pixmix}
Hendrycks, D., Zou, A., Mazeika, M., Tang, L., Li, B., Song, D., Steinhardt, J.: {PixMix: Dreamlike Pictures Comprehensively Improve Safety Measures}. In: Proc.\ of CVPR. pp. 16783--16792. New Orleans, LA, USA (Jun 2022)

\bibitem{hoffman2018cycada}
Hoffman, J., Tzeng, E., Park, T., Zhu, J.Y., Isola, P., Saenko, K., Efros, A.A., Darrell, T.: {{CyCADA}: Cycle-Consistent Adversarial Domain Adaptation}. In: Proc.\ of ICML. pp. 1989--1998 (Jul 2018)

\bibitem{Hoyer2022daformer}
Hoyer, L., Dai, D., Van~Gool, L.: {DAFormer: Improving Network Architectures and Training Strategies for Domain-Adaptive Semantic Segmentation}. In: Proc.\ of CVPR. pp. 9924--9935 (Jun 2022)

\bibitem{hoyer2022hrda}
Hoyer, L., Dai, D., Van~Gool, L.: {HRDA: Context-Aware High-Resolution Domain-Adaptive Semantic Segmentation}. In: Proc.\ of ECCV. pp. 372--391 (Oct 2022)

\bibitem{hoyer2023domain}
Hoyer, L., Dai, D., Van~Gool, L.: {Domain Adaptive and Generalizable Network Architectures and Training Strategies for Semantic Image Segmentation}. arXiv:2304.13615 pp. 1--15 (Apr 2023)

\bibitem{hoyer2023mic}
Hoyer, L., Dai, D., Wang, H., Van~Gool, L.: {MIC: Masked Image Consistency for Context-Enhanced Domain Adaptation}. In: Proc.\ of CVPR. pp. 11721--11732 (Jun 2023)

\bibitem{Huang2021}
Huang, J., Guan, D., Xiao, A., Lu, S.: {FSDR: Frequency Space Domain Randomization for Domain Generalization}. In: Proc.\ of CVPR. pp. 6891--6902 (Jun 2021)

\bibitem{huang2021fsdr}
Huang, J., Guan, D., Xiao, A., Lu, S.: Fsdr: Frequency space domain randomization for domain generalization. In: Proc.\ of CVPR. pp. 6891--6902 (2021)

\bibitem{huh2016makes}
Huh, M., Agrawal, P., Efros, A.A.: What makes imagenet good for transfer learning? arXiv preprint arXiv:1608.08614  (2016)

\bibitem{jia2021scaling}
Jia, C., Yang, Y., Xia, Y., Chen, Y.T., Parekh, Z., Pham, H., Le, Q., Sung, Y.H., Li, Z., Duerig, T.: Scaling up visual and vision-language representation learning with noisy text supervision. In: Proc.\ of ICML. pp. 4904--4916. PMLR (2021)

\bibitem{jia2023dginstyle}
Jia, Y., Hoyer, L., Huang, S., Wang, T., Van~Gool, L., Schindler, K., Obukhov, A.: Dginstyle: Domain-generalizable semantic segmentation with image diffusion models and stylized semantic control. In: Synthetic Data for Computer Vision Workshop@ CVPR 2024 (2023)

\bibitem{kerssies2024benchmark}
Kerssies, T., De~Geus, D., Dubbelman, G.: How to benchmark vision foundation models for semantic segmentation? In: Proceedings of the IEEE/CVF Conference on Computer Vision and Pattern Recognition. pp. 1162--1171 (2024)

\bibitem{kim2021wedge}
Kim, N., Son, T., Lan, C., Zeng, W., Kwak, S.: {WEDGE: Web-Image Assisted Domain Generalization for Semantic Segmentation}. arXiv:2109.14196 pp. 1--14 (Sep 2021)

\bibitem{kim2023texture}
Kim, S., Kim, D.h., Kim, H.: Texture learning domain randomization for domain generalized segmentation. Proc. of ICCV  (2023)

\bibitem{kirillov2019panoptic}
Kirillov, A., Girshick, R., He, K., Doll{\'a}r, P.: Panoptic feature pyramid networks. In: Proc.\ of CVPR. pp. 6399--6408 (2019)

\bibitem{kirillov2023segment}
Kirillov, A., Mintun, E., Ravi, N., Mao, H., Rolland, C., Gustafson, L., Xiao, T., Whitehead, S., Berg, A.C., Lo, W.Y., et~al.: Segment anything. arXiv preprint arXiv:2304.02643  (2023)

\bibitem{klingner2022unsupervised}
Klingner, M., Term{\"o}hlen, J.A., Ritterbach, J., Fingscheidt, T.: Unsupervised batchnorm adaptation (ubna): A domain adaptation method for semantic segmentation without using source domain representations. In: Proc.\ of WACV. pp. 210--220 (2022)

\bibitem{kornblith2019better}
Kornblith, S., Shlens, J., Le, Q.V.: Do better imagenet models transfer better? In: Proc.\ of CVPR. pp. 2661--2671 (2019)

\bibitem{Lee2022wildnet}
Lee, S., Seong, H., Lee, S., Kim, E.: {WildNet: Learning Domain Generalized Semantic Segmentation from the Wild}. In: Proc.\ of CVPR. pp. 9936--9946 (Jun 2022)

\bibitem{li2023otter}
Li, B., Zhang, Y., Chen, L., Wang, J., Yang, J., Liu, Z.: Otter: A multi-modal model with in-context instruction tuning. arXiv preprint arXiv:2305.03726  (2023)

\bibitem{li2024prompt}
Li, D., Wu, A., Wang, Y., Han, Y.: Prompt-driven dynamic object-centric learning for single domain generalization. In: Proc. of CVPR. pp. 17606--17615 (2024)

\bibitem{LiMGH2022}
Li, Y., Mao, H., Girshick, R.B., He, K.: Exploring plain vision transformer backbones for object detection. In: Proc. of ECCV (2022)

\bibitem{Liang_2023_CVPR}
Liang, F., Wu, B., Dai, X., Li, K., Zhao, Y., Zhang, H., Zhang, P., Vajda, P., Marculescu, D.: Open-vocabulary semantic segmentation with mask-adapted clip. In: Proc.\ of CVPR. pp. 7061--7070 (June 2023)

\bibitem{loshchilov2018decoupled}
Loshchilov, I., Hutter, F.: Decoupled weight decay regularization. In: Proc.\ of ICLR (2018)

\bibitem{mattolin2023confmix}
Mattolin, G., Zanella, L., Ricci, E., Wang, Y.: Confmix: Unsupervised domain adaptation for object detection via confidence-based mixing. In: Proceedings of the IEEE/CVF Winter Conference on Applications of Computer Vision. pp. 423--433 (2023)

\bibitem{michaelis2019benchmarking}
Michaelis, C., Mitzkus, B., Geirhos, R., Rusak, E., Bringmann, O., Ecker, A.S., Bethge, M., Brendel, W.: Benchmarking robustness in object detection: Autonomous driving when winter is coming. arXiv preprint arXiv:1907.07484  (2019)

\bibitem{neuhold2017mapillary}
Neuhold, G., Ollmann, T., Rota~Bulo, S., Kontschieder, P.: The mapillary vistas dataset for semantic understanding of street scenes. In: Proc.\ of ICCV. pp. 4990--4999 (2017)

\bibitem{niemeijer2024generalization}
Niemeijer, J., Schwonberg, M., Term{\"o}hlen, J.A., Schmidt, N.M., Fingscheidt, T.: Generalization by adaptation: Diffusion-based domain extension for domain-generalized semantic segmentation. In: Proceedings of the IEEE/CVF Winter Conference on Applications of Computer Vision. pp. 2830--2840 (2024)

\bibitem{oquab2023dinov2}
Oquab, M., Darcet, T., Moutakanni, T., Vo, H., Szafraniec, M., Khalidov, V., Fernandez, P., Haziza, D., Massa, F., El-Nouby, A., et~al.: Dinov2: Learning robust visual features without supervision. arXiv preprint arXiv:2304.07193  (2023)

\bibitem{Pan2018}
Pan, X., Luo, P., Shi, J., Tang, X.: {Two at Once: Enhancing Learning and Generalization Capacities via IBN-Net}. In: Proc.\ of ECCV. pp. 464--479 (Sep 2018)

\bibitem{pan2019switchable}
Pan, X., Zhan, X., Shi, J., Tang, X., Luo, P.: {Switchable Whitening for Deep Representation Learning}. In: Proc.\ of ICCV. pp. 1863--1871 (Oct 2019)

\bibitem{Peng2022semanticaware}
Peng, D., Lei, Y., Hayat, M., Guo, Y., Li, W.: {Semantic-Aware Domain Generalized Segmentation}. In: Proc.\ of CVPR. pp. 2594--2605 (Jun 2022)

\bibitem{peng2021global}
Peng, D., Lei, Y., Liu, L., Zhang, P., Liu, J.: {Global and Local Texture Randomization for Synthetic-to-Real Semantic Segmentation}. \textnormal{In} IEEE TIP  \textbf{30},  6594--6608 (2021)

\bibitem{peng2022beit}
Peng, Z., Dong, L., Bao, H., Ye, Q., Wei, F.: Beit v2: Masked image modeling with vector-quantized visual tokenizers. arXiv preprint arXiv:2208.06366  (2022)

\bibitem{kosmos-2}
Peng, Z., Wang, W., Dong, L., Hao, Y., Huang, S., Ma, S., Wei, F.: Kosmos-2: Grounding multimodal large language models to the world. ArXiv  \textbf{abs/2306} (2023)

\bibitem{radford2021learning}
Radford, A., Kim, J.W., Hallacy, C., Ramesh, A., Goh, G., Agarwal, S., Sastry, G., Askell, A., Mishkin, P., Clark, J., et~al.: Learning transferable visual models from natural language supervision. In: Proc.\ of ICML. pp. 8748--8763. PMLR (2021)

\bibitem{rao2022denseclip}
Rao, Y., Zhao, W., Chen, G., Tang, Y., Zhu, Z., Huang, G., Zhou, J., Lu, J.: Denseclip: Language-guided dense prediction with context-aware prompting. In: Proc.\ of CVPR. pp. 18082--18091 (2022)

\bibitem{richter2016playing}
Richter, S.R., Vineet, V., Roth, S., Koltun, V.: Playing for data: Ground truth from computer games. In: Proc.\ of ECCV. pp. 102--118. Springer (2016)

\bibitem{ridnik2021imagenet}
Ridnik, T., Ben-Baruch, E., Noy, A., Zelnik-Manor, L.: Imagenet-21k pretraining for the masses. arXiv preprint arXiv:2104.10972  (2021)

\bibitem{rombach2021highresolution}
Rombach, R., Blattmann, A., Lorenz, D., Esser, P., Ommer, B.: High-resolution image synthesis with latent diffusion models (2021)

\bibitem{rombach2022invertible}
Rombach, R., Esser, P., Blattmann, A., Ommer, B.: Invertible neural networks for understanding semantics of invariances of cnn representations. In: Deep Neural Networks and Data for Automated Driving: Robustness, Uncertainty Quantification, and Insights Towards Safety, pp. 197--224. Springer International Publishing Cham (2022)

\bibitem{ros2016synthia}
Ros, G., Sellart, L., Materzynska, J., Vazquez, D., Lopez, A.M.: The synthia dataset: A large collection of synthetic images for semantic segmentation of urban scenes. In: Proc.\ of CVPR. pp. 3234--3243 (2016)

\bibitem{sakaridis2023condition}
Sakaridis, C., Bruggemann, D., Yu, F., Van~Gool, L.: Condition-invariant semantic segmentation. arXiv preprint arXiv:2305.17349  (2023)

\bibitem{sakaridis2021acdc}
Sakaridis, C., Dai, D., Van~Gool, L.: {ACDC: The Adverse Conditions Dataset with Correspondences for Semantic Driving Scene Understanding}. In: Proc.\ of ICCV. pp. 10765--10775 (Oct 2021)

\bibitem{schuhmann2022laionb}
Schuhmann, C., Beaumont, R., Vencu, R., Gordon, C.W., Wightman, R., Cherti, M., Coombes, T., Katta, A., Mullis, C., Wortsman, M., Schramowski, P., Kundurthy, S.R., Crowson, K., Schmidt, L., Kaczmarczyk, R., Jitsev, J.: {LAION}-5b: An open large-scale dataset for training next generation image-text models. In: Proc.\ of NeurIPS Datasets and Benchmarks Track (2022), \url{https://openreview.net/forum?id=M3Y74vmsMcY}

\bibitem{schwonberg2023augmentation}
Schwonberg, M., El~Bouazati, F., Schmidt, N.M., Gottschalk, H.: {Augmentation-Based Domain Generalization for Semantic Segmentation}. In: Proc.\ of IV - Workshops. pp.~1--8 (Jun 2023)

\bibitem{Schwonberg2023Survey}
Schwonberg, M., Niemeijer, J., Termöhlen, J.A., Schäfer, J.P., Schmidt, N.M., Gottschalk, H., Fingscheidt, T.: {Survey on Unsupervised Domain Adaptation for Semantic Segmentation for Visual Perception in Automated Driving}. IEEE Access  \textbf{11},  54296--54336 (May 2023)

\bibitem{shen2022k}
Shen, S., Li, C., Hu, X., Xie, Y., Yang, J., Zhang, P., Rohrbach, A., Gan, Z., Wang, L., Yuan, L., et~al.: K-lite: Learning transferable visual models with external knowledge. In: Proc.\ of NeurIPS (2022)

\bibitem{Shi_2023_ICCV}
Shi, C., Yang, S.: Edadet: Open-vocabulary object detection using early dense alignment. In: Proc.\ of ICCV (October 2023)

\bibitem{singh2022}
Singh, A., Hu, R., Goswami, V., Couairon, G., Galuba, W., Rohrbach, M., Kiela, D.: {FLAVA:} {A} foundational language and vision alignment model. In: Proc.\ of CVPR (2022)

\bibitem{sun2023ibaformer}
Sun, Q., Chen, H., Zheng, M., Wu, Z., Felsberg, M., Tang, Y.: {IBAFormer: Intra-batch Attention Transformer for Domain Generalized Semantic Segmentation}. arXiv:2309.06282 pp. 1--10 (Sep 2023)

\bibitem{sun2023augment}
Sun, Q., Melnyk, P., Felsberg, M., Tang, Y.: {Augment Features Beyond Color for Domain Generalized Segmentation}. arXiv:2307.01703 pp. 1--10 (Jul 2023)

\bibitem{EVA-CLIP}
Sun, Q., Fang, Y., Wu, L., Wang, X., Cao, Y.: Eva-clip: Improved training techniques for clip at scale. arXiv preprint arXiv:2303.15389  (2023)

\bibitem{tao2020hierarchical}
Tao, A., Sapra, K., Catanzaro, B.: Hierarchical multi-scale attention for semantic segmentation. arXiv preprint arXiv:2005.10821  (2020)

\bibitem{Termoehlen2023arxiv}
Term{\"o}hlen, J.A., Bartels, T., Fing\-scheidt, T.: {A Re-Parameterized Vision Transformer (ReVT) for Domain-Generalized Semantic Segmentation}. arXiv:2308.13331 pp. 1--18 (Aug 2023)

\bibitem{Termoehlen2023}
Term{\"o}hlen, J.A., Bartels, T., Fing\-scheidt, T.: {A Re-Parameterized Vision Transformer (ReVT) for Domain-Generalized Semantic Segmentation}. In: Proc.\ of ICCV - Workshops. pp. 1--10. Paris, France (Oct 2023)

\bibitem{touvron2022deit}
Touvron, H., Cord, M., J{\'e}gou, H.: Deit iii: Revenge of the vit. In: Proc.\ of ECCV (2022)

\bibitem{tranheden2021dacs}
Tranheden, W., Olsson, V., Pinto, J., Svensson, L.: {DACS: Domain Adaptation via Cross-Domain Mixed Sampling}. In: Proc.\ of WACV. pp. 1379--1389 (Jan 2021)

\bibitem{tsai2018learning}
Tsai, Y.H., Hung, W.C., Schulter, S., Sohn, K., Yang, M.H., Chandraker, M.: {Learning to Adapt Structured Output Space for Semantic Segmentation}. In: Proc.\ of CVPR. pp. 7472--7481 (Jun 2018)

\bibitem{vidit2023clip}
Vidit, V., Engilberge, M., Salzmann, M.: Clip the gap: A single domain generalization approach for object detection. In: Proceedings of the IEEE/CVF Conference on Computer Vision and Pattern Recognition. pp. 3219--3229 (2023)

\bibitem{Wang_2023_CVPR}
Wang, T.: Learning to detect and segment for open vocabulary object detection. In: Proc.\ of CVPR. pp. 7051--7060 (June 2023)

\bibitem{wang2023internimage}
Wang, W., Dai, J., Chen, Z., Huang, Z., Li, Z., Zhu, X., Hu, X., Lu, T., Lu, L., Li, H., et~al.: Internimage: Exploring large-scale vision foundation models with deformable convolutions. In: Proc.\ of CVPR. pp. 14408--14419 (2023)

\bibitem{Wang2023}
Wang, W., Bao, H., Dong, L., Bjorck, J., Peng, Z., Liu, Q., Aggarwal, K., Mohammed, O.K., Singhal, S., Som, S., Wei, F.: Image as a foreign language: Beit pretraining for vision and vision-language tasks. In: Proc.\ of CVPR. pp. 19175--19186 (June 2023)

\bibitem{wang2020differential}
Wang, Z., Yu, M., Wei, Y., Feris, R., Xiong, J., Hwu, W., Huang, T.S., Shi, H.: {Differential Treatment for Stuff and Things: A Simple Unsupervised Domain Adaptation Method for Semantic Segmentation}. In: Proc.\ of CVPR. pp. 12635--12644 (Jun 2020)

\bibitem{wei2024stronger}
Wei, Z., Chen, L., Jin, Y., Ma, X., Liu, T., Ling, P., Wang, B., Chen, H., Zheng, J.: Stronger fewer \& superior: Harnessing vision foundation models for domain generalized semantic segmentation. In: Proceedings of the IEEE/CVF Conference on Computer Vision and Pattern Recognition. pp. 28619--28630 (2024)

\bibitem{wu2022single}
Wu, A., Deng, C.: Single-domain generalized object detection in urban scene via cyclic-disentangled self-distillation. In: Proceedings of the IEEE/CVF Conference on computer vision and pattern recognition. pp. 847--856 (2022)

\bibitem{wu2024g}
Wu, F., Gao, J., Hong, L., Wang, X., Zhou, C., Ye, N.: G-nas: Generalizable neural architecture search for single domain generalization object detection. In: Proceedings of the AAAI Conference on Artificial Intelligence. vol.~38, pp. 5958--5966 (2024)

\bibitem{Xie2022segformer}
Xie, E., Wang, W., Yu, Z., Anandkumar, A., Alvarez, J.M., Luo, P.: {SegFormer: Simple and Efficient Design for Semantic Segmentation with Transformers}. In: Proc.\ of NeurIPS. pp. 12077--12090 (Dec 2021)

\bibitem{xie2022simmim}
Xie, Z., Zhang, Z., Cao, Y., Lin, Y., Bao, J., Yao, Z., Dai, Q., Hu, H.: Simmim: A simple framework for masked image modeling. In: Proc.\ of CVPR. pp. 9653--9663 (2022)

\bibitem{xu2022dirl}
Xu, Q., Yao, L., Jiang, Z., Jiang, G., Chu, W., Han, W., Zhang, W., Wang, C., Tai, Y.: {DIRL: Domain-Invariant Representation Learning for Generalizable Semantic Segmentation}. In: Proc.\ of AAAI. pp. 2884--2892 (Jun 2022)

\bibitem{yamada2022does}
Yamada, Y., Otani, M.: Does robustness on imagenet transfer to downstream tasks? In: Proc.\ of CVPR. pp. 9215--9224 (2022)

\bibitem{yu2020bdd100k}
Yu, F., Chen, H., Wang, X., Xian, W., Chen, Y., Liu, F., Madhavan, V., Darrell, T.: Bdd100k: A diverse driving dataset for heterogeneous multitask learning. In: Proc.\ of CVPR. pp. 2636--2645 (2020)

\bibitem{yu2023fcclip}
Yu, Q., He, J., Deng, X., Shen, X., Chen, L.C.: Convolutions die hard: Open-vocabulary segmentation with single frozen convolutional clip. In: Proc.\ of NeurIPS (2023)

\bibitem{Yue2019}
Yue, X., Zhang, Y., Zhao, S., Sangiovanni-Vincentelli, A., Keutzer, K., Gong, B.: {Domain Randomization and Pyramid Consistency: Simulation-to-Real Generalization Without Accessing Target Domain Data}. In: Proc.\ of ICCV. pp. 2100--2110 (Oct 2019)

\bibitem{yue2019domain}
Yue, X., Zhang, Y., Zhao, S., Sangiovanni-Vincentelli, A., Keutzer, K., Gong, B.: Domain randomization and pyramid consistency: Simulation-to-real generalization without accessing target domain data. In: Proc.\ of ICCV. pp. 2100--2110 (2019)

\bibitem{zeng2022x}
Zeng, Y., Zhang, X., Li, H., Wang, J., Zhang, J., Zhou, W.: X $^{}2$-vlm: All-in-one pre-trained model for vision-language tasks. arXiv preprint arXiv:2211.12402  (2022)

\bibitem{zhang2021multiple}
Zhang, K., Sun, Y., Wang, R., Li, H., Hu, X.: Multiple fusion adaptation: A strong framework for unsupervised semantic segmentation adaptation. arXiv preprint arXiv:2112.00295  (2021)

\bibitem{zhang2021prototypical}
Zhang, P., Zhang, B., Zhang, T., Chen, D., Wang, Y., Wen, F.: {Prototypical Pseudo Label Denoising and Target Structure Learning for Domain Adaptive Semantic Segmentation}. In: Proc.\ of CVPR. pp. 12414--12424 (2021)

\bibitem{zhao2022style}
Zhao, Y., Zhong, Z., Zhao, N., Sebe, N., Lee, G.H.: {Style-Hallucinated Dual Consistency Learning for Domain Generalized Semantic Segmentation}. In: Proc.\ of ECCV. pp. 535--552 (2022)

\bibitem{zhong2022regionclip}
Zhong, Y., Yang, J., Zhang, P., Li, C., Codella, N., Li, L.H., Zhou, L., Dai, X., Yuan, L., Li, Y., et~al.: Regionclip: Region-based language-image pretraining. In: Proc.\ of CVPR. pp. 16793--16803 (2022)

\bibitem{zhong2022adversarial}
Zhong, Z., Zhao, Y., Lee, G.H., Sebe, N.: {Adversarial Style Augmentation for Domain Generalized Urban-Scene Segmentation}. In: Proc.\ of NeurIPS. pp. 338--350 (Dec 2022)

\bibitem{Zhou2017}
Zhou, B., Zhao, H., Puig, X., Fidler, S., Barriuso, A., Torralba, A.: {Scene Parsing through ADE20K Dataset}. In: Proc.\ of CVPR. pp. 633--641 (Jul 2017)

\bibitem{zhou2021ibot}
Zhou, J., Wei, C., Wang, H., Shen, W., Xie, C., Yuille, A., Kong, T.: ibot: Image bert pre-training with online tokenizer. arXiv preprint arXiv:2111.07832  (2021)

\bibitem{Zhou2022ConditionalPL}
Zhou, K., Yang, J., Loy, C.C., Liu, Z.: Conditional prompt learning for vision-language models. In Proc.\ of CVPR pp. 16795--16804 (2022), \url{https://api.semanticscholar.org/CorpusID:247363011}

\bibitem{Zhou_2023_CVPR}
Zhou, Z., Lei, Y., Zhang, B., Liu, L., Liu, Y.: Zegclip: Towards adapting clip for zero-shot semantic segmentation. In: Proc.\ of CVPR

\bibitem{zoph2020rethinking}
Zoph, B., Ghiasi, G., Lin, T.Y., Cui, Y., Liu, H., Cubuk, E.D., Le, Q.: Rethinking pre-training and self-training. \textnormal{In} Proc.\ of NeurIPS  \textbf{33},  3833--3845 (2020)

\end{thebibliography}
